\documentclass[10pt,twocolumn,letterpaper]{article}

\usepackage[pagenumbers]{cvpr}

\usepackage{amsmath,amssymb,amsthm}
\usepackage{bm}
\usepackage{booktabs}
\usepackage{multirow}
\usepackage{graphicx}
\usepackage[inline]{enumitem}
\usepackage{xcolor}
\usepackage{tcolorbox}
\usepackage{colortbl}
\definecolor{hloursbg}{HTML}{E6F0FA}
\providecommand{\hlours}{\rowcolor{hloursbg}}
\makeatletter
\providecommand{\tablenoplus}{\let\tnp@ensuremath\ensuremath\def\ensuremath##1{\ifx+##1\else\ifx\pm##1\else\tnp@ensuremath{##1}\fi\fi}}
\makeatother
\providecommand{\grayed}[1]{\textit{#1}}

\usepackage{soul}
\definecolor{inkours}{HTML}{1D5FBF}
\definecolor{inkread}{HTML}{0F766E}
\definecolor{inkprior}{HTML}{B45309}
\definecolor{boxband}{HTML}{EAF1FB}
\newcommand{\oursc}[1]{\textbf{#1}}
\newcommand{\readc}[1]{\emph{#1}}
\newcommand{\priorc}[1]{\emph{#1}}
\newcommand{\keyhl}[1]{{\boldmath\textbf{#1}}}
\usepackage{xspace}
\usepackage{algorithm}
\usepackage{algpseudocode}

\newcommand{\method}{PriorSplit\xspace}

\theoremstyle{plain}
\newtheorem{theorem}{Theorem}
\newtheorem{proposition}{Proposition}
\newtheorem{corollary}{Corollary}
\newtheorem{sprop}{Proposition}

\newtheorem{slemma}[sprop]{Lemma}
\theoremstyle{definition}
\newtheorem{definition}{Definition}

\DeclareMathOperator*{\argmax}{arg\,max}
\DeclareMathOperator{\logit}{logit}

\usepackage{placeins}

\definecolor{linkblue}{rgb}{0.21,0.49,0.74}
\usepackage[pagebackref,breaklinks,colorlinks,allcolors=linkblue]{hyperref}

\hypersetup{pdftitle={Prior-as-Skill: Remember Outside and Read Inside with On-Policy Self-Distillation for Self-Evolving VLM Agents}}
\title{Prior-as-Skill: Remember Outside and Read Inside with On-Policy Self-Distillation for Self-Evolving VLM Agents}

\author{
Yan Lin$^{1}$\quad
Jingyu Sun$^{2,3}$\quad
Yifei Qian$^{4}$\quad
Zhengtao Yao$^{5}$\quad
Yuxia Wang$^{6}$\\
Daizong Liu$^{7}$\quad
Chun Pong Lau$^{8}$\quad
Zhongliang Guo$^{9,*}$\quad
Stephen McGough$^{1}$\\[6pt]
{\normalsize
$^{1}$Newcastle University\quad
$^{2}$University of Manchester\quad
$^{3}$University of Melbourne\quad
$^{4}$University of Nottingham}\\
{\normalsize
$^{5}$University of Southern California\quad
$^{6}$INSAIT, Sofia University ``St. Kliment Ohridski''}\\
{\normalsize
$^{7}$Wuhan University\quad
$^{8}$City University of Hong Kong\quad
$^{9}$University of Aberdeen}
}
\newcommand{\contactlabel}{Contact: }
\newcommand{\firstpagenote}{\begingroup\renewcommand{\thefootnote}{}%
\footnotetext{$^{*}$Corresponding author.}%
\footnotetext{\contactlabel\texttt{y.lin64@ncl.ac.uk}}%
\footnotetext{\phantom{\contactlabel}\texttt{zhongliang.guo@abdn.ac.uk}}\endgroup}

\newcommand{\ShortPS}{73.5}
\newcommand{\ShortOPSD}{51.1}
\newcommand{\ShortBayes}{73.8}

\newcommand{\ZeroOPSD}{45.8}

\newcommand{\IntPS}{98.6}
\newcommand{\IntOPSD}{98.8}
\newcommand{\ConfPS}{96.5}
\newcommand{\ConfOPSD}{98.3}
\newcommand{\LongEightPS}{97.4}
\newcommand{\LongEightOPSD}{97.2}
\newcommand{\LongTTPS}{83.0}
\newcommand{\LongTTOPSD}{80.3}
\newcommand{\GainShortOPSD}{22.3}

\newcommand{\GainShortBest}{4.4}

\newcommand{\MaxGainLessConf}{5.7}

\newcommand{\LossTenLo}{6.5}
\newcommand{\LossTenHi}{14.5}

\newcommand{\AblSumShort}{54.7}

\newcommand{\HeldoutPS}{78.0}
\newcommand{\HeldoutOPSD}{51.2}
\newcommand{\ProtoBayesGapMax}{0.9}
\newcommand{\ProtoOPSDGapMax}{0.3}
\newcommand{\ProtoEnvGapMax}{1.3}
\newcommand{\ProtoAboveBothFrom}{12}
\newcommand{\GateMiss}{5.3}
\newcommand{\GateFalseYes}{2.1}

\newcommand{\DeltaLPriorShort}{\ensuremath{-}0.7}
\newcommand{\DeltaLPriorShortlo}{\ensuremath{-}1.9}
\newcommand{\DeltaLPriorShorthi}{\ensuremath{+}0.5}
\newcommand{\DeltaLPriorHeldout}{\ensuremath{-}29.3}
\newcommand{\DeltaLPriorHeldoutlo}{\ensuremath{-}37.7}
\newcommand{\DeltaLPriorHeldouthi}{\ensuremath{-}20.8}

\newcommand{\DFixBMConf}{\ensuremath{-}9.2}

\newcommand{\DVOneBMConf}{\ensuremath{-}12.9}

\newcommand{\TextPriorBaseConf}{9.1}
\newcommand{\IntTieMargin}{0.4}
\newcommand{\IntTieN}{6}

\newcommand{\MGainShort}{13.3}
\newcommand{\MGainShortlo}{9.4}
\newcommand{\MGainShorthi}{17.4}

\newcommand{\MConfPS}{97.6}

\newcommand{\MGainShortBest}{10.4}

\newcommand{\NGainShort}{21.3}
\newcommand{\NGainShortlo}{16.6}
\newcommand{\NGainShorthi}{25.8}

\newcommand{\BBGainMin}{19.4}
\newcommand{\BBGainMax}{22.2}
\newcommand{\BBBestGainMin}{4.2}

\newcommand{\CGainShort}{16.3}
\newcommand{\CGainShortlo}{13.1}
\newcommand{\CGainShorthi}{19.4}

\newcommand{\CNClusters}{109}

\newcommand{\CGoneMissMin}{2.8}
\newcommand{\CGoneMissMax}{8.6}
\newcommand{\CGoneNFail}{2}
\newcommand{\CGoneNPass}{1}

\newcommand{\CDevNClusters}{62}

\newcommand{\TFDevMissMax}{11.5}

\newcommand{\TFDevFalseYesMax}{29.9}

\newcommand{\TFDevNPassOrder}{0}

\newcommand{\TFOrdNCells}{2917}

\newcommand{\TFOrdShortOPSD}{62.5}

\newcommand{\TFOrdGainShort}{7.3}

\newcommand{\TFNxtNCells}{1531}

\newcommand{\TFNxtGainVOne}{12.3}

\newcommand{\HAOrderGainOPSD}{29.0}
\newcommand{\HAOrderGainOPSDlo}{25.9}
\newcommand{\HAOrderGainOPSDhi}{32.0}
\newcommand{\HAOrderGainBest}{3.9}
\newcommand{\HAOrderGainBestlo}{2.6}
\newcommand{\HAOrderGainBesthi}{5.3}

\newcommand{\HAConflictAcc}{93.0}
\newcommand{\HAConflictAcclo}{88.4}
\newcommand{\HAConflictAcchi}{97.0}

\newcommand{\HAEightConflictAcc}{95.5}

\newcommand{\HANConfOrder}{67}
\newcommand{\HANConfNext}{184}
\newcommand{\HAPriorConf}{0.0}

\newcommand{\HADevOrderRouteHi}{6.7}
\newcommand{\HADevOrderRouteLo}{6.2}
\newcommand{\HAEightDevOrderRoute}{8.2}

\newcommand{\HADevMissMax}{4.7}
\newcommand{\HADevFalseYesMax}{3.5}
\newcommand{\HADevOrderRouteMax}{6.7}
\newcommand{\HADevOrderNPass}{1}
\newcommand{\CGainShortBest}{5.5}
\newcommand{\CGainShortBestlo}{4.0}
\newcommand{\CGainShortBesthi}{7.0}
\newcommand{\CNextGainBest}{14.3}
\newcommand{\CNextGainBestlo}{11.1}
\newcommand{\CNextGainBesthi}{17.6}

\newcommand{\BBInternMiss}{0.0}

\newcommand{\MaxConfLossRealMain}{1.7}

\newcommand{\MaxConfLossShownFlag}{5.2}

\newcommand{\ZeroFrameMOPSD}{53.8}
\newcommand{\ZeroFrameMPD}{59.6}
\newcommand{\ZeroFrameMPS}{69.8}
\newcommand{\ZeroFrameHAOPSD}{52.6}
\newcommand{\ZeroFrameHAPD}{84.5}
\newcommand{\ZeroFrameHAPS}{89.1}

\newcommand{\ZeroFrameCOPSD}{52.0}
\newcommand{\ZeroFrameCPD}{67.6}
\newcommand{\ZeroFrameCPS}{73.5}

\newcommand{\MGainShortBestlo}{7.3}
\newcommand{\MGainShortBesthi}{13.5}

\newcommand{\OracleGapShort}{0.5}

\newcommand{\HAVOneConfOrder}{86.6}
\newcommand{\HAEightVOneConfOrder}{93.0}
\newcommand{\MaskMGainShort}{12.1}

\newcommand{\MaskMOPSDShort}{\ensuremath{+}0.5}

\newcommand{\MaskMOPSDConf}{\ensuremath{+}0.2}

\newcommand{\MaskMPSShort}{\ensuremath{-}0.7}

\newcommand{\MaskMPSConf}{\ensuremath{+}0.6}

\newcommand{\MaskMCheckYesMask}{95.4}
\newcommand{\MaskMCheckNoMask}{99.9}
\newcommand{\PixelCheckAgreeMin}{50.8}
\newcommand{\PixelCheckAgreeMax}{56.0}

\newcommand{\RawNConf}{339}

\newcommand{\RawNRec}{109}
\newcommand{\RawOPSDConfZero}{51.5}
\newcommand{\RawOPSDConf}{70.3}

\newcommand{\RawGainShortOPSD}{5.3}
\newcommand{\RawGainShortOPSDlo}{2.8}
\newcommand{\RawGainShortOPSDhi}{7.8}
\newcommand{\RawGainShortPD}{5.5}
\newcommand{\RawGainShortPDlo}{3.7}
\newcommand{\RawGainShortPDhi}{7.3}
\newcommand{\RawBaseFrameGain}{16.5}
\newcommand{\RawBaseFrameGainlo}{10.6}
\newcommand{\RawBaseFrameGainhi}{22.5}

\newcommand{\RawPSMinusPriorShort}{\ensuremath{+}0.2}
\newcommand{\RawPSMinusPriorShortlo}{\ensuremath{-}0.3}
\newcommand{\RawPSMinusPriorShorthi}{\ensuremath{+}0.6}
\newcommand{\RawCheckMissMin}{20.6}
\newcommand{\RawCheckMissMax}{40.4}
\newcommand{\RawCheckFYMin}{15.2}
\newcommand{\RawCheckFYMax}{25.1}

\newcommand{\RawCheckAgreeBase}{57.1}
\newcommand{\RawCheckAlwaysYes}{69.0}

\newcommand{\HOTGainShort}{11.5}
\newcommand{\HOTGainShortlo}{8.4}
\newcommand{\HOTGainShorthi}{14.5}
\newcommand{\HOSGainShort}{12.2}
\newcommand{\HOSGainShortlo}{8.9}
\newcommand{\HOSGainShorthi}{15.4}
\newcommand{\HOTGainShortPD}{14.2}
\newcommand{\HOTGainShortPDlo}{11.2}
\newcommand{\HOTGainShortPDhi}{17.2}
\newcommand{\HOSGainShortPD}{13.8}
\newcommand{\HOSGainShortPDlo}{10.7}
\newcommand{\HOSGainShortPDhi}{16.8}

\newcommand{\HOTNTrain}{87}
\newcommand{\HOTNAll}{122}
\newcommand{\HOTNSites}{35}
\newcommand{\HOTNQ}{550}
\newcommand{\HOTNConsume}{706}
\newcommand{\HOTNPool}{849}
\newcommand{\HOSNSkills}{25}

\newcommand{\HOSNSites}{10}
\newcommand{\HOSNConsume}{340}
\newcommand{\HOSNPool}{486}

\newcommand{\HOLeadSubMin}{10.9}
\newcommand{\HOLeadSubMax}{12.2}
\newcommand{\HOLeadSubLoMin}{6.9}
\newcommand{\HOProbeMissMax}{1.5}
\newcommand{\HOProbeFYMax}{4.2}

\newcommand{\TriFirstGain}{28.8}
\newcommand{\TriFirstGainlo}{26.1}
\newcommand{\TriFirstGainhi}{31.6}

\newcommand{\TriOrderGain}{28.3}
\newcommand{\TriOrderGainlo}{25.7}
\newcommand{\TriOrderGainhi}{30.8}
\newcommand{\DriftGain}{22.7}
\newcommand{\DriftGainlo}{20.2}
\newcommand{\DriftGainhi}{25.2}
\newcommand{\DriftHalfLife}{400}
\newcommand{\DriftT}{1{,}600}
\newcommand{\TrajPS}{6.2}
\newcommand{\TrajOPSD}{0.0}
\newcommand{\TrajPD}{0.7}
\newcommand{\TrajGain}{6.2}
\newcommand{\TrajGainlo}{2.7}
\newcommand{\TrajGainhi}{10.3}
\newcommand{\TrajN}{97}

\newcommand{\AlfItemPS}{83.1}
\newcommand{\AlfItemOPSD}{70.0}
\newcommand{\AlfItemPD}{60.5}
\newcommand{\AlfNItems}{670}
\newcommand{\AlfNItemGames}{124}
\newcommand{\AlfItemGainOPSD}{13.1}
\newcommand{\AlfItemGainOPSDlo}{11.5}
\newcommand{\AlfItemGainOPSDhi}{14.8}
\newcommand{\AlfItemGainPD}{22.6}
\newcommand{\AlfItemGainPDlo}{20.3}
\newcommand{\AlfItemGainPDhi}{24.9}
\newcommand{\AlfSuccPS}{47.5}
\newcommand{\AlfSuccOPSD}{51.2}
\newcommand{\AlfSuccPD}{45.0}
\newcommand{\AlfSuccRecipe}{48.5}
\newcommand{\AlfSuccNone}{32.8}
\newcommand{\AlfNGames}{134}
\newcommand{\AlfReaderShare}{8.7}
\newcommand{\AlfSuccGainNone}{14.7}
\newcommand{\AlfSuccGainNonelo}{7.5}
\newcommand{\AlfSuccGainNonehi}{22.1}
\newcommand{\AlfStepSavePD}{1.8}
\newcommand{\AlfStepSavePDlo}{0.2}
\newcommand{\AlfStepSavePDhi}{3.3}
\newcommand{\AlfTokSave}{2{,}939}
\newcommand{\AlfTokSavelo}{1{,}118}
\newcommand{\AlfTokSavehi}{4{,}900}

\newcommand{\SwapBaseConf}{52.6}

\newcommand{\MSwapBaseConf}{60.0}

\newcommand{\CalibHAFourNextPSOnline}{56.5}
\newcommand{\CalibHAFourNextCalOnline}{72.1}
\newcommand{\CalibHAFourNextOracleOnline}{73.7}

\newcommand{\CalibHAFourNextPSConf}{97.6}
\newcommand{\CalibHAFourNextCalConf}{83.2}

\newcommand{\CalibHAEightNextPSOnline}{66.7}
\newcommand{\CalibHAEightNextCalOnline}{73.1}
\newcommand{\CalibHAEightNextOracleOnline}{73.7}

\newcommand{\CalibHAEightNextPSConf}{98.6}
\newcommand{\CalibHAEightNextCalConf}{97.3}

\newcommand{\HeadTFMissMax}{0.2}
\newcommand{\HeadTFFYMax}{0.9}

\newcommand{\HeadTFNPass}{3}

\newcommand{\HeadTFPSConf}{81.7}
\newcommand{\HeadTFShort}{71.0}
\newcommand{\HeadTFConf}{89.5}

\newcommand{\HeadTFOracleConf}{89.5}
\newcommand{\HeadTFNConf}{146}

\newcommand{\HANSOrderGainPD}{4.4}
\newcommand{\HANSOrderGainPDlo}{2.9}
\newcommand{\HANSOrderGainPDhi}{6.0}

\newcommand{\HANSSteps}{1220}
\newcommand{\HAOrderSteps}{616}

\newcommand{\HeadHADevMissMax}{0.0}
\newcommand{\HeadHADevFYMax}{0.0}

\newcommand{\HeadHANPass}{3}

\newcommand{\HeadHAPSConf}{94.6}
\newcommand{\HeadHAConf}{97.1}
\newcommand{\HeadHAOracleConf}{97.1}
\newcommand{\HeadHANConf}{93}
\newcommand{\HeadHADeltaConf}{\ensuremath{+}2.5}
\newcommand{\HeadHADeltaConflo}{\ensuremath{+}0.7}
\newcommand{\HeadHADeltaConfhi}{\ensuremath{+}4.8}

\newcommand{\HeadCCDevMissMax}{0.0}
\newcommand{\HeadCCDevFYMax}{0.0}

\newcommand{\HeadCCNPass}{3}

\newcommand{\HeadCCPSConf}{94.8}
\newcommand{\HeadCCConf}{99.6}
\newcommand{\HeadCCOracleConf}{99.6}
\newcommand{\HeadCCNConf}{746}
\newcommand{\HeadCCDeltaConf}{\ensuremath{+}4.8}
\newcommand{\HeadCCDeltaConflo}{\ensuremath{+}3.1}
\newcommand{\HeadCCDeltaConfhi}{\ensuremath{+}6.6}

\newcommand{\HeadCCGainShort}{16.5}
\newcommand{\HeadCCGainShortlo}{13.3}
\newcommand{\HeadCCGainShorthi}{19.6}

\newcommand{\KVNQ}{800}
\newcommand{\KVCheckRecomp}{2.0}
\newcommand{\KVCheckShared}{1.0}
\newcommand{\KVCheckRecompLong}{2.0}
\newcommand{\KVCheckSharedLong}{0.3}

\newcommand{\TvAdN}{72}
\newcommand{\TvAdIn}{63}

\newcommand{\TvBaN}{16}
\newcommand{\TvBaIn}{16}
\newcommand{\TvBbN}{9}
\newcommand{\TvBbIn}{9}

\newcommand{\TvBiN}{5}
\newcommand{\TvBiIn}{5}

\newcommand{\TvCcN}{6}
\newcommand{\TvCcIn}{6}
\newcommand{\TvCdN}{1}
\newcommand{\TvCdIn}{1}

\newcommand{\ThOpsdPiMax}{0.38}

\newcommand{\ThBandAll}{72}
\newcommand{\ThBandInf}{34}
\newcommand{\ThBandInfIn}{31}
\newcommand{\ThBandInfAbove}{3}
\newcommand{\ThBandBelow}{5}

\newcommand{\ThConfLossMain}{1.73}

\newcommand{\ThConfLossDevPred}{1.27}

\newcommand{\ThLibPred}{78.3}
\newcommand{\ThLibMeas}{78.0}

\newcommand{\TvAllIn}{326}
\newcommand{\TvAllN}{351}
\newcommand{\OrderFreeMeanOPSD}{\ensuremath{-}0.01}
\newcommand{\OrderFreeMeanVOne}{0.90}
\newcommand{\OrderConfPastOPSD}{1.7}
\newcommand{\OrderConfPastVOne}{16.2}
\newcommand{\OrderConfPastPS}{3.5}
\newcommand{\OrderNFree}{1800}
\newcommand{\OrderNConf}{519}
\newcommand{\MTraj}{97}
\newcommand{\MWeb}{34}

\begin{document}
\twocolumn[{%
\renewcommand\twocolumn[1][]{#1}%
\maketitle
\begin{center}
\centering
\captionsetup{type=figure}
\includegraphics[width=\textwidth]{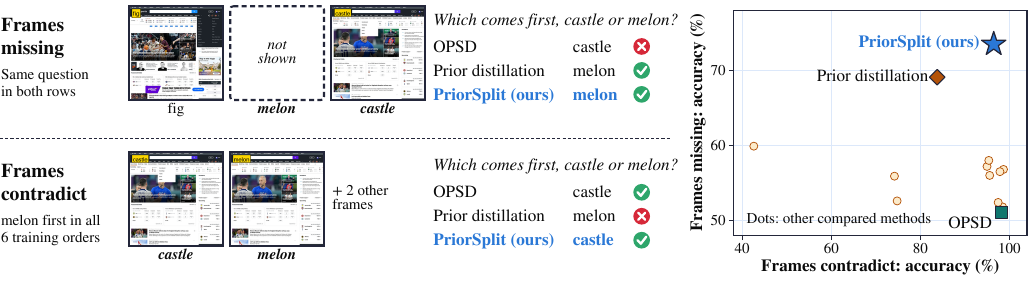}
\captionof{figure}{\textbf{Missing and contradicting frames.} Only PriorSplit is high on both axes. Left: one Mind2Web test question in both rows, picked to favor PriorSplit. Right: ranked methods on synthetic skills with Qwen3-VL-4B. Screenshots~\cite{deng2023mind2web,zheng2024seeact} (OpenRAIL).}
\label{fig:teaser}
\end{center}%

}]
\firstpagenote
\begin{abstract}
Self-evolving visual agents keep what they learn as \textbf{skills}: key frames with step notes on a task's steps and their order. On-policy self-distillation (OPSD) is now a popular way to train such skills into a vision-language model that matches, on its own answers, a teacher that also reads the written order. At deployment, frames often go missing or show an unusual order, as when a person or a website reorders the steps. Existing internalization methods keep any order they store in the frame-reading weights: trained on full frames, the model \textbf{guesses} when they are missing, and trained with frames withheld, it stores the order, so \textbf{memory can override what it sees}. We propose \oursc{PriorSplit}, which fills in missing frames from an order kept outside the model, with no training beyond OPSD. It \priorc{counts which of two neighboring steps comes first in training episodes}, and the model checks if both asked steps are in view: if so, it \readc{answers from the frames}, and otherwise the counts answer. Extensive experiments on synthetic skills with five models and on step-tagged order questions in new episodes of seen web, cooking and egocentric tasks demonstrate that, with frames missing, \keyhl{PriorSplit outperforms every compared method} at OPSD's training cost. With Qwen3-VL-4B on synthetic skills, it also \keyhl{stays close to OPSD} on unusual orders, where methods that gain most by storing the order fall behind.
\end{abstract}

\section{Introduction}
\label{sec:intro}

Self-evolving agents turn past runs into reusable skills~\cite{xia2026skillrl} and increasingly internalize them, so a skill helps after leaving the context window~\cite{skill0,skill052026,skillsd2026}. The agent must then answer from what it \priorc{remembers} and what it \readc{sees}. For a visual agent, a skill is a keyframe sequence with step notes on how a task unfolds on a website~\cite{deng2023mind2web,zheng2024seeact} or in egocentric videos of cooking and other tasks~\cite{captaincook4d,holoassist}. On-policy self-distillation (OPSD)~\cite{zhao2026opsd} suits this: a teacher given privileged information supervises the student's own outputs, so the student matches it without that information.

At test time, frames fail the student in two ways. They go \textbf{missing} when context budgets truncate long skills, retrieval returns some of them~\cite{yeo2026worldmm} or a mid-task agent has not seen later steps. They \textbf{contradict} the usual order when a person reorders steps or a website changes its flow. An internalized skill should decide per question whether its usual order or the frames answer.

\begin{figure}[t]
\centering
\includegraphics[width=\columnwidth]{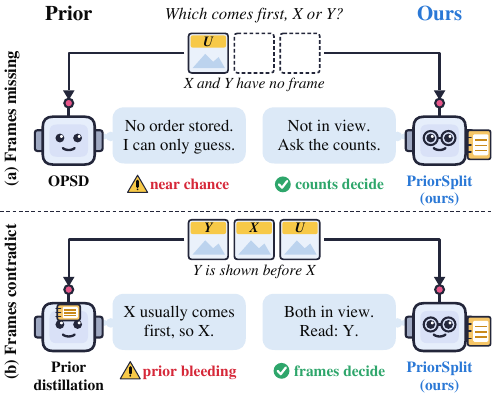}
\caption{\textbf{Two ways frames fail.} PriorSplit lets the counts decide in (a) and the frames in (b). Steps X, Y and U. (a) Frames missing: an evidence-only reader guesses. (b) Frames contradict: a shared-weight reader follows its stored order (prior bleeding).}
\label{fig:motivation}
\end{figure}

As Fig.~\ref{fig:teaser} and Fig.~\ref{fig:motivation} show, existing internalization methods fall into two groups, and neither escapes a choice between guessing and trusting memory over the frames. In the evidence-only group, such as our OPSD baseline~\cite{zhao2026opsd}, the student's frames explain the whole training signal, so its weights need not store the order. On synthetic skills with Qwen3-VL-4B, its students reach $\IntOPSD\%$ with all six frames but $\ZeroOPSD\%$ on two options without them. In the shared-weight group, the student answers from less context, so its weights absorb the skill~\cite{snell2022context,opcd,skill0,nca} and the stored order sits where frames are read. On the same skills, students gaining more than ten short-side points (zero or two of six frames shown) lose $\LossTenLo$ to $\LossTenHi$ points on conflict items, whose frames contradict the usual order. The stored prior's part of this loss is \textbf{prior bleeding} (Section~\ref{sec:revisit}). This raises an intriguing question: can a skill fill in missing frames from its order without storing it in the frame-reading weights?

An ideal method should (i) answer from the usual order without decisive frames, which evidence-only methods cannot, as their weights hold almost no order. It should also (ii) keep that order from overriding frames that decide the answer, which shared-weight students fail to do once they store much order, as their loss does not ask to switch it off.

The answer is positive: \oursc{PriorSplit} keeps the order outside the frame-reading weights. To meet (i), it \priorc{counts} the privileged orders the OPSD teacher already sees, and the counts answer when a named step is out of view. To meet (ii), the OPSD reader, trained on full frame windows, checks per question if both named steps are in view. If they are, it \readc{reads the frames} alone, so the prior can override them only on a check miss, as Theorem~\ref{thm:switch}(a) quantifies. Both parts add no training beyond OPSD. This follows \textbf{\mbox{Occam's} razor}: counting the order takes no training, so it need not live in the frame-reading weights. Like the parameter-free shortcut of ResNet~\cite{he2016resnet}, what need not be learned bypasses the weights, which only read the frames.

Our main contributions are as follows.

\begin{itemize}
\item We revisit skill internalization with OPSD and reveal prior bleeding: an order stored in the frame-reading weights overrides frames that contradict it. Our log-odds model identifies this stored prior from frame-free scores.
\item We propose PriorSplit, which adds no training beyond OPSD: counted orders outside the reader answer when its check finds an asked step out of view. We derive this switch from a Bayes posterior with the counts as its prior, and an identity of its routes predicts the conflict loss.
\item PriorSplit leads every compared method with frames missing at OPSD's training budget, on synthetic skills with five models, new episodes of seen web and cooking tasks and order questions of seen egocentric tasks.
\end{itemize}

\section{Related Work}
\label{sec:related}

\noindent\textbf{Skills for self-evolving agents.} Agents distill experience into reusable skills retrieved by context~\cite{xia2026skillrl}. AtlasVA~\cite{atlasva2026} uses a visual skill memory as shaping rewards, and EgoProceVQA~\cite{egoprocevqa2026} asks key-step questions about egocentric procedures. MMSkills~\cite{mmskills2026} builds state-card and keyframe skills from agent trajectories and checks them live at inference. ContractSkill~\cite{contractskill2026} turns web-agent skills into explicit contracts that are verified and locally repaired outside the model. Others internalize skills. SKILL0~\cite{skill0} removes the full skill from context over training, Skill0.5~\cite{skill052026} internalizes general skills and keeps task-specific ones in context, and several use a skill as privileged teacher context in on-policy distillation~\cite{skillsd2026,opid2026,seed2026}. UCOB~\cite{ucob2026} picks skill-conditioned or skill-free views as a local teacher, and LatentSkill~\cite{latentskill2026} maps textual skills to weights with a hypernetwork. Chen \etal~\cite{rethinking2026} find repeated experience internalization can collapse. These methods store the skill in the weights that read the context, at a cost we measure for ordered visual skills.

\noindent\textbf{On-policy and context distillation.} Distillation~\cite{hinton2015distilling} with privileged information~\cite{vapnik2009lupi} lets a student learn from a teacher that sees more. Context distillation~\cite{snell2022context} drops the prompt from the student, on-policy distillation trains on the student's own samples~\cite{agarwal2024onpolicy}, on-policy context distillation~\cite{opcd} combines the two, and OPSD~\cite{zhao2026opsd} conditions the teacher on privileged reference solutions. Self-distillation serves continual learning~\cite{continualsd2026} and is studied for overconfidence~\cite{antisd2026}, GATES~\cite{gates2026} gates distillation by tutor consensus, and Imagine-OPD~\cite{imagineopd2026} gives the teacher zoomed views. Closest to us, NCA~\cite{nca} adds a consistency regularizer because an internalized student degrades when the context returns, and cross-world distillation~\cite{cwopd} matches the teacher's belief change across two visual worlds differing in answer-critical evidence. Unlike PriorSplit, they keep prior and evidence in shared weights.

\noindent\textbf{Contrasting contexts and replacing priors.} Context-aware decoding~\cite{shi2024cad,wang2025adacad}, prior-aware decoding~\cite{pad2024} and contextual calibration~\cite{zhao2021calibrate} contrast outputs on full and reduced inputs to remove prior beliefs, and knowledge-conflict studies ask when context beats parametric knowledge~\cite{xie2024chameleon,zhou2023contextfaithful}. Vision-language models default to the image, but a few attention heads let memorized knowledge override it~\cite{lietzow2026vision}. PhD~\cite{liu2025phd} tests whether misleading text or counter-common-sense images pull answers away from what the image shows. Our conflict items ask the same of step order. Logit arithmetic with experts~\cite{liu2021dexperts,proxytuning2024} mixes models at inference, bias-only branches~\cite{clark2019easy,cadene2019rubi,niu2021cfvqa} join them in training, and prior-shift correction replaces a classifier's training prior~\cite{saerens2002adjusting,menon2021logit}. Task graphs learned from egocentric video give structured step priors~\cite{seminara2024tgml}. Video-LLMs separate spatial and temporal tokens to curb action-scene hallucination~\cite{bae2025mashvlm}, and multi-image models soften the causal mask between images or calibrate logits with attention to reduce position bias~\cite{tian2025position,xian2026lad}. Selective classification~\cite{geifman2017selective} abstains at low confidence, and mixtures of experts~\cite{jacobs1991adaptive,shazeer2017outrageously}, Octopus~\cite{suo2025octopus} and a long-video agent~\cite{yeo2026worldmm} route by a trained gate, a trained decoding selector or a prompted memory pick. PriorSplit trains no gate and routes each question by its reader's own check.

\section{Revisiting Skill Internalization}
\label{sec:revisit}

\begin{figure}[t]
\centering
\includegraphics[width=\columnwidth]{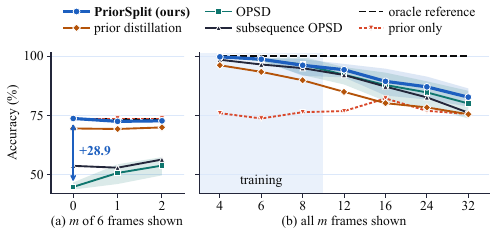}
\caption{\textbf{Few frames and long windows.} PriorSplit stays within $\ProtoEnvGapMax$ points of the better of OPSD and prior only (the counts alone) at every $m$ and is above both from $m = \ProtoAboveBothFrom$. Accuracy (\%) by shown frames $m$ of a $K$-frame window, seed mean and range.}
\label{fig:protocol}
\end{figure}

\subsection{Limitations of Evidence-Only Internalization}
\label{sec:revisit_reader}

We revisit why internalization fails on contexts shorter than in training, with Qwen3-VL-4B on the synthetic test set (Section~\ref{sec:setup}). In Fig.~\ref{fig:protocol}, Table~\ref{tab:main} and Table~\ref{tab:s_protocol}, the OPSD~\cite{zhao2026opsd} student, trained on full frame sequences, reads almost perfectly on windows of the training range and answers near chance without decisive frames. \textbf{\emph{The teacher always sees the privileged order, but the student's frames explain the whole signal, so evidence-only OPSD stores almost no order and guesses without decisive frames.}}

\subsection{Limitations of Shared-Weight Internalization}
\label{sec:revisit_shared}

\begin{table}[t]
\centering
\footnotesize
\setlength{\tabcolsep}{2.5pt}
\caption{\textbf{Shared-weight students.} Each shared-weight student gains on the short side and loses on conflict items, more as it gains more. Accuracy (\%), Qwen3-VL-4B, and $\Delta$ to OPSD on the same seeds and questions. Italic rows: questions neither all-pair variant trained on, where OPSD has 50.7 short and 98.4 conflict.}
\label{tab:revisit}
\begin{tabular}{lrrrr}
\toprule
Method & Short & $\Delta$ & Conflict & $\Delta$ \\
\midrule
OPSD & 51.1 & 0.0 & 98.3 & 0.0 \\
Subsequence OPSD & 56.0 & 4.9 & 95.6 & $-$2.7 \\
\grayed{All-pair subsequence OPSD (same steps)} & \grayed{61.1} & \grayed{10.3} & \grayed{91.9} & \grayed{$-$6.5} \\
\grayed{All-pair subsequence OPSD (5$\times$ steps)} & \grayed{66.0} & \grayed{15.3} & \grayed{90.2} & \grayed{$-$8.1} \\
Prior distillation & 69.1 & 17.9 & 83.8 & $-$14.5 \\
\bottomrule
\end{tabular}
\end{table}

\begin{figure}[t]
\centering
\includegraphics[width=\columnwidth]{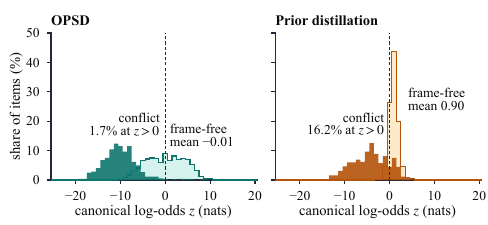}
\caption{\textbf{Where the order is stored.} Prior distillation moves its frame-free scores to the canonical side, and $\OrderConfPastVOne\%$ of its conflict scores cross zero against $\OrderConfPastOPSD\%$ for OPSD. Log-odds $z$ on frame-free (light) and conflict items (solid), Qwen3-VL-4B.}
\label{fig:order_store}
\end{figure}

Subsequence OPSD (its student sees a random subsequence of the window), its all-pair variants, skill-shrinking curricula~\cite{skill0} and prior distillation (which distills the order into the reader) apply the natural remedy. The student answers from less context, so its weights store the order. This assumes the stored prior does not interfere with the frames, which may fail when they contradict the usual order. In Table~\ref{tab:revisit}, every such student gains on the short side (zero or two of six frames) and loses on conflict items, interval questions whose order departs from the canonical one, and the loss grows with the gain. Trained five times longer, subsequence OPSD and prior distillation do the same and stay below PriorSplit on both (Table~\ref{tab:s_budget}). Theorem~\ref{thm:identify} separates the stored prior's part, prior bleeding, from worse reading.

A reader scores a question on the canonical pair $X$ before $Y$ by its canonical log-odds $z(V) = \log p(X \mid q, V) - \log p(Y \mid q, V)$, with $\delta$ and $\mathrm{vis}(V)$ as in Section~\ref{sec:formulation}.

\begin{theorem}[What stored scores identify]
\label{thm:identify}
For a reader with canonical log-odds $z(V) = e(V) + \pi(1 - \rho\,\mathrm{vis}(V))$, evidence term $e$, stored prior $\pi$, switch quality $\rho \in [0,1]$ and $\kappa = (1-\rho)\pi$, let $Z_F, Z_N, Z_C$ be $z$ on random frame-free, visible non-conflict and conflict items of one $\delta$, with $P(z=0)=0$ on each. If $\mathbb{E}[e] = 0$ without frames and the margin toward the gold has one law on both visible sets, (i) $\pi = \mathbb{E}Z_F$, (ii) the conflict accuracy is $P(Z_N > 2\kappa)$, between $P(Z_N > 2\pi)$ and $P(Z_N > 0)$, and (iii) with the prior off on visible items and no tie at $e = 0$ it is $P(Z_N > \kappa)$.
\end{theorem}
By part (i), OPSD stores at most $\ThOpsdPiMax$ nats of frame-free prior per $\delta$ on four synthetic backbones. As Fig.~\ref{fig:order_store} shows, its frame-free log-odds are wide and symmetric with mean $\OrderFreeMeanOPSD$ nats, against $\OrderFreeMeanVOne$ nats for prior distillation. By part (ii), $\ThBandInfIn$ of the $\ThBandInf$ student cases there with a band over five points wide are consistent with ${\rho = 0}$, weights that never switch the prior off. Training asks no more: with a canonical majority, the cross-entropy on visible items is least with the prior not switched off (Proposition~\ref{sprop:incentive}). \textbf{\emph{Once shared-weight students store much order in the frame-reading weights, they lose conflict accuracy, and the loss grows with the short-side gain.}}

\subsection{Limitations of Two Simpler Fixes}
\label{sec:revisit_fixes}

Counting adjacent step pairs in the privileged orders gives an order prior with no training, and two simple ways to combine it with the reader each give up one side (Table~\ref{tab:ablation}). Added to the reader's scores, the counts are outweighed by those scores without decisive frames, so reader plus prior reaches $\AblSumShort\%$ on the short side, near OPSD's $\ShortOPSD\%$. Written into the prompt, the counts override the frames, and the base model then answers $\TextPriorBaseConf\%$ of conflict items correctly. \textbf{\emph{PriorSplit therefore keeps the prior outside the reader and combines the two only at inference, where a check of the named steps decides which one answers.}}

\section{Methodology}
\label{sec:method}

\begin{figure*}[t]
\centering
\includegraphics[width=\textwidth]{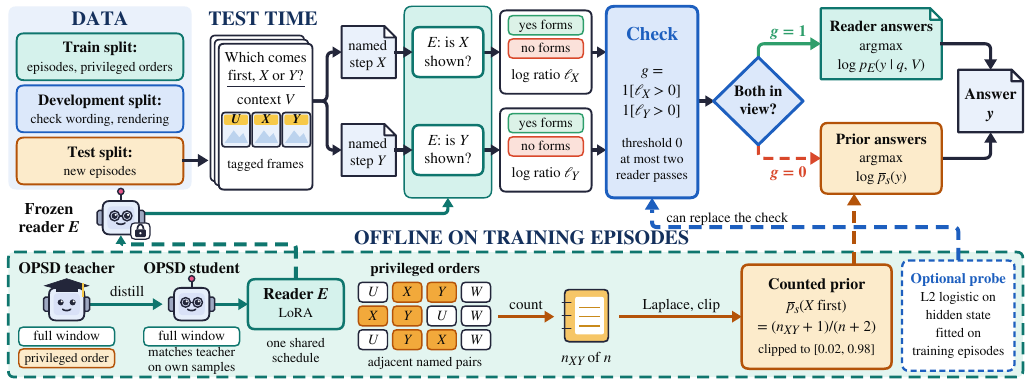}
\caption{\textbf{PriorSplit.} The counted prior answers when the check finds a named step out of view, and the reader answers otherwise. Offline: OPSD trains the reader $E$ and privileged orders give $\bar p_s$. Test time: $E$ (or the optional probe) checks the named steps and $g$ picks $E$ or $\bar p_s$.}
\label{fig:method}
\end{figure*}

\subsection{Problem Formulation}
\label{sec:formulation}

As Fig.~\ref{fig:method} shows, PriorSplit keeps the prior out of the reader and trains nothing beyond OPSD. A visual skill $s$ is a canonical sequence of units $u_1,\dots,u_L$, each a keyframe with a short step note. An episode shows a window of $K$ units in a presented order $S$, the canonical order under order noise of per-skill strength $\delta$. A question $q$ names two canonically adjacent units $X$ and $Y$ and asks which comes first in $S$. At test time the context $V$ is an order-preserving subsequence of $S$ with $m \ge 0$ frames. We write $\mathrm{vis}(V)=1$ when both named units are in $V$, and $a_s=P(X \text{ before } Y \text{ in } S)$ for the canonical order $X$ before $Y$. A conflict item is a visible question with $Y$ first in $S$.

\begin{definition}[Oracle reference]
\label{def:oracle}
This reference answers by the presented order when both named units are in view and by the prior otherwise:
$P(y^\ast = X \mid V) = \mathbf{1}[X \text{ before } Y \text{ in } V]$ if $\mathrm{vis}(V)=1$, and $a_s$ otherwise.
\end{definition}

Definition~\ref{def:oracle} is a posterior. Let $T(V)$ record $\mathrm{vis}(V)$ and, when both named units are in view, which of them comes first in $V$. Bayes' rule in log-odds splits the posterior into a frame term and the skill prior,
\begin{equation}
\logit P(y^\ast{=}X \mid T) = \underbrace{\log\frac{P(T \mid y^\ast{=}X)}{P(T \mid y^\ast{=}Y)}}_{\Lambda(T)} + \logit a_s .
\label{eq:posterior}
\end{equation}

\begin{proposition}[Bayes reader]
\label{prop:bayes}
Let $\mathrm{vis}(V)$ be independent of the presented order of the named pair. (i) $\Lambda(T)$ is $+\infty$ if $X$ comes first in $V$, $-\infty$ if $Y$ comes first in $V$, and $0$ if $\mathrm{vis}(V)=0$, so Definition~\ref{def:oracle} is the posterior of Eq.~\eqref{eq:posterior}. (ii) With $v = P(\mathrm{vis}(V)=1)$, its most probable answer is right with probability $A^\ast = v + (1-v)\max(a_s, 1-a_s)$, and no rule that uses the frames through $T$ is right more often.
\end{proposition}
The frame term is all or nothing, so the Bayes reader never averages frames and prior: it switches between them. The condition is exact with no frame and with full windows, and Section~\ref{sec:s_bayes} proves the general form.

\subsection{Reader}
\label{sec:reader}

The reader $E$ is the OPSD student~\cite{zhao2026opsd} of Fig.~\ref{fig:method}, trained on the shared schedule (Section~\ref{sec:s_training}) with a full window, since on the main model shorter contexts store the prior where it overrides contradicting frames (Section~\ref{sec:revisit_shared}).

\subsection{Counted Order Prior}
\label{sec:prior}

Eq.~\eqref{eq:posterior} needs $a_s$, which the OPSD reader barely stores (Section~\ref{sec:revisit_reader}), so we estimate it from the privileged orders. Let $n$ episodes hold an adjacent pair $(X,Y)$, with $X$ first in $n_{XY}$ of them. We model these orders as independent given $\theta = P(X \text{ first})$ and place a uniform prior on $\theta$. The posterior is $\mathrm{Beta}(n_{XY}+1,\, n-n_{XY}+1)$, and its mean, the probability that $X$ comes first in the next episode, is
\begin{equation}
\bar p_s(X \text{ first}) = \mathbb{E}[\theta \mid n_{XY}, n] = \frac{n_{XY}+1}{n+2},
\label{eq:prior}
\end{equation}
clipped to $[0.02, 0.98]$, so rare pairs stay near $1/2$. Every symmetric prior $\mathrm{Beta}(c,c)$ puts this mean on the side of the larger count, and only this side reaches an answer (Lemma~\ref{slem:direction}). The generator's orders are independent but not identically distributed, the case that Theorem~\ref{thm:switch}(b) bounds. A frame-free adapter fitted to the counts changes short-side accuracy by $\DeltaLPriorShort$ points (Table~\ref{tab:s_learned_prior}).

\subsection{Check and Switch}
\label{sec:recombine}

The check asks $E$ once per named unit $u$ whether $u$ is shown in the same context (wordings in Table~\ref{tab:s_wordings}). Its log-odds $\ell_u$ is the log ratio of summed yes and no probabilities at the first answer token, and $g=\mathbf{1}[\ell_X>0]\,\mathbf{1}[\ell_Y>0]$. No visibility label trains the check, and its threshold is zero everywhere without tuning. On the real settings, development data chose only its wording, frame rendering, tag list and number of next-step check questions (Section~\ref{sec:s_priorsplit}). PriorSplit answers by
\begin{equation}
s(y) = g\log p_E(y \mid q, V) + (1-g)\log \bar p_s(y)
\label{eq:score}
\end{equation}
for each option $y$, with ties to the option named first. For exchangeable orders, Eq.~\eqref{eq:posterior} given the counts holds with the unclipped count of Eq.~\eqref{eq:prior} in place of $a_s$ (Section~\ref{sec:s_bayes}). Under Proposition~\ref{prop:bayes}'s condition, Eq.~\eqref{eq:score} is this posterior with $g$ for $\mathrm{vis}(V)$ and the reader's answer for the unbounded frame term $\Lambda$. For next-step questions, $\bar p_s$ counts which option follows the named step, whose presented successor is the gold, and $g=1$ when the check finds the named step and another shown step after it (Section~\ref{sec:s_m2w}). The check adds at most two reader passes. With the shared prompt encoded once they cost the time of $\KVCheckShared$ answer passes per synthetic question and $\KVCheckSharedLong$ on 32-frame windows (Section~\ref{sec:s_priorsplit}). A probe can replace this check on order questions with Qwen3-VL-4B readers: per seed, an L2-regularized logistic regression on the frozen reader's final hidden state at the check prompt, fitted on training episodes and labeled by the shown steps. Development data choose regularization and threshold.

\subsection{Theoretical Analysis}
\label{sec:theory}

\begin{theorem}[Switch and counts]
\label{thm:switch}
(a) Let the check miss a share $\alpha$ of some visible questions and pass a share $\beta$ of some out of view, with $A_E$ and $A_{\bar p}$ the accuracies of the reader and of the prior on the misrouted ones. Then \method is below the reader by exactly $\alpha(A_E - A_{\bar p}) \le \alpha$ on the first and below the prior by $\beta(A_{\bar p} - A_E)$ on the second. (b) Let the $n_p$ counted orders of a pair $p$ be independent given their windows and the test item, with mean canonical probability $\bar q_p > \frac12$. On a frame-free item of $p$, canonical with probability $a_s \ge \frac12$ and named by a fair coin, the prior and \method with a correct check are right with probability at most $a_s$ and at least $a_s - (2a_s - 1)\exp\big(-2n_p(\bar q_p - \tfrac12)^2\big)$.
\end{theorem}
\begin{corollary}[Regret to the oracle reference]
\label{cor:regret}
On a set of questions with visible share $v$, let $R_{\mathrm{vis}}$ be the reader's accuracy on the visible ones. Let $P_{\mathrm{inv}}$ and $a^\ast$ be the accuracies of the counted prior and of the oracle reference on the others, and $L_\alpha$, $L_\beta$ the two losses of Theorem~\ref{thm:switch}(a). The accuracies $A$ of \method and $A^\ast$ of the oracle reference satisfy
\begin{equation}
\begin{aligned}
A^\ast - A ={}& \underbrace{v(1 - R_{\mathrm{vis}})}_{\text{reading}} + \underbrace{vL_\alpha + (1-v)L_\beta}_{\text{check}} \\
&+ \underbrace{(1-v)(a^\ast - P_{\mathrm{inv}})}_{\text{counting}} .
\end{aligned}
\label{eq:regret}
\end{equation}
With a correct check, under the conditions of Proposition~\ref{prop:bayes} and of Theorem~\ref{thm:switch}(b), the expected regret on a pair is at most $v(1 - R_{\mathrm{vis}}) + (1-v)(2a_s - 1)\exp\big(-2n_p(\bar q_p - \tfrac12)^2\big)$.
\end{corollary}
Each part of \method owns one term, and the counting bound decays exponentially in $n_p$. A correct check keeps the prior off every visible item, the case $\rho = 1$ (Proposition~\ref{sprop:frontier}). Conflict accuracy is the worst visible accuracy of a shared-weight reader favoring the usual order (Proposition~\ref{sprop:shift}). On development data, part (a) predicts the test conflict loss within its interval on $\TvBiIn$ of $\TvBiN$ sets (Table~\ref{tab:s_ledger}). Proposition~\ref{sprop:tolerable} gives the check errors below which \method beats both parts. Proofs are in Section~\ref{sec:s_proofs}.

\section{Experiments}
\label{sec:experiments}

\begin{figure*}[t]
\centering
\includegraphics[width=\textwidth]{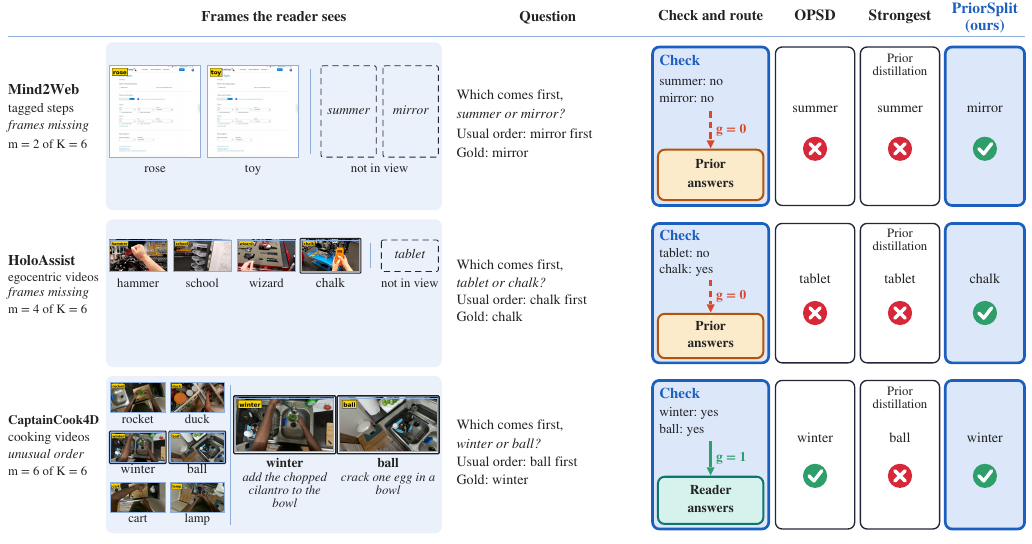}
\caption{\textbf{Missing and contradicting real frames.} PriorSplit is right in every row and the strongest compared method in none. Frames the reader saw, in order, with their banner words. Qwen3-VL-4B readers, seed 0. Rows 1 and 2 lack a frame of a named step, and row 3 shows an unusual order. Strongest: strongest compared method. Each row is picked to favor PriorSplit. Multimodal-Mind2Web~\cite{deng2023mind2web,zheng2024seeact} (OpenRAIL), HoloAssist~\cite{holoassist} (CDLA-Permissive-2.0) and CaptainCook4D~\cite{captaincook4d} (Apache 2.0) frames.}
\label{fig:qualitative}
\end{figure*}

\subsection{Experimental Setup}
\label{sec:setup}

\noindent\textbf{Skills and models.} Readers use Qwen3-VL-4B~\cite{qwen3vl} (main model), Qwen2.5-VL-7B~\cite{qwen25vl}, Qwen3-VL-8B~\cite{qwen3vl}, Idefics3-8B~\cite{idefics3} and InternVL3.5-8B~\cite{internvl35}. Nine synthetic skills (Section~\ref{sec:s_synth}) of $32$ rendered word cards, three per order noise $\delta \in \{0.15, 0.3, 0.5\}$, train on $K \in [4,8]$-frame windows. Tests show zero or two of six window frames (short side), all six (interval) and full $8$- and $32$-frame windows (Long8 and Long32). Real skills are $97$ Multimodal-Mind2Web trajectories~\cite{zheng2024seeact} from $34$ websites (Section~\ref{sec:s_m2w}). Each screenshot gets a banner with the step name that questions use, and we ask the same order questions and the next of four actions. Each question names its skill, which selects the count table. Test episodes are kept apart from training and development (Section~\ref{sec:s_baselines}).

\noindent\textbf{Baselines.} At equal information we compare OPSD~\cite{zhao2026opsd}, the shared-weight students of Table~\ref{tab:main}, including subsequence OPSD (Section~\ref{sec:revisit_shared}), its all-pair variants at the same or $5\times$ the steps and prior distillation. Training-free baselines are CAD~\cite{shi2024cad} with $\gamma = 0.5$, its limit pointwise mutual information (PMI)~\cite{holtzman2021surface} and PAD~\cite{pad2024} (Section~\ref{sec:s_baselines}). Every trained method learns from the same order questions, on Mind2Web and CaptainCook4D next-step questions only test them, and PriorSplit counts next steps in the same privileged orders.

\noindent\textbf{Protocol.} Methods answer by the higher option score and share one LoRA~\cite{hu2022lora} schedule, OPSD included (Section~\ref{sec:s_training}). Paired bootstrap intervals resample test questions or real episodes against same-seed OPSD with seeds held fixed (Section~\ref{sec:s_baselines}). The strongest compared method of a setting has the highest short-side accuracy among the training-free corrections and methods with OPSD's step budget run on it.

\subsection{Main Results}
\label{sec:main_results}
\begin{table*}[t]
\centering
\footnotesize
\setlength{\tabcolsep}{4pt}
\renewcommand{\arraystretch}{0.95}
\setlength{\aboverulesep}{0.2ex}\setlength{\belowrulesep}{0.3ex}
\caption{\textbf{Results on synthetic skills.} PriorSplit leads every ranked method in each short-side column, Long8 and Long32. Accuracy (\%) of Qwen3-VL-4B over three training seeds. Short pools 0, the first 2 (Prefix 2) and 2 strided (Stride 2) of six frames, Interval all six, Conflict its out-of-order items, Long8 and Long32 hold 8 and 32 frames. Bold, underline: top two per column. Subscripts: seed standard deviations. Italic rows are not ranked: all-pair rows score untrained questions, the oracle reference uses true visibility and presented order.}
\label{tab:main}
\begin{tabular}{lrrrrrrrr}
\toprule
 & \multicolumn{4}{c}{Short side} & \multicolumn{4}{c}{Full context} \\
\cmidrule(lr){2-5} \cmidrule(lr){6-9}
Method & Short & 0 frames & Prefix 2 & Stride 2 & Interval & Conflict & Long8 & Long32 \\
\midrule
\multicolumn{9}{l}{\textbf{\emph{Evidence-only}}} \\
OPSD~\cite{zhao2026opsd} & 51.1\,{\scriptsize$\pm$2.6} & 45.8 & 57.7 & 49.9 & \textbf{98.8} & 98.3\,{\scriptsize$\pm$0.6} & 97.2 & 80.3 \\
\midrule
\multicolumn{9}{l}{\textbf{\emph{Shared-weight}}} \\
SFT on gold answers & 56.8\,{\scriptsize$\pm$1.2} & 54.2 & 61.8 & 54.6 & 98.5 & \textbf{98.7}\,{\scriptsize$\pm$0.7} & \underline{97.3} & \underline{82.8} \\
SFT on teacher answers & 57.2\,{\scriptsize$\pm$1.3} & 54.1 & 62.1 & 55.4 & 96.5 & 95.0\,{\scriptsize$\pm$2.3} & 95.2 & 76.9 \\
Context distillation~\cite{snell2022context} & 59.9\,{\scriptsize$\pm$0.9} & 58.9 & 63.9 & 57.0 & 61.8 & 42.6\,{\scriptsize$\pm$3.8} & 65.3 & 62.6 \\
Subsequence OPSD & 56.0\,{\scriptsize$\pm$2.1} & 53.9 & 59.8 & 54.3 & 96.2 & 95.6\,{\scriptsize$\pm$1.5} & 94.6 & 76.4 \\
\grayed{All-pair subsequence OPSD (same steps)} & \grayed{61.1\,{\scriptsize$\pm$1.5}} & \grayed{57.9} & \grayed{67.9} & \grayed{57.3} & \grayed{94.8} & \grayed{91.9\,{\scriptsize$\pm$2.3}} & \grayed{95.1} & \grayed{76.8} \\
\grayed{All-pair subsequence OPSD (5$\times$ steps)} & \grayed{66.0\,{\scriptsize$\pm$2.3}} & \grayed{64.0} & \grayed{69.6} & \grayed{64.3} & \grayed{95.7} & \grayed{90.2\,{\scriptsize$\pm$10.7}} & \grayed{95.3} & \grayed{79.4} \\
Prior distillation & \underline{69.1}\,{\scriptsize$\pm$0.6} & \underline{67.1} & \underline{73.3} & \underline{66.9} & 92.7 & 83.8\,{\scriptsize$\pm$4.4} & 91.3 & 75.4 \\
OPCD~\cite{opcd} & 52.6\,{\scriptsize$\pm$1.8} & 47.3 & 59.9 & 50.6 & 81.8 & 74.8\,{\scriptsize$\pm$0.9} & 75.8 & 52.5 \\
SKILL0~\cite{skill0} & 58.0\,{\scriptsize$\pm$1.2} & 56.3 & 64.7 & 52.9 & 96.3 & 95.4\,{\scriptsize$\pm$2.3} & 93.8 & 72.7 \\
NCA~\cite{nca} & 55.9\,{\scriptsize$\pm$1.7} & 54.2 & 58.8 & 54.5 & 83.6 & 74.2\,{\scriptsize$\pm$5.5} & 84.8 & 69.7 \\
CW-OPD~\cite{cwopd} & 56.5\,{\scriptsize$\pm$0.5} & 54.4 & 61.8 & 53.3 & 98.4 & 97.9\,{\scriptsize$\pm$0.3} & 96.4 & 76.1 \\
\midrule
\multicolumn{9}{l}{\textbf{\emph{Inference-time}}} \\
CAD~\cite{shi2024cad} & 51.8\,{\scriptsize$\pm$3.0} & 45.8 & 58.7 & 50.9 & \textbf{98.8} & \underline{98.5}\,{\scriptsize$\pm$0.3} & 97.0 & 80.0 \\
PMI~\cite{holtzman2021surface} & 52.4\,{\scriptsize$\pm$1.1} & 44.2 & 61.2 & 51.8 & 98.5 & 97.5\,{\scriptsize$\pm$1.3} & 95.8 & 74.3 \\
PAD~\cite{pad2024} & 51.3\,{\scriptsize$\pm$2.8} & 45.8 & 57.8 & 50.2 & \underline{98.7} & 98.1\,{\scriptsize$\pm$0.3} & 97.0 & 78.2 \\
\midrule
\multicolumn{9}{l}{\textbf{\emph{Ours}}} \\
\hlours \textbf{PriorSplit} & \textbf{73.5}\,{\scriptsize$\pm$0.4} & \textbf{71.3} & \textbf{78.5} & \textbf{70.6} & 98.6 & 96.5\,{\scriptsize$\pm$2.1} & \textbf{97.4} & \textbf{83.0} \\
\midrule
\grayed{Oracle reference (Def. 1)} & \grayed{73.8} & \grayed{71.2} & \grayed{79.2} & \grayed{71.2} & \grayed{100.0} & \grayed{100.0} & \grayed{100.0} & \grayed{100.0} \\
\bottomrule
\end{tabular}
\end{table*}

\noindent\textbf{PriorSplit leads on short contexts.} With all six frames, $\IntTieN$ ranked methods of Table~\ref{tab:main} score within $\IntTieMargin$ points of OPSD. On the short side, PriorSplit reaches $\ShortPS\%$ (oracle reference $\ShortBayes\%$), $\GainShortOPSD$ points above OPSD and $\GainShortBest$ above the strongest compared method, prior distillation. Inference-time corrections stay near OPSD there, as contrasting views cannot supply an unstored order.

\noindent\textbf{PriorSplit keeps full-context accuracy high.} With all six frames, PriorSplit reaches $\IntPS\%$ (OPSD $\IntOPSD\%$) and $\ConfPS\%$ on conflict items (OPSD $\ConfOPSD\%$). On Long8 and Long32 it reaches $\LongEightPS\%$ and $\LongTTPS\%$ (OPSD $\LongEightOPSD\%$ and $\LongTTOPSD\%$). Shared-weight methods losing less conflict accuracy than PriorSplit gain at most $\MaxGainLessConf$ short-side points (Fig.~\ref{fig:tradeoff_full}).

\noindent\textbf{PriorSplit leads on four more backbones.} It gains $\BBGainMin$ to $\BBGainMax$ short-side points over OPSD and at least $\BBBestGainMin$ over the strongest compared method of each (Table~\ref{tab:general}). On InternVL3.5-8B its check misses $\BBInternMiss\%$ of visible questions, so conflict accuracy stays at OPSD's.

\noindent\textbf{Held-out skills need only new counts.} On three synthetic skills held out of training, PriorSplit takes its counts from a library of their episodes without retraining and reaches $\HeldoutPS\%$ with no frame in view against $\HeldoutOPSD\%$ for OPSD (Table~\ref{tab:s_heldout}), near the $\ThLibPred\%$ predicted under Theorem~\ref{thm:switch}(b).

\subsection{Missing and Conflicting Real Evidence}
\label{sec:real}
\begin{table}[t]
\centering
\footnotesize
\setlength{\tabcolsep}{3pt}
\caption{\textbf{Results on real frames.} PriorSplit gains short-side points over OPSD and the strongest compared method in every row. Qwen3-VL-4B and 8B readers, points averaged over seeds.}
\label{tab:real}
\begin{tabular}{llrrrr}
\toprule
 & & \multicolumn{2}{c}{Short-side gain over} & & \\
\cmidrule(lr){3-4}
Setting & Reader & OPSD & Strongest & Conflict & Miss \\
\midrule
\multicolumn{6}{l}{\textbf{\emph{Mind2Web}}} \\
Order & 4B & 13.3 & 10.4 & $-$1.7 & 3.6 \\
Order & 8B & 13.5 & 11.9 & $-$1.1 & 1.9 \\
\midrule
\multicolumn{6}{l}{\textbf{\emph{HoloAssist}}} \\
Order & 4B & 29.0 & 3.9 & $-$3.0 & 3.1 \\
Order & 8B & 29.7 & 4.4 & $-$4.0 & 2.2 \\
\midrule
\multicolumn{6}{l}{\textbf{\emph{CaptainCook4D}}} \\
Order & 4B & 16.3 & 5.5 & $-$5.2 & 7.7 \\
\bottomrule
\end{tabular}
\end{table}

\begin{figure}[t]
\centering
\includegraphics[width=\columnwidth]{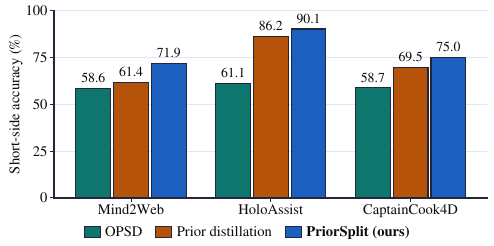}
\caption{\textbf{Short contexts on real frames.} PriorSplit is highest on every dataset. Short-side accuracy (\%), Qwen3-VL-4B.}
\label{fig:real_bars}
\end{figure}

Table~\ref{tab:real} covers real frames with the tag banner: Mind2Web, $20$ egocentric tasks of HoloAssist~\cite{holoassist} and recipes of CaptainCook4D~\cite{captaincook4d}. On HoloAssist and CaptainCook4D the shown order is the person's own, and Mind2Web uses the synthetic order noise. Evidence goes missing when a named step has no frame (Fig.~\ref{fig:qualitative} rows 1 and 2) and conflicts when the shown order is unusual (row 3, more in Fig.~\ref{fig:s_qualitative}).

\noindent\phantomsection\label{sec:m2w}\textbf{PriorSplit fills in missing frames.} On short contexts PriorSplit beats OPSD and the strongest compared method in every row of Table~\ref{tab:real} (Fig.~\ref{fig:real_bars}). With 4B readers it gains $\MGainShort$ $[\MGainShortlo, \MGainShorthi]$, $\HAOrderGainOPSD$ $[\HAOrderGainOPSDlo, \HAOrderGainOPSDhi]$ and $\CGainShort$ $[\CGainShortlo, \CGainShorthi]$ points over OPSD on Mind2Web, HoloAssist and CaptainCook4D. Over prior distillation the leads are $\MGainShortBest$ $[\MGainShortBestlo, \MGainShortBesthi]$, $\HAOrderGainBest$ $[\HAOrderGainBestlo, \HAOrderGainBesthi]$ and $\CGainShortBest$ $[\CGainShortBestlo, \CGainShortBesthi]$ ($95\%$ intervals, Table~\ref{tab:s_best}). On HoloAssist order with 4B readers, prior distillation trained on both question types for $\HANSSteps$ steps instead of $\HAOrderSteps$ stays $\HANSOrderGainPD$ points $[\HANSOrderGainPDlo, \HANSOrderGainPDhi]$ behind. With no frame in view only memory can answer. OPSD stays near chance there ($\ZeroFrameMOPSD\%$, $\ZeroFrameHAOPSD\%$ and $\ZeroFrameCOPSD\%$ on Mind2Web, HoloAssist and CaptainCook4D), prior distillation reaches $\ZeroFrameMPD\%$, $\ZeroFrameHAPD\%$ and $\ZeroFrameCPD\%$, and PriorSplit $\ZeroFrameMPS\%$, $\ZeroFrameHAPS\%$ and $\ZeroFrameCPS\%$. On next-step questions it leads by $\NGainShort$ points on Mind2Web and $\CNextGainBest$ $[\CNextGainBestlo, \CNextGainBesthi]$ on CaptainCook4D.

\begin{figure}[t]
\centering
\includegraphics[width=\columnwidth]{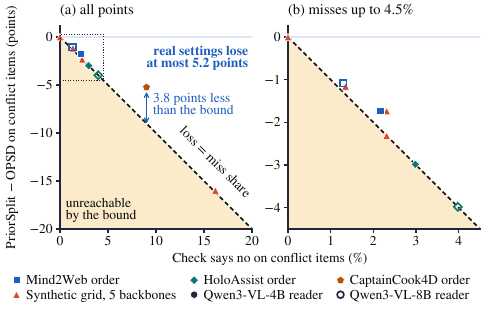}
\caption{\textbf{Check misses bound the conflict loss.} On every seed, setting and reader, PriorSplit loses at most the miss share (conflict items handed to the prior) against OPSD on conflict items. Seed means per setting and reader. Dashed: loss equal to the miss share.}
\label{fig:check_bound}
\end{figure}

\noindent\textbf{When the check sees the steps, the frames decide.} PriorSplit scores $\HAConflictAcc\%$ $[\HAConflictAcclo, \HAConflictAcchi]$ (4B) and $\HAEightConflictAcc\%$ (8B) on the $\HANConfOrder$ full-context HoloAssist order questions with an unusual shown order, against $\HAVOneConfOrder\%$ and $\HAEightVOneConfOrder\%$ for prior distillation and $\HAPriorConf\%$ for the prior alone. It departs from OPSD only on the conflict items its check hands to the prior, within their share of all conflict items (Theorem~\ref{thm:switch}(a), Fig.~\ref{fig:check_bound}): at most $\MaxConfLossRealMain$ points on tagged Mind2Web and $\MaxConfLossShownFlag$ on tagged HoloAssist and CaptainCook4D.

\noindent\textbf{The check routes close to the oracle switch.} On the tagged order questions of all three datasets with 4B readers, PriorSplit stays within $\OracleGapShort$ short-side points of the oracle switch, which knows the steps shown. On HoloAssist and CaptainCook4D with 4B readers, the probe (Section~\ref{sec:s_probe}) lifts accuracy on all visible conflicts to OPSD's, $\HeadHADeltaConf$ and $\HeadCCDeltaConf$ over the reader's check, and on CaptainCook4D gains $\HeadCCGainShort$ short-side points over OPSD.

\noindent\textbf{PriorSplit leads OPSD with either banners or frame content removed.} The banner names each step, so Table~\ref{tab:real} tests memory and routing rather than recognizing a step from its content. Masking frame content at test time leaves PriorSplit's Mind2Web order gain over OPSD at $\MaskMGainShort$ short-side points (Section~\ref{sec:s_content}). Without the tag banner on Mind2Web, with action texts as step names, PriorSplit leads the strongest compared method by $\TFOrdGainShort$ short-side points on order and $\TFNxtGainVOne$ on next-step questions (Table~\ref{tab:s_best}), and on order the probe fitted on training episodes reaches $\HeadTFShort\%$ against OPSD's $\TFOrdShortOPSD\%$ at OPSD's conflict accuracy. On held-out Mind2Web skills without the banner, PriorSplit with the probe leads OPSD by $\HOTGainShort$ $[\HOTGainShortlo, \HOTGainShorthi]$ short-side points on held-out trajectories of seen websites and $\HOSGainShort$ $[\HOSGainShortlo, \HOSGainShorthi]$ on held-out websites, and matches it with all frames (Section~\ref{sec:s_heldout_m2w}). Without the CaptainCook4D banner, frames lift OPSD from $\RawOPSDConfZero\%$ with no frame to $\RawOPSDConf\%$ on the $\RawNConf$ full-context departures from the usual order, and PriorSplit leads OPSD and prior distillation on the short side by $\RawGainShortOPSD$ $[\RawGainShortOPSDlo, \RawGainShortOPSDhi]$ and $\RawGainShortPD$ $[\RawGainShortPDlo, \RawGainShortPDhi]$ points (Section~\ref{sec:s_content}).

\subsection{Ablations and Analysis}
\label{sec:ablation}
\begin{table}[tb]
\centering
\footnotesize
\setlength{\tabcolsep}{3pt}
\caption{\textbf{Ablations.} PriorSplit (switch) leads every ranked variant on the short side. Accuracy (\%) with Qwen3-VL-4B. Reader ratio plus prior, checked: where the check says yes, log prior plus the reader's score with minus without frames. Italic: not ranked.}
\label{tab:ablation}
\begin{tabular}{lrrr}
\toprule
Variant & Short & Interval & Conflict \\
\midrule
\hlours \textbf{PriorSplit (switch)} & \textbf{73.5} & 98.6 & 96.5 \\
\grayed{Oracle switch} & \grayed{74.0} & \grayed{98.8} & \grayed{98.3} \\
Reader only (OPSD) & 51.1 & \textbf{98.8} & \textbf{98.3} \\
Prior only & 71.3 & 71.3 & 1.5 \\
Reader plus prior & 54.7 & \underline{98.7} & \underline{97.7} \\
Reader ratio plus prior, checked & \underline{73.4} & 98.3 & 94.8 \\
\midrule
\multicolumn{4}{l}{\textbf{\emph{Counts in the prompt}}} \\
\grayed{OPSD reader} & \grayed{68.4} & \grayed{97.2} & \grayed{90.9} \\
\grayed{Base model} & \grayed{70.0} & \grayed{72.5} & \grayed{9.1} \\
\bottomrule
\end{tabular}
\end{table}

In Table~\ref{tab:ablation}, PriorSplit's gap to the oracle switch is the check's cost in Theorem~\ref{thm:switch}(a): it misses $\GateMiss\%$ of visible questions and says yes on $\GateFalseYes\%$ of questions with a named step out of view. The oracle's conflict difference to OPSD vanishes on five synthetic backbones and tagged Mind2Web (Table~\ref{tab:s_gate}). With the base model as reader and check, PriorSplit reaches $\SwapBaseConf\%$ on conflict items (Table~\ref{tab:s_ablation}).

\noindent\textbf{The counts extend to three steps, drift and trajectories.} Ranking steps by pair majorities, the same counts and readers answer three-step synthetic questions, where PriorSplit leads OPSD by $\TriFirstGain$ points on the first step and $\TriOrderGain$ on their full order with none or half of six frames shown. Decayed counts restore a $\DriftGain$-point short-side lead over OPSD $\DriftT$ episodes after a synthetic drift. On the short side PriorSplit is right on all asked next-step questions of $\TrajPS\%$ of Mind2Web trajectories, OPSD $\TrajOPSD\%$ (Section~\ref{sec:s_beyond}). On ALFWorld~\cite{alfworld}, with its text rendered as frames, PriorSplit answers next-step questions $\AlfItemGainOPSD$ points above OPSD and $\AlfItemGainPD$ above prior distillation. As an agent's step hint it reaches $\AlfSuccPS\%$ task success against OPSD's $\AlfSuccOPSD\%$ while calling the reader on $\AlfReaderShare\%$ of steps, where OPSD calls it on every step (Section~\ref{sec:s_alfworld}).

\section{Conclusion}
\label{sec:conclusion}

An internalized skill must answer when frames are missing or contradict its usual order. On synthetic skills with Qwen3-VL-4B, OPSD stores almost no order, and every tested subsequence variant stores it in the same weights and loses conflict accuracy. \oursc{PriorSplit} \priorc{counts} privileged orders outside the reader and \readc{reads} the frames when its check finds the steps. With frames missing, it \keyhl{leads every compared method at OPSD's training budget} on synthetic skills, seen web and cooking tasks and seen egocentric order questions, and it loses conflict accuracy only on check misses.%
\label{body-end}\par

\section*{Acknowledgements}
YL was supported by a five-year PGTA scholarship from the School of Computing, Newcastle University. This research was supported by the NVIDIA Academic Grant Program.

\section*{LLM Usage Statement}
LLM tools were used only to polish author-written text, related work, code, experimental analysis, and figures. They were not used for data creation or annotation, or for developing research ideas, claims, or conclusions. All outputs were reviewed and verified by the authors.
The authors take full responsibility for all experimental results and for the entire content of this paper.

{
    \small
    \bibliographystyle{ieeenat_fullname}
    \bibliography{refs}

@inproceedings{zhao2026opsd,
  author       = {Siyan Zhao and
                  Zhihui Xie and
                  Mengchen Liu and
                  Jing Huang and
                  Guan Pang and
                  Feiyu Chen and
                  Aditya Grover},
  title        = {{Self-Distilled Reasoner}: On-Policy Self-Distillation for Large Language Models},
  booktitle    = {Forty-third International Conference on Machine Learning, ICML 2026, Seoul, South Korea, July 6-11, 2026},
  series       = {Proceedings of Machine Learning Research},
  volume       = {306},
  pages        = {162433--162448},
  publisher    = {PMLR},
  year         = {2026}
}

@inproceedings{skill0,
  author       = {Zhengxi Lu and
                  Zhiyuan Yao and
                  Jinyang Wu and
                  Chengcheng Han and
                  Xierui Song and
                  Xunliang Cai and
                  Weiming Lu and
                  Jun Xiao and
                  Yueting Zhuang and
                  Yongliang Shen},
  title        = {{SKILL0}: In-Context Agentic Reinforcement Learning for Skill Internalization},
  booktitle    = {Findings of the Association for Computational Linguistics: {EMNLP} 2026, Budapest, Hungary},
  pages        = {4485--4505},
  publisher    = {Association for Computational Linguistics},
  year         = {2026},
  doi          = {10.18653/2026.findings-emnlp.215}
}

@article{snell2022context,
  author       = {Charlie Snell and
                  Dan Klein and
                  Ruiqi Zhong},
  title        = {Learning by Distilling Context},
  journal      = {CoRR},
  volume       = {abs/2209.15189},
  year         = {2022},
  doi          = {10.48550/ARXIV.2209.15189},
  eprinttype   = {arXiv},
  eprint       = {2209.15189}
}

@article{opcd,
  author       = {Tianzhu Ye and
                  Li Dong and
                  Xun Wu and
                  Shaohan Huang and
                  Furu Wei},
  title        = {On-Policy Context Distillation for Language Models},
  journal      = {CoRR},
  volume       = {abs/2602.12275},
  year         = {2026},
  doi          = {10.48550/ARXIV.2602.12275},
  eprinttype   = {arXiv},
  eprint       = {2602.12275}
}

@article{nca,
  author       = {Xun Wang and
                  Ruishuo Chen and
                  Zhuoran Li and
                  Yu Chen and
                  Longbo Huang},
  title        = {When Context Returns: Toward Robust Internalization in On-Policy Distillation},
  journal      = {CoRR},
  volume       = {abs/2606.11627},
  year         = {2026},
  doi          = {10.48550/ARXIV.2606.11627},
  eprinttype   = {arXiv},
  eprint       = {2606.11627}
}

@inproceedings{shi2024cad,
  author       = {Weijia Shi and
                  Xiaochuang Han and
                  Mike Lewis and
                  Yulia Tsvetkov and
                  Luke Zettlemoyer and
                  Wen{-}tau Yih},
  editor       = {Kevin Duh and
                  Helena G{\'o}mez-Adorno and
                  Steven Bethard},
  title        = {Trusting Your Evidence: Hallucinate Less with Context-aware Decoding},
  booktitle    = {Proceedings of the 2024 Conference of the North American Chapter of the Association for Computational Linguistics: Human Language Technologies, NAACL 2024, Volume 2: Short Papers, Mexico City, Mexico, June 16-21, 2024},
  pages        = {783--791},
  publisher    = {Association for Computational Linguistics},
  year         = {2024},
  doi          = {10.18653/V1/2024.NAACL-SHORT.69}
}

@inproceedings{zhao2021calibrate,
  author       = {Zihao Zhao and
                  Eric Wallace and
                  Shi Feng and
                  Dan Klein and
                  Sameer Singh},
  editor       = {Marina Meila and
                  Tong Zhang},
  title        = {Calibrate Before Use: Improving Few-shot Performance of Language Models},
  booktitle    = {The 38th International Conference on Machine Learning, ICML 2021, Virtual Event, July 18-24, 2021},
  series       = {Proceedings of Machine Learning Research},
  volume       = {139},
  pages        = {12697--12706},
  publisher    = {PMLR},
  year         = {2021}
}

@article{saerens2002adjusting,
  author       = {Marco Saerens and
                  Patrice Latinne and
                  Christine Decaestecker},
  title        = {Adjusting the Outputs of a Classifier to New a Priori Probabilities: A Simple Procedure},
  journal      = {Neural Computation},
  volume       = {14},
  number       = {1},
  pages        = {21--41},
  year         = {2002},
  doi          = {10.1162/089976602753284446}
}

@inproceedings{menon2021logit,
  author       = {Aditya Krishna Menon and
                  Sadeep Jayasumana and
                  Ankit Singh Rawat and
                  Himanshu Jain and
                  Andreas Veit and
                  Sanjiv Kumar},
  title        = {Long-tail learning via logit adjustment},
  booktitle    = {The Ninth International Conference on Learning Representations, {ICLR} 2021, Virtual Event, May 3-7, 2021},
  publisher    = {OpenReview.net},
  year         = {2021}
}

@article{cwopd,
  author       = {Yuanhao Sun and
                  Huawei Ji and
                  Jiaxin Ding and
                  Luoyi Fu and
                  Xinbing Wang},
  title        = {Distill the Visual Evidence, Not Just the Answer: Cross-World On-Policy Distillation for Vision-Language Models},
  journal      = {CoRR},
  volume       = {abs/2609.38777},
  year         = {2026},
  doi          = {10.48550/ARXIV.2609.38777},
  eprinttype   = {arXiv},
  eprint       = {2609.38777v1},
  note         = {Version 1, withdrawn on 2026-10-19}
}

@inproceedings{alfworld,
  author       = {Mohit Shridhar and
                  Xingdi Yuan and
                  Marc{-}Alexandre C{\^{o}}t{\'{e}} and
                  Yonatan Bisk and
                  Adam Trischler and
                  Matthew Hausknecht},
  title        = {{ALFWorld}: Aligning Text and Embodied Environments for Interactive Learning},
  booktitle    = {The Ninth International Conference on Learning Representations, {ICLR} 2021, Virtual Event, May 3-7, 2021},
  publisher    = {OpenReview.net},
  year         = {2021}
}

@inproceedings{contractskill2026,
  author       = {Zijian Lu and
                  Yiping Zuo and
                  Yupeng Nie and
                  Xin He and
                  Weibei Fan and
                  Lianyong Qi and
                  Shi Jin},
  title        = {{ContractSkill}: Repairable Contract-Based Skills for Multimodal Web Agents},
  booktitle    = {Proceedings of the 34th ACM International Conference on Multimedia, MM 2026, Rio de Janeiro, Brazil, November 10-14, 2026},
  publisher    = {ACM},
  year         = {2026}
}

@inproceedings{skill052026,
  author       = {Jiapeng Zhu and
                  Jianxiang Yu and
                  Yibo Zhao and
                  Chengcheng Han and
                  Xierui Song and
                  Xunliang Cai and
                  Xiang Li and
                  Weining Qian},
  title        = {{Skill0.5}: Joint Skill Internalization and Utilization for Out-of-Distribution Generalization in Agentic Reinforcement Learning},
  booktitle    = {Findings of the Association for Computational Linguistics: {EMNLP} 2026, Budapest, Hungary},
  pages        = {19381--19396},
  publisher    = {Association for Computational Linguistics},
  year         = {2026},
  doi          = {10.18653/2026.findings-emnlp.917}
}

@inproceedings{skillsd2026,
  author       = {Hao Wang and
                  Guozhi Wang and
                  Han Xiao and
                  Yufeng Zhou and
                  Yue Pan and
                  Jichao Wang and
                  Ke Xu and
                  Yafei Wen and
                  Xiaohu Ruan and
                  Xiaoxin Chen and
                  Honggang Qi},
  title        = {{Skill-SD}: Skill-Conditioned Self-Distillation for Multi-turn {LLM} Agents},
  booktitle    = {Advances in Neural Information Processing Systems 39: Annual Conference on Neural Information Processing Systems 2026, NeurIPS 2026},
  year         = {2026}
}

@inproceedings{continualsd2026,
  author       = {Idan Shenfeld and
                  Mehul Damani and
                  Jonas H{\"u}botter and
                  Pulkit Agrawal},
  title        = {Self-Distillation Enables Continual Learning},
  booktitle    = {Forty-third International Conference on Machine Learning, ICML 2026, Seoul, South Korea, July 6-11, 2026},
  series       = {Proceedings of Machine Learning Research},
  volume       = {306},
  pages        = {111542--111558},
  publisher    = {PMLR},
  year         = {2026}
}

@inproceedings{antisd2026,
  author       = {Guobin Shen and
                  Xiang Cheng and
                  Chenxiao Zhao and
                  Lei Huang and
                  Jindong Li and
                  Dongcheng Zhao and
                  XingYu Li},
  title        = {{Anti-Self-Distillation} for Reasoning {RL} via Pointwise Mutual Information},
  booktitle    = {Advances in Neural Information Processing Systems 39: Annual Conference on Neural Information Processing Systems 2026, NeurIPS 2026},
  year         = {2026}
}

@inproceedings{xia2026skillrl,
  author       = {Peng Xia and
                  Jianwen Chen and
                  Hanyang Wang and
                  Jiaqi Liu and
                  Kaide Zeng and
                  Yu Wang and
                  Siwei Han and
                  Yiyang Zhou and
                  Xujiang Zhao and
                  Haifeng Chen and
                  Zeyu Zheng and
                  Cihang Xie and
                  Huaxiu Yao},
  title        = {{SkillRL}: Evolving Agents via Recursive Skill-Augmented Reinforcement Learning},
  booktitle    = {Advances in Neural Information Processing Systems 39: Annual Conference on Neural Information Processing Systems 2026, NeurIPS 2026},
  year         = {2026}
}

@article{rethinking2026,
  author       = {Jingwen Chen and
                  Wenkai Yang and
                  Shengda Fan and
                  Wenbo Nie and
                  Chenxing Sun and
                  Shaodong Zheng and
                  Yangen Hu and
                  Lu Pan and
                  Ke Zeng and
                  Yankai Lin},
  title        = {Rethinking Continual Experience Internalization for Self-Evolving {LLM} Agents},
  journal      = {CoRR},
  volume       = {abs/2606.04703},
  year         = {2026},
  doi          = {10.48550/ARXIV.2606.04703},
  eprinttype   = {arXiv},
  eprint       = {2606.04703}
}

@inproceedings{agarwal2024onpolicy,
  author       = {Rishabh Agarwal and
                  Nino Vieillard and
                  Yongchao Zhou and
                  Piotr Stanczyk and
                  Sabela {Ramos Garea} and
                  Matthieu Geist and
                  Olivier Bachem},
  title        = {On-Policy Distillation of Language Models: Learning from Self-Generated Mistakes},
  booktitle    = {The Twelfth International Conference on Learning Representations, ICLR 2024, Vienna, Austria, May 7-11, 2024},
  publisher    = {OpenReview.net},
  year         = {2024}
}

@inproceedings{hu2022lora,
  author       = {Edward Hu and
                  Yelong Shen and
                  Phillip Wallis and
                  Zeyuan Allen-Zhu and
                  Yuanzhi Li and
                  Shean Wang and
                  Lu Wang and
                  Weizhu Chen},
  title        = {{LoRA}: Low-Rank Adaptation of Large Language Models},
  booktitle    = {The Tenth International Conference on Learning Representations, ICLR 2022, Virtual Event, April 25-29, 2022},
  publisher    = {OpenReview.net},
  year         = {2022}
}

@article{qwen3vl,
  author       = {Shuai Bai and
                  Yuxuan Cai and
                  Ruizhe Chen and
                  Keqin Chen and
                  Xionghui Chen and
                  Zesen Cheng and
                  Lianghao Deng and
                  Wei Ding and
                  Chang Gao and
                  Chunjiang Ge and
                  Wenbin Ge and
                  Zhifang Guo and
                  Qidong Huang and
                  Jie Huang and
                  Fei Huang and
                  Binyuan Hui and
                  Shutong Jiang and
                  Zhaohai Li and
                  Mingsheng Li and
                  Mei Li and
                  Kaixin Li and
                  Zicheng Lin and
                  Junyang Lin and
                  Xuejing Liu and
                  Jiawei Liu and
                  Chenglong Liu and
                  Yang Liu and
                  Dayiheng Liu and
                  Shixuan Liu and
                  Dunjie Lu and
                  Ruilin Luo and
                  Chenxu Lv and
                  Rui Men and
                  Lingchen Meng and
                  Xuancheng Ren and
                  Xingzhang Ren and
                  Sibo Song and
                  Yuchong Sun and
                  Jun Tang and
                  Jianhong Tu and
                  Jianqiang Wan and
                  Peng Wang and
                  Pengfei Wang and
                  Qiuyue Wang and
                  Yuxuan Wang and
                  Tianbao Xie and
                  Yiheng Xu and
                  Haiyang Xu and
                  Jin Xu and
                  Zhibo Yang and
                  Mingkun Yang and
                  Jianxin Yang and
                  An Yang and
                  Bowen Yu and
                  Fei Zhang and
                  Hang Zhang and
                  Xi Zhang and
                  Bo Zheng and
                  Humen Zhong and
                  Jingren Zhou and
                  Fan Zhou and
                  Jing Zhou and
                  Yuanzhi Zhu and
                  Ke Zhu},
  title        = {{Qwen3-VL} Technical Report},
  journal      = {CoRR},
  volume       = {abs/2511.21631},
  year         = {2025},
  doi          = {10.48550/ARXIV.2511.21631},
  eprinttype   = {arXiv},
  eprint       = {2511.21631}
}

@article{qwen25vl,
  author       = {Shuai Bai and
                  Keqin Chen and
                  Xuejing Liu and
                  Jialin Wang and
                  Wenbin Ge and
                  Sibo Song and
                  Kai Dang and
                  Peng Wang and
                  Shijie Wang and
                  Jun Tang and
                  Humen Zhong and
                  Yuanzhi Zhu and
                  Mingkun Yang and
                  Zhaohai Li and
                  Jianqiang Wan and
                  Pengfei Wang and
                  Wei Ding and
                  Zheren Fu and
                  Yiheng Xu and
                  Jiabo Ye and
                  Xi Zhang and
                  Tianbao Xie and
                  Zesen Cheng and
                  Hang Zhang and
                  Zhibo Yang and
                  Haiyang Xu and
                  Junyang Lin},
  title        = {{Qwen2.5-VL} Technical Report},
  journal      = {CoRR},
  volume       = {abs/2502.13923},
  year         = {2025},
  doi          = {10.48550/ARXIV.2502.13923},
  eprinttype   = {arXiv},
  eprint       = {2502.13923}
}

@inproceedings{deng2023mind2web,
  author       = {Xiang Deng and
                  Yu Gu and
                  Boyuan Zheng and
                  Shijie Chen and
                  Sam Stevens and
                  Boshi Wang and
                  Huan Sun and
                  Yu Su},
  editor       = {Alice Oh and
                  Tristan Naumann and
                  Amir Globerson and
                  Kate Saenko and
                  Moritz Hardt and
                  Sergey Levine},
  title        = {{Mind2Web}: Towards a Generalist Agent for the Web},
  booktitle    = {Advances in Neural Information Processing Systems 36: Annual Conference on Neural Information Processing Systems 2023, NeurIPS 2023, New Orleans, LA, USA, December 10-16, 2023},
  pages        = {28091--28114},
  year         = {2023}
}

@inproceedings{zheng2024seeact,
  author       = {Boyuan Zheng and
                  Boyu Gou and
                  Jihyung Kil and
                  Huan Sun and
                  Yu Su},
  editor       = {Ruslan Salakhutdinov and
                  Zico Kolter and
                  Katherine A. Heller and
                  Adrian Weller and
                  Nuria Oliver and
                  Jonathan Scarlett and
                  Felix Berkenkamp},
  title        = {{GPT-4V(ision)} is a Generalist Web Agent, if Grounded},
  booktitle    = {Forty-first International Conference on Machine Learning, ICML 2024, Vienna, Austria, July 21-27, 2024},
  series       = {Proceedings of Machine Learning Research},
  volume       = {235},
  pages        = {61349--61385},
  publisher    = {PMLR},
  year         = {2024}
}

@inproceedings{liu2021dexperts,
  author       = {Alisa Liu and
                  Maarten Sap and
                  Ximing Lu and
                  Swabha Swayamdipta and
                  Chandra Bhagavatula and
                  Noah A. Smith and
                  Yejin Choi},
  editor       = {Chengqing Zong and
                  Fei Xia and
                  Wenjie Li and
                  Roberto Navigli},
  title        = {{DExperts}: Decoding-Time Controlled Text Generation with Experts and Anti-Experts},
  booktitle    = {Proceedings of the 59th Annual Meeting of the Association for Computational Linguistics and the 11th International Joint Conference on Natural Language Processing, ACL/IJCNLP 2021, Volume 1: Long Papers, Virtual Event, August 1-6, 2021},
  pages        = {6691--6706},
  publisher    = {Association for Computational Linguistics},
  year         = {2021},
  doi          = {10.18653/V1/2021.ACL-LONG.522}
}

@inproceedings{clark2019easy,
  author       = {Christopher Clark and
                  Mark Yatskar and
                  Luke Zettlemoyer},
  editor       = {Kentaro Inui and
                  Jing Jiang and
                  Vincent Ng and
                  Xiaojun Wan},
  title        = {Don't Take the Easy Way Out: Ensemble Based Methods for Avoiding Known Dataset Biases},
  booktitle    = {Proceedings of the 2019 Conference on Empirical Methods in Natural Language Processing and the 9th International Joint Conference on Natural Language Processing, EMNLP-IJCNLP 2019, Hong Kong, China, November 3-7, 2019},
  pages        = {4069--4082},
  publisher    = {Association for Computational Linguistics},
  year         = {2019},
  doi          = {10.18653/V1/D19-1418}
}

@inproceedings{niu2021cfvqa,
  author       = {Yulei Niu and
                  Kaihua Tang and
                  Hanwang Zhang and
                  Zhiwu Lu and
                  Xian-Sheng Hua and
                  Ji-Rong Wen},
  title        = {Counterfactual {VQA}: A Cause-Effect Look at Language Bias},
  booktitle    = {{IEEE/CVF} Conference on Computer Vision and Pattern Recognition, {CVPR} 2021, Virtual Event, June 19-25, 2021},
  pages        = {12700--12710},
  publisher    = {Computer Vision Foundation / IEEE},
  year         = {2021},
  doi          = {10.1109/CVPR46437.2021.01251}
}

@inproceedings{cadene2019rubi,
  author       = {Remi Cadene and
                  Corentin Dancette and
                  Hedi {Ben younes} and
                  Matthieu Cord and
                  Devi Parikh},
  editor       = {Hanna M. Wallach and
                  Hugo Larochelle and
                  Alina Beygelzimer and
                  Florence d'Alch{\'e}-Buc and
                  Emily B. Fox and
                  Roman Garnett},
  title        = {{RUBi}: Reducing Unimodal Biases for Visual Question Answering},
  booktitle    = {Advances in Neural Information Processing Systems 32: Annual Conference on Neural Information Processing Systems 2019, NeurIPS 2019, Vancouver, BC, Canada, December 8-14, 2019},
  pages        = {841--852},
  year         = {2019}
}

@inproceedings{wang2025adacad,
  author       = {Han Wang and
                  Archiki Prasad and
                  Elias Stengel-Eskin and
                  Mohit Bansal},
  editor       = {Luis Chiruzzo and
                  Alan Ritter and
                  Lu Wang},
  title        = {{AdaCAD}: Adaptively Decoding to Balance Conflicts between Contextual and Parametric Knowledge},
  booktitle    = {Proceedings of the 2025 Conference of the Nations of the Americas Chapter of the Association for Computational Linguistics: Human Language Technologies, NAACL 2025, Volume 1: Long Papers, Albuquerque, NM, USA, April 29 - May 4, 2025},
  pages        = {11636--11652},
  publisher    = {Association for Computational Linguistics},
  year         = {2025},
  doi          = {10.18653/V1/2025.NAACL-LONG.581}
}

@inproceedings{xie2024chameleon,
  author       = {Jian Xie and
                  Kai Zhang and
                  Jiangjie Chen and
                  Renze Lou and
                  Yu Su},
  title        = {Adaptive Chameleon or Stubborn Sloth: Revealing the Behavior of Large Language Models in Knowledge Conflicts},
  booktitle    = {The Twelfth International Conference on Learning Representations, ICLR 2024, Vienna, Austria, May 7-11, 2024},
  publisher    = {OpenReview.net},
  year         = {2024}
}

@inproceedings{zhou2023contextfaithful,
  author       = {Wenxuan Zhou and
                  Sheng Zhang and
                  Hoifung Poon and
                  Muhao Chen},
  editor       = {Houda Bouamor and
                  Juan Pino and
                  Kalika Bali},
  title        = {Context-faithful Prompting for Large Language Models},
  booktitle    = {Findings of the Association for Computational Linguistics: EMNLP 2023, Singapore, December 6-10, 2023},
  series       = {Findings of ACL},
  volume       = {EMNLP 2023},
  pages        = {14544--14556},
  publisher    = {Association for Computational Linguistics},
  year         = {2023},
  doi          = {10.18653/V1/2023.FINDINGS-EMNLP.968}
}

@article{hinton2015distilling,
  author       = {Geoffrey Hinton and
                  Oriol Vinyals and
                  Jeff Dean},
  title        = {Distilling the Knowledge in a Neural Network},
  journal      = {CoRR},
  volume       = {abs/1503.02531},
  year         = {2015},
  doi          = {10.48550/ARXIV.1503.02531},
  eprinttype   = {arXiv},
  eprint       = {1503.02531}
}

@inproceedings{holtzman2021surface,
  author       = {Ari Holtzman and
                  Peter West and
                  Vered Shwartz and
                  Yejin Choi and
                  Luke Zettlemoyer},
  editor       = {Marie{-}Francine Moens and
                  Xuanjing Huang and
                  Lucia Specia and
                  Scott Wen{-}tau Yih},
  title        = {Surface Form Competition: Why the Highest Probability Answer Isn't Always Right},
  booktitle    = {Proceedings of the 2021 Conference on Empirical Methods in Natural Language Processing, EMNLP 2021, Virtual Event / Punta Cana, Dominican Republic, 7-11 November, 2021},
  pages        = {7038--7051},
  publisher    = {Association for Computational Linguistics},
  year         = {2021},
  doi          = {10.18653/V1/2021.EMNLP-MAIN.564}
}

@article{pad2024,
  author       = {Raymond Douglas and
                  Andis Draguns and
                  Tom{\'a}{\v s} Gaven{\v c}iak},
  title        = {Mitigating the Influence of Distractor Tasks in {LMs} with Prior-Aware Decoding},
  journal      = {CoRR},
  volume       = {abs/2401.17692},
  year         = {2024},
  doi          = {10.48550/ARXIV.2401.17692},
  eprinttype   = {arXiv},
  eprint       = {2401.17692}
}

@article{opid2026,
  author       = {Shuo Yang and
                  Jinyang Wu and
                  Zhengxi Lu and
                  Yuhao Shen and
                  Fan Zhang and
                  Lang Feng and
                  Shuai Zhang and
                  Haoran Luo and
                  Zheng Lian and
                  Zhengqi Wen and
                  Jianhua Tao},
  title        = {{OPID}: On-Policy Skill Distillation for Agentic Reinforcement Learning},
  journal      = {CoRR},
  volume       = {abs/2606.26790},
  year         = {2026},
  doi          = {10.48550/ARXIV.2606.26790},
  eprinttype   = {arXiv},
  eprint       = {2606.26790}
}

@article{seed2026,
  author       = {Jinyang Wu and
                  Shuo Yang and
                  Zhengxi Lu and
                  Fan Zhang and
                  Yuhao Shen and
                  Lang Feng and
                  Haoran Luo and
                  Zheng Lian and
                  Shuai Zhang and
                  Zhengqi Wen and
                  Jianhua Tao},
  title        = {{SEED}: Self-Evolving On-Policy Distillation for Agentic Reinforcement Learning},
  journal      = {CoRR},
  volume       = {abs/2607.14777},
  year         = {2026},
  doi          = {10.48550/ARXIV.2607.14777},
  eprinttype   = {arXiv},
  eprint       = {2607.14777}
}

@inproceedings{proxytuning2024,
  author       = {Alisa Liu and
                  Xiaochuang Han and
                  Yizhong Wang and
                  Yulia Tsvetkov and
                  Yejin Choi and
                  Noah A. Smith},
  title        = {Tuning Language Models by Proxy},
  booktitle    = {First Conference on Language Modeling, {COLM} 2024},
  year         = {2024}
}

@article{vapnik2009lupi,
  author       = {Vladimir Vapnik and
                  Akshay Vashist},
  title        = {A new learning paradigm: Learning using privileged information},
  journal      = {Neural Networks},
  volume       = {22},
  number       = {5--6},
  pages        = {544--557},
  year         = {2009},
  doi          = {10.1016/j.neunet.2009.06.042}
}

@article{idefics3,
  author       = {Hugo Lauren{\c{c}}on and
                  Andr{\'{e}}s Marafioti and
                  Victor Sanh and
                  L{\'{e}}o Tronchon},
  title        = {Building and better understanding vision-language models: insights
                  and future directions},
  journal      = {CoRR},
  volume       = {abs/2408.12637},
  year         = {2024},
  doi          = {10.48550/ARXIV.2408.12637},
  eprinttype   = {arXiv},
  eprint       = {2408.12637}
}

@article{gates2026,
  author       = {Alex Stein and
                  Furong Huang and
                  Tom Goldstein},
  title        = {{GATES}: Self-Distillation under Privileged Context with Consensus Gating},
  journal      = {CoRR},
  volume       = {abs/2602.20574},
  year         = {2026},
  doi          = {10.48550/ARXIV.2602.20574},
  eprinttype   = {arXiv},
  eprint       = {2602.20574}
}

@article{ucob2026,
  author       = {Songjun Tu and
                  Chengdong Xu and
                  Qichao Zhang and
                  Yiwen Ma and
                  Yaocheng Zhang and
                  Linjing Li and
                  Dong Li and
                  Xiangyuan Lan and
                  Dongbin Zhao},
  title        = {{UCOB}: Learning to Utilize and Evolve Agentic Skills via Credit-Aware On-Policy Bidirectional Self-Distillation},
  journal      = {CoRR},
  volume       = {abs/2606.29502},
  year         = {2026},
  doi          = {10.48550/ARXIV.2606.29502},
  eprinttype   = {arXiv},
  eprint       = {2606.29502}
}

@article{latentskill2026,
  author       = {Aofan Yu and
                  Chenyu Zhou and
                  Tianyi Xu and
                  Zihan Guo and
                  Rong Shan and
                  Zhihui Fu and
                  Jun Wang and
                  Weiwen Liu and
                  Yong Yu and
                  Weinan Zhang and
                  Jianghao Lin},
  title        = {{LatentSkill}: From In-Context Textual Skills to In-Weight Latent Skills for {LLM} Agents},
  journal      = {CoRR},
  volume       = {abs/2606.06087},
  year         = {2026},
  doi          = {10.48550/ARXIV.2606.06087},
  eprinttype   = {arXiv},
  eprint       = {2606.06087}
}

@inproceedings{captaincook4d,
  author       = {Rohith Peddi and
                  Shivvrat Arya and
                  Bharath Challa and
                  Likhitha Pallapothula and
                  Akshay Vyas and
                  Bhavya Gouripeddi and
                  Qifan Zhang and
                  Jikai Wang and
                  Vasundhara Komaragiri and
                  Eric Ragan and
                  Nicholas Ruozzi and
                  Yu Xiang and
                  Vibhav Gogate},
  title        = {{CaptainCook4D}: A Dataset for Understanding Errors in Procedural Activities},
  booktitle    = {Advances in Neural Information Processing Systems 37: Annual Conference on Neural Information Processing Systems 2024, NeurIPS 2024, Vancouver, BC, Canada, December 10-15, 2024},
  pages        = {135626--135679},
  year         = {2024},
  eprinttype   = {arXiv},
  eprint       = {2312.14556}
}

@inproceedings{seminara2024tgml,
  author       = {Luigi Seminara and
                  Giovanni Maria Farinella and
                  Antonino Furnari},
  title        = {Differentiable Task Graph Learning: Procedural Activity Representation and Online Mistake Detection from Egocentric Videos},
  booktitle    = {Advances in Neural Information Processing Systems 37: Annual Conference on Neural Information Processing Systems 2024, NeurIPS 2024, Vancouver, BC, Canada, December 10-15, 2024},
  pages        = {59373--59407},
  year         = {2024},
  eprinttype   = {arXiv},
  eprint       = {2406.01486}
}

@article{internvl35,
  author       = {Weiyun Wang and
                  Zhangwei Gao and
                  Lixin Gu and
                  Hengjun Pu and
                  Long Cui and
                  Xingguang Wei and
                  Zhaoyang Liu and
                  Linglin Jing and
                  Shenglong Ye and
                  Jie Shao and
                  Zhaokai Wang and
                  Zhe Chen and
                  Hongjie Zhang and
                  Ganlin Yang and
                  Haomin Wang and
                  Qi Wei and
                  Jinhui Yin and
                  Wenhao Li and
                  Erfei Cui and
                  Guanzhou Chen and
                  Zichen Ding and
                  Changyao Tian and
                  Zhenyu Wu and
                  Jingjing Xie and
                  Zehao Li and
                  Bowen Yang and
                  Yuchen Duan and
                  Xuehui Wang and
                  Zhi Hou and
                  Haoran Hao and
                  Tianyi Zhang and
                  Songze Li and
                  Xiangyu Zhao and
                  Haodong Duan and
                  Nianchen Deng and
                  Bin Fu and
                  Yinan He and
                  Yi Wang and
                  Conghui He and
                  Botian Shi and
                  Junjun He and
                  Yingtong Xiong and
                  Han Lv and
                  Lijun Wu and
                  Wenqi Shao and
                  Kaipeng Zhang and
                  Huipeng Deng and
                  Biqing Qi and
                  Jiaye Ge and
                  Qipeng Guo and
                  Wenwei Zhang and
                  Songyang Zhang and
                  Maosong Cao and
                  Junyao Lin and
                  Kexian Tang and
                  Jianfei Gao and
                  Haian Huang and
                  Yuzhe Gu and
                  Chengqi Lyu and
                  Huanze Tang and
                  Rui Wang and
                  Haijun Lv and
                  Wanli Ouyang and
                  Limin Wang and
                  Min Dou and
                  Xizhou Zhu and
                  Tong Lu and
                  Dahua Lin and
                  Jifeng Dai and
                  Weijie Su and
                  Bowen Zhou and
                  Kai Chen and
                  Yu Qiao and
                  Wenhai Wang and
                  Gen Luo},
  title        = {{InternVL3.5}: Advancing Open-Source Multimodal Models in Versatility, Reasoning, and Efficiency},
  journal      = {CoRR},
  volume       = {abs/2508.18265},
  year         = {2025},
  doi          = {10.48550/ARXIV.2508.18265},
  eprinttype   = {arXiv},
  eprint       = {2508.18265}
}

@inproceedings{holoassist,
  author       = {Xin Wang and
                  Taein Kwon and
                  Mahdi Rad and
                  Bowen Pan and
                  Ishani Chakraborty and
                  Sean Andrist and
                  Dan Bohus and
                  Ashley Feniello and
                  Bugra Tekin and
                  Felipe Vieira Frujeri and
                  Neel Joshi and
                  Marc Pollefeys},
  title        = {{HoloAssist}: an Egocentric Human Interaction Dataset for Interactive {AI} Assistants in the Real World},
  booktitle    = {{IEEE/CVF} International Conference on Computer Vision, {ICCV} 2023, Paris, France, October 1-6, 2023},
  pages        = {20213--20224},
  publisher    = {{IEEE}},
  year         = {2023},
  doi          = {10.1109/ICCV51070.2023.01854}
}

@inproceedings{geifman2017selective,
  author       = {Yonatan Geifman and
                  Ran El{-}Yaniv},
  editor       = {Isabelle Guyon and
                  Ulrike von Luxburg and
                  Samy Bengio and
                  Hanna M. Wallach and
                  Rob Fergus and
                  S. V. N. Vishwanathan and
                  Roman Garnett},
  title        = {Selective Classification for Deep Neural Networks},
  booktitle    = {Advances in Neural Information Processing Systems 30: Annual Conference on Neural Information Processing Systems 2017, NIPS 2017, Long Beach, CA, USA, December 4-9, 2017},
  pages        = {4878--4887},
  year         = {2017}
}

@article{atlasva2026,
  author       = {Pan Wang and
                  Yihao Hu and
                  Xiujin Liu and
                  Jingchu Yang and
                  Hang Wang and
                  Zhihao Wen},
  title        = {{AtlasVA}: Self-Evolving Visual Skill Memory for Teacher-Free {VLM} Agents},
  journal      = {CoRR},
  volume       = {abs/2605.17933},
  year         = {2026},
  doi          = {10.48550/ARXIV.2605.17933},
  eprinttype   = {arXiv},
  eprint       = {2605.17933}
}

@inproceedings{lietzow2026vision,
  author       = {Niclas Lietzow and
                  Danielle Bitterman and
                  Carsten Eickhoff and
                  William Rudman and
                  Michal Golovanevsky},
  title        = {Vision-Default, Prior-Override: Causal Mechanisms of Perception-Knowledge Conflict in Vision-Language Models},
  booktitle    = {Findings of the Association for Computational Linguistics: {EMNLP} 2026, Budapest, Hungary},
  pages        = {43282--43298},
  publisher    = {Association for Computational Linguistics},
  year         = {2026}
}

@inproceedings{egoprocevqa2026,
  author       = {Junlong Li and
                  Junxi Li and
                  Yuxiang Yang and
                  Wenbin Zou and
                  Lap{-}Pui Chau and
                  Yi Wang},
  title        = {{EgoProceVQA}: A Novel Egocentric Procedural Understanding Task with Self-Skill-Exploration Agent},
  booktitle    = {Proceedings of the 34th ACM International Conference on Multimedia, MM 2026, Rio de Janeiro, Brazil, November 10-14, 2026},
  publisher    = {ACM},
  year         = {2026}
}

@inproceedings{mmskills2026,
  author       = {Kangning Zhang and
                  Shuai Shao and
                  Wenxiang Jiao and
                  Qingyao Li and
                  Jianghao Lin and
                  Lingyue Fu and
                  Shijian Wang and
                  Yuan Lu and
                  Weiwen Liu and
                  Weinan Zhang and
                  Yong Yu},
  title        = {{MMSkills}: Towards Multimodal Skills for General Visual Agents},
  booktitle    = {Advances in Neural Information Processing Systems 39: Annual Conference on Neural Information Processing Systems 2026, NeurIPS 2026},
  year         = {2026}
}

@article{imagineopd2026,
  author       = {Yishuo Cai and
                  Jiahui Liu and
                  Yuanxin Liu and
                  Haobo Deng and
                  Linli Yao and
                  Yuhao Zheng and
                  Kun Ouyang and
                  Zhimo Li and
                  Ziyue Wang and
                  Xu Sun and
                  Haoli Bai and
                  Xiaohui Li},
  title        = {Thinking Without Images: Internalizing Visual Manipulation with On-Policy Self-Distillation},
  journal      = {CoRR},
  volume       = {abs/2606.08719},
  year         = {2026},
  doi          = {10.48550/ARXIV.2606.08719},
  eprinttype   = {arXiv},
  eprint       = {2606.08719}
}

@article{jacobs1991adaptive,
  author       = {Robert A. Jacobs and
                  Michael I. Jordan and
                  Steven J. Nowlan and
                  Geoffrey E. Hinton},
  title        = {Adaptive Mixtures of Local Experts},
  journal      = {Neural Computation},
  volume       = {3},
  number       = {1},
  pages        = {79--87},
  year         = {1991},
  doi          = {10.1162/NECO.1991.3.1.79}
}

@inproceedings{shazeer2017outrageously,
  author       = {Noam Shazeer and
                  Azalia Mirhoseini and
                  Krzysztof Maziarz and
                  Andy Davis and
                  Quoc Le and
                  Geoffrey Hinton and
                  Jeff Dean},
  title        = {Outrageously Large Neural Networks: The Sparsely-Gated Mixture-of-Experts Layer},
  booktitle    = {The Fifth International Conference on Learning Representations, {ICLR} 2017, Toulon, France, April 24-26, 2017},
  publisher    = {OpenReview.net},
  year         = {2017}
}

@inproceedings{bae2025mashvlm,
  author       = {Kyungho Bae and
                  Jinhyung Kim and
                  Sihaeng Lee and
                  Soonyoung Lee and
                  Gunhee Lee and
                  Jinwoo Choi},
  title        = {{MASH-VLM}: Mitigating Action-Scene Hallucination in Video-{LLMs} through Disentangled Spatial-Temporal Representations},
  booktitle    = {{IEEE/CVF} Conference on Computer Vision and Pattern Recognition,
                  {CVPR} 2025, Nashville, TN, USA, June 11-15, 2025},
  pages        = {13744--13753},
  publisher    = {Computer Vision Foundation / {IEEE}},
  year         = {2025},
  doi          = {10.1109/CVPR52734.2025.01283}
}

@inproceedings{tian2025position,
  author       = {Xinyu Tian and
                  Shu Zou and
                  Zhaoyuan Yang and
                  Jing Zhang},
  title        = {Identifying and Mitigating Position Bias of Multi-image Vision-Language
                  Models},
  booktitle    = {{IEEE/CVF} Conference on Computer Vision and Pattern Recognition,
                  {CVPR} 2025, Nashville, TN, USA, June 11-15, 2025},
  pages        = {10599--10609},
  publisher    = {Computer Vision Foundation / {IEEE}},
  year         = {2025},
  doi          = {10.1109/CVPR52734.2025.00991}
}

@inproceedings{yeo2026worldmm,
  author       = {Woongyeong Yeo and
                  Kangsan Kim and
                  Jaehong Yoon and
                  Sung Ju Hwang},
  title        = {{WorldMM}: Dynamic Multimodal Memory Agent for Long Video Reasoning},
  booktitle    = {{IEEE/CVF} Conference on Computer Vision and Pattern Recognition,
                  {CVPR} 2026, Denver, CO, USA, June 3-7, 2026},
  pages        = {25599--25609},
  publisher    = {Computer Vision Foundation},
  year         = {2026}
}

@inproceedings{xian2026lad,
  author       = {Mingtao Xian and
                  Yifeng Yang and
                  Qinying Gu and
                  Xinbing Wang and
                  Nanyang Ye},
  title        = {{Logit-Attention Divergence}: Mitigating Position Bias in Multi-Image Retrieval via Attention-Guided Calibration},
  booktitle    = {Forty-third International Conference on Machine Learning, ICML 2026, Seoul, South Korea, July 6-11, 2026},
  series       = {Proceedings of Machine Learning Research},
  volume       = {306},
  pages        = {138447--138462},
  publisher    = {PMLR},
  year         = {2026}
}

@inproceedings{he2016resnet,
  author       = {Kaiming He and
                  Xiangyu Zhang and
                  Shaoqing Ren and
                  Jian Sun},
  title        = {Deep Residual Learning for Image Recognition},
  booktitle    = {{IEEE} Conference on Computer Vision and Pattern Recognition, {CVPR} 2016, Las Vegas, NV, USA, June 27-30, 2016},
  pages        = {770--778},
  publisher    = {{IEEE} Computer Society},
  year         = {2016},
  doi          = {10.1109/CVPR.2016.90}
}

@inproceedings{liu2025phd,
  author       = {Jiazhen Liu and
                  Yuhan Fu and
                  Ruobing Xie and
                  Runquan Xie and
                  Xingwu Sun and
                  Fengzong Lian and
                  Zhanhui Kang and
                  Xirong Li},
  title        = {{PhD}: A {ChatGPT}-Prompted Visual Hallucination Evaluation Dataset},
  booktitle    = {{IEEE/CVF} Conference on Computer Vision and Pattern Recognition,
                  {CVPR} 2025, Nashville, TN, USA, June 11-15, 2025},
  pages        = {19857--19866},
  publisher    = {Computer Vision Foundation / {IEEE}},
  year         = {2025},
  doi          = {10.1109/CVPR52734.2025.01849}
}

@inproceedings{suo2025octopus,
  author       = {Wei Suo and
                  Lijun Zhang and
                  Mengyang Sun and
                  Lin Yuanbo Wu and
                  Peng Wang and
                  Yanning Zhang},
  title        = {{Octopus}: Alleviating Hallucination via Dynamic Contrastive Decoding},
  booktitle    = {{IEEE/CVF} Conference on Computer Vision and Pattern Recognition,
                  {CVPR} 2025, Nashville, TN, USA, June 11-15, 2025},
  pages        = {29904--29914},
  publisher    = {Computer Vision Foundation / {IEEE}},
  year         = {2025},
  doi          = {10.1109/CVPR52734.2025.02783}
}
}
\clearpage
\appendix
\setcounter{table}{0}
\setcounter{figure}{0}
\renewcommand{\thesection}{\Alph{section}}
\renewcommand{\thetable}{S\arabic{table}}
\renewcommand{\thefigure}{S\arabic{figure}}
\renewcommand{\thealgorithm}{S\arabic{algorithm}}
\setcounter{dbltopnumber}{3}
\renewcommand{\dbltopfraction}{0.95}
\renewcommand{\dblfloatpagefraction}{0.8}
\setlength{\textfloatsep}{12pt plus 2pt minus 4pt}
\setlength{\dbltextfloatsep}{12pt plus 2pt minus 4pt}
\setlength{\floatsep}{8pt plus 2pt minus 2pt}
\setlength{\dblfloatsep}{8pt plus 2pt minus 2pt}
\renewcommand{\arraystretch}{0.92}
\renewcommand{\theHtable}{S\arabic{table}}
\renewcommand{\theHfigure}{S\arabic{figure}}
\twocolumn[\centering\Large\textbf{Appendix}\vspace{1.0em}]

Section~\ref{sec:s_proofs} proves the theorems of the main text, Section~\ref{sec:s_impl} details the methods and Section~\ref{sec:s_results} adds results. Numbers without the prefix S refer to the main text.

\section{Proofs}
\label{sec:s_proofs}

\subsection{Setting and two lemmas}
\label{sec:s_setting}

\noindent\textbf{Generator.} An episode of skill $s$ draws a window length $K$ (uniform on $\{4, \dots, 8\}$ for training and the held-out library, fixed on each test cell), a window start uniform over the canonical positions, and the order noise of Section~\ref{sec:s_synth}. A question names a canonically adjacent pair, $X$ before $Y$ in the canonical order, and $x$ is the option named first, a fair coin under the generator. The gold $y^\ast$ comes first in the presented order $S$, and a conflict item is a visible item with $y^\ast = Y$.

\noindent\textbf{Readers and \method.} A reader answers $X$ if its canonical log-odds $z(V)$ is positive, $Y$ if it is negative and $x$ if $z = 0$. \method uses the counts $n_{XY}$ and $n$ of Eq.~(\ref{eq:prior}), the clipped $\bar p_s$, and the check $g = \mathbf{1}[\min(\ell_X, \ell_Y) > 0]$. It answers by the argmax of the reader's option scores if $g = 1$ and by the argmax of $\log \bar p_s(y)$ if $g = 0$, with ties to $x$.

\begin{slemma}[Order noise]
\label{slem:noise}
Let $p_j$ be the probability that the canonical pair at window offset $j$ is presented swapped. Then $p_0 = \delta$ and $p_j = \delta(1 - p_{j-1})$, so $0 \le p_j \le \delta$ for every $\delta \in [0,1]$ and the canonical probability $a_j = 1 - p_j$ is at least $1 - \delta$. For $\delta \le \tfrac12$, $a_j \ge \tfrac12$, with equality only for $\delta = \tfrac12$ and $j = 0$. For $\delta < 1$ the $p_j$ converge to $\delta/(1+\delta)$.
\end{slemma}

\noindent\emph{Proof.} Swaps do not overlap and move every unit by at most one place, so the pair at offset $j$ is swapped exactly when the swap at position $j$ fires. That swap is tried only if position $j$ was not consumed by a swap at $j-1$, and then fires with probability $\delta$, which gives the recursion. Since $1 - p_{j-1} \in [0,1]$, $0 \le p_j \le \delta$. For $\delta \le \tfrac12$ this gives $a_j \ge \tfrac12$. Equality needs $\delta = \tfrac12$ and $p_j = \delta$, that is $p_{j-1} = 0$ for $j \ge 1$, which fails because $p_{j-1} \ge \delta(1-\delta) > 0$. The map $p \mapsto \delta(1-p)$ is a contraction with factor $\delta < 1$ and fixed point $\delta/(1+\delta)$. \qed

The pair $(u_1, u_2)$ sits only at offset $0$, so at $\delta = \tfrac12$ its canonical probability is $\tfrac12$, and results that need a strict majority hold on the order noise grid for every other pair.

\begin{slemma}[Direction of the counted prior]
\label{slem:direction}
$\bar p_s(X \text{ first}) > \tfrac12$ if and only if $n_{XY} > n/2$, and $\bar p_s(X \text{ first}) = \tfrac12$ if and only if $n_{XY} = n/2$, including $n = 0$. The same holds for the posterior mean $(n_{XY}+c)/(n+2c)$ under every symmetric prior $\mathrm{Beta}(c,c)$ with $c > 0$. The switch uses only the argmax of $\log \bar p_s$, so neither the prior strength $c$ nor the clipping changes an answer of \method.
\end{slemma}

\noindent\emph{Proof.} For every $c > 0$, $(k+c)/(n+2c) > \tfrac12$ if and only if $2k + 2c > n + 2c$, that is $k > n/2$, and equality likewise. Eq.~(\ref{eq:prior}) is the case $c = 1$. Clipping to $[0.02, 0.98]$ is nondecreasing and fixes $\tfrac12$. \qed

\noindent\textbf{Posterior mean.} Let the $n$ orders of a pair be independent given $\theta = P(X \text{ first})$, with prior $\theta \sim \mathrm{Beta}(c,c)$. The likelihood of the counts is proportional to $\theta^{n_{XY}}(1-\theta)^{n-n_{XY}}$, so the posterior is $\mathrm{Beta}(n_{XY}+c,\, n-n_{XY}+c)$ with mean $(n_{XY}+c)/(n+2c)$. By the tower rule a further order of the pair has $X$ first with probability $\mathbb{E}[\theta \mid n_{XY}, n]$, which is Eq.~(\ref{eq:prior}) at the uniform prior $c = 1$. The orders of the generator are independent with unequal canonical probabilities (Lemma~\ref{slem:noise}), the law under which Proposition~\ref{sprop:counts} bounds the direction error of the counts.

Under (A3.1), the law of large numbers gives $\bar p_s-\operatorname{clip}_{[0.02,0.98]}(\bar q_p)\to0$ as $n_p\to\infty$, with $\bar q_p$ from Proposition~\ref{sprop:counts}. By Lemma~\ref{slem:direction}, only its side of $\tfrac12$ matters. The test sampler drops every candidate whose key belongs to the training pool, the development questions or another evaluation set, so the test items follow a rejection-sampled law, and the checks of Theorem~\ref{thm:identify} and Proposition~\ref{sprop:counts} condition on the realized test items.

\subsection{Theorem \ref{thm:identify}}
\label{sec:s_t1}

\noindent\textbf{Family.} Fix an evidence term $e(V)$, any real function of the context and the question, and a prior shape $\psi_q$, a real function of the question such as $\logit \bar p_s$ or a constant. The shared-weight family $R(\lambda, \rho)$ has $z(V) = e(V) + \lambda\psi_q(1 - \rho\,\mathrm{vis}(V))$ with $\lambda \ge 0$ and $\rho \in [0,1]$, and $\mu = (1-\rho)\lambda$ is the part of the prior strength that reaches visible items. Let $F$, $C$ and $N$ be sets of frame-free items, visible conflict items and visible non-conflict items, with margin toward the gold $M = e$ on $N$ and $M = -e$ on $C$. Items with $\psi_q = e_q = 0$ tie for every $\lambda$ and are left out. On every other item a tie, $z = 0$, occurs at one value of $\lambda$ or $\mu$ at most, the accuracy statements below hold at every value without a tie, and $\mathbb{E}A_F$, $A_C$ and $A_N$ denote the displayed sums at every value.
\begin{enumerate}[label=(A1.\arabic*),leftmargin=*,nosep]
\item Given its question, the presented order of a frame-free item is independent of $e$, and its gold is canonical with probability $a_q$ under the law that drew the items. At $m = 0$, $e$ is a function of the question, so this holds for any law of the presented order.
\item $\psi_q > 0$ and $a_q \ge \tfrac12$ on every item considered.
\item Symmetric reading: for each $\delta$, the margin has the same law on the visible conflict and on the visible non-conflict items of that $\delta$.
\item Normalization: $\mathbb{E}[e \mid F, \delta] = 0$, so the stored strength is the mean frame-free log-odds and any canonical bias of frame-free evidence is stored prior.
\end{enumerate}
The fair naming coin makes this bias zero for a reader whose frame-free evidence depends on the naming order and the words but not on which option is canonical.

\begin{sprop}[Frontier]
\label{sprop:frontier}
Assume (A1.1). Over the presented orders of the frame-free items given their questions, $R(\lambda, \rho)$ has
\begin{align*}
\mathbb{E} A_F(\lambda) &= \frac{1}{|F|}\sum_{q \in F}\Big[(1 - a_q) \\
&\quad + (2a_q - 1)\,\mathbf{1}\big[e_q > -\lambda\psi_q\big]\Big], \\
A_C(\mu) &= \frac{1}{|C|}\sum_{c \in C}\mathbf{1}\big[M_c > \mu\psi_c\big], \\
A_N(\mu) &= \frac{1}{|N|}\sum_{n \in N}\mathbf{1}\big[M_n > -\mu\psi_n\big].
\end{align*}
Under (A1.2), $\mathbb{E}A_F$ is nondecreasing in $\lambda$, $A_C$ is nonincreasing and $A_N$ nondecreasing in $\mu$, and the prior alone has frame-free accuracy $A_F^{\mathrm{prior}} = |F|^{-1}\sum_q a_q$. For $\varepsilon \ge 0$ let $\lambda_\varepsilon = \inf\{\lambda \ge 0 : \mathbb{E}A_F(\lambda) \ge A_F^{\mathrm{prior}} - \varepsilon\}$. Among the members with switch quality $\rho$ whose frame-free accuracy is within $\varepsilon$ of the prior's, the supremum of the conflict accuracy is $A_C((1-\rho)\lambda_\varepsilon)$, so the least conflict cost is
\begin{equation*}
\mathcal{C}_\rho(\varepsilon) = \frac{\#\{c \in C : 0 < M_c \le (1-\rho)\lambda_\varepsilon\psi_c\}}{|C|},
\end{equation*}
nonincreasing in $\rho$ and in $\varepsilon$ and $0$ at $\rho = 1$. Every pair $(\mathbb{E}A_F, A_C)$ reachable with switch quality at most $\bar\rho$ lies on or below a point of $\Gamma_{\bar\rho} = \{(\mathbb{E}A_F(\lambda), A_C((1-\bar\rho)\lambda)) : \lambda \ge 0\}$ with the same frame-free accuracy.
\end{sprop}

\noindent\emph{Proof.} On $F$, $\mathrm{vis} = 0$ and $z = e_q + \lambda\psi_q$. By (A1.1) the gold is canonical with probability $a_q$ independently of $z$, so the item is right with probability $a_q\mathbf{1}[z > 0] + (1-a_q)\mathbf{1}[z < 0]$, the summand when $z \ne 0$. On $C$, $z = e_c + \mu\psi_c$, the gold is $Y$, and the item is right if and only if $z < 0$, that is $M_c > \mu\psi_c$. On $N$ it is right if and only if $z > 0$, that is $M_n > -\mu\psi_n$. With $\psi_q > 0$ and $2a_q - 1 \ge 0$ every summand of $\mathbb{E}A_F$ is nondecreasing in $\lambda$ and tends to $a_q$, and the indicators on $C$ and $N$ are monotone in $\mu$ as stated. For fixed $\rho$ the feasible set of $\lambda$ is an interval, unbounded above, with left end $\lambda_\varepsilon$, and $A_C((1-\rho)\lambda)$ is nonincreasing in $\lambda$, so its supremum over the set is its right limit at $\lambda_\varepsilon$. Each $\mathbf{1}[M_c > \mu\psi_c]$ is right-continuous in $\mu$, so this limit is $A_C((1-\rho)\lambda_\varepsilon)$. The items right at $\mu = 0$ and wrong at $\mu$ are those with $0 < M_c \le \mu\psi_c$, which gives the count, and $(1-\rho)\lambda_\varepsilon$ is nonincreasing in $\rho$ and in $\varepsilon$. For fixed $\lambda$, $\rho \in [0, \bar\rho]$ gives $\mu \in [(1-\bar\rho)\lambda, \lambda]$, and $A_C$ is largest at the smallest $\mu$. \qed

A correct check is the case $\rho = 1$: \method then answers frame-free items by the prior and visible items by the reader (Lemma~\ref{slem:routes} with $\alpha = \beta = 0$), as the member $\rho = 1$ of the family built from its two parts ($e$ its reader's evidence, $\psi = \logit \bar p_s$), whose conflict cost $\mathcal{C}_1(\varepsilon)$ is $0$ for every $\varepsilon$.

\begin{sprop}[What stored scores identify, full form]
\label{sprop:identify}
In the setting of Theorem~\ref{thm:identify}, with $\pi \in \mathbb{R}$, $\kappa = (1-\rho)\pi$ and margin toward the gold $e$ on a non-conflict item and $-e$ on a conflict item, assume (A1.3) and (A1.4). Then (i) to (iii) of Theorem~\ref{thm:identify} hold. $Z_C$ has the law of $2\kappa - Z_N$. The bounds of (ii) are the values that $\rho = 0$ and $\rho = 1$ give for the same law of $Z_N$. The reader's conflict loss against any reference splits into reading, the reference minus $P(Z_N > \kappa)$, and bleeding, $P(Z_N > \kappa) - P(Z_N > 2\kappa)$.
\end{sprop}

\noindent\textbf{Proof of Proposition~\ref{sprop:identify} and Theorem~\ref{thm:identify}.} The reader is $R(\lambda, \rho)$ with constant shape on the items of $\delta$ and $\pi = \lambda\psi$, and the two conditions of Theorem~\ref{thm:identify} are (A1.3) and (A1.4). (i) On $N$, $z = e + \kappa$ and $M = e$, so $Z_N = M_N + \kappa$. On $C$, $z = e + \kappa = \kappa - M_C$. By (A1.3), $Z_C = \kappa - M_C$ has the law of $\kappa - M_N = 2\kappa - Z_N$. On $F$, $z = e + \pi$, so (A1.4) gives $\mathbb{E}Z_F = \pi$. (ii) A conflict item is right when $Z_C < 0$, with probability $P(2\kappa - Z_N < 0) = P(Z_N > 2\kappa)$, nonincreasing in $\kappa$ for a fixed law of $Z_N$, and $\kappa$ lies between $0$ and $\pi$. (iii) With the prior switched off on visible items and $P(e = 0) = 0$ there, the decision is $\mathrm{sign}(e)$, so the conflict accuracy is $P(M_C > 0) = P(M_N > 0) = P(Z_N > \kappa)$. The split is the identity $\mathrm{ref} - P(Z_N > 2\kappa) = [\mathrm{ref} - P(Z_N > \kappa)] + [P(Z_N > \kappa) - P(Z_N > 2\kappa)]$, in which the second term is identified by the law of $Z_C$ shown in (i). \qed

\noindent\textbf{Closed form.} Since $Z_C$ has the law of $2\kappa - Z_N$, $\kappa = \tfrac12(\mathbb{E}Z_N + \mathbb{E}Z_C)$ when the means exist, and likewise with unique medians. Stored scores thus identify $\kappa$ and, for $\pi \ne 0$, $\rho = 1 - \kappa/\pi$ in closed form, and the reflection of the whole law of $Z_N$ is a testable consequence of (A1.3). Theorem~\ref{thm:identify} is a measurement model: (A1.4) defines the stored prior and stored scores give $\rho$. Proposition~\ref{sprop:incentive} adds why training leaves the prior on.

\noindent\textbf{Estimation.} Per $\delta$, stored option scores give $\hat\pi_\delta$, the mean frame-free canonical log-odds, the laws of $Z_N$ and $Z_C$, and $\hat\rho$, the first point of a grid on $[-3, 1]$ ($\rho < 0$ fits accuracies below the band) minimizing the distance between the measured conflict accuracy and $P(Z_N > 2(1-\rho)\hat\pi^+_\delta)$ with $\hat\pi_\delta^+ = \max(0, \hat\pi_\delta)$. A case is informative when the band of (ii) is wider than five points, since a narrower band leaves the prediction nearly flat in $\rho$. The family is rejected where the measured accuracy falls below the band.

\subsection{Proposition \ref{prop:bayes}}
\label{sec:s_bayes}

Let $0 < a_s < 1$, $v_X = P(\mathrm{vis}(V)=1 \mid y^\ast = X)$ and $v_Y = P(\mathrm{vis}(V)=1 \mid y^\ast = Y)$, so that $v = a_s v_X + (1-a_s)v_Y$. The statistic $T$ takes the values $(1,X)$ and $(1,Y)$, both named units in view with $X$ or $Y$ first in $V$, and $0$, not both in view. Eq.~(\ref{eq:posterior}) is the logarithm of Bayes' rule
\begin{equation*}
\frac{P(y^\ast = X \mid T)}{P(y^\ast = Y \mid T)} = \frac{P(T \mid y^\ast = X)}{P(T \mid y^\ast = Y)} \cdot \frac{a_s}{1-a_s},
\end{equation*}
with $\Lambda = \pm\infty$ where one of the two likelihoods is zero.

\noindent\emph{Proof of Proposition~\ref{prop:bayes}.} (i) $V$ is an order-preserving subsequence of $S$, so when both named units are in view the one that comes first in $V$ comes first in $S$ and is the gold. Hence $P(T = (1,X) \mid y^\ast = Y) = 0$ and $P(T = (1,Y) \mid y^\ast = X) = 0$, which gives $\Lambda(1,X) = +\infty$ and $\Lambda(1,Y) = -\infty$ with the posteriors $1$ and $0$. Out of view, $\Lambda(0) = \log[(1-v_X)/(1-v_Y)]$, which is $0$ under the condition $v_X = v_Y$, and the posterior is $a_s$. These are the three values of Definition~\ref{def:oracle}. (ii) A rule $d$ that uses the frames through $T$, with any randomization independent of the gold given $T$, is right with probability $\mathbb{E}[P(y^\ast = d \mid T)] \le \mathbb{E}[\max_y P(y^\ast = y \mid T)]$, with equality for the most probable answer. The maximum is $1$ in view and $\max(a_s, 1-a_s)$ out of view, which gives $A^\ast$. \qed

\noindent\textbf{General form.} Without the condition, (i) and (ii) hold by the same proof with $a_s$ replaced out of view by $a_s^0 = P(y^\ast = X \mid \mathrm{vis}(V)=0)$, which Bayes' rule gives as
\begin{equation*}
a_s^0 = \frac{a_s(1-v_X)}{a_s(1-v_X) + (1-a_s)(1-v_Y)},
\end{equation*}
and $A^\ast = v + (1-v)\max(a_s^0, 1-a_s^0)$. The condition holds wherever $\mathrm{vis}(V)$ is constant given the question: with no frame, where $v = 0$, and on the interval and full-window cells, where $v = 1$. On the two-frame short cells the view depends on the presented order of some pairs: with the first two frames shown, the first pair of the window is out of view only in its canonical order, so $a_s^0 \ne a_s$. The oracle reference keeps $a_s$ on every cell. It is a rule that uses the frames through $T$, so its accuracy is at most the general $A^\ast$.

\noindent\textbf{Posterior with the counts.} Let the presented order of the test item and the $n$ counted orders of its pair be exchangeable: independent given the rate $\theta$ of $X$ first, with a uniform prior on $\theta$. Let $\mathrm{vis}(V)$ be independent of the test order given the counts. Then the posterior probability $P(y^\ast = X \mid T, n_{XY}, n)$ is
\begin{equation*}
\begin{cases}
\mathbf{1}[X \text{ before } Y \text{ in } V], & \mathrm{vis}(V) = 1,\\[2pt]
\dfrac{n_{XY}+1}{n+2}, & \mathrm{vis}(V) = 0.
\end{cases}
\end{equation*}
In view the gold is read off $V$ as in (i). Out of view $T$ leaves the test order as it was given the counts, and the tower rule of Section~\ref{sec:s_setting} gives $\mathbb{E}[\theta \mid n_{XY}, n]$. So Eq.~(\ref{eq:score}), with a correct check and a reader that follows the shown order, returns the most probable answer of one posterior, in which the count of Eq.~(\ref{eq:prior}) before its clip is the posterior itself, and the clip keeps its side (Lemma~\ref{slem:direction}). Averaged over independent window draws of one law, the generator's counted orders are independent with one rate, the mean of the $a_j$ of Lemma~\ref{slem:noise} over the pair's offsets, hence exchangeable, and so is a test item whose window follows the same law. A test cell that fixes $K$ draws its order at a slightly different rate, and given the windows the rates differ, the law of Proposition~\ref{sprop:counts}.

\noindent\textbf{Sum against switch.} A sum of the log prior and the reader's scores is the posterior of Eq.~(\ref{eq:posterior}) only if those scores are its frame term $\Lambda$, zero without decisive frames. Measured, the OPSD reader's frame-free scores are wide and symmetric (Fig.~\ref{fig:s_order_store}), so the unchecked sum stays near the reader alone on the short side (Table~\ref{tab:s_ablation}), and the check supplies what the scores lack: whether the frames are decisive.

\noindent\textbf{Counts as a sufficient statistic.} In the model above $(n_{XY}, n)$ is sufficient for $\theta$, and the posterior is updated by adding counts. A held-out skill therefore needs only a library of its episodes, and the counts with a half-life of Section~\ref{sec:s_beyond} are the posterior mean of the tempered update that raises the likelihood of each old order to its decay weight. For next-step questions the counts with Laplace smoothing are the posterior mean of a multinomial rate over the offered options under a uniform Dirichlet prior.

\subsection{Theorem \ref{thm:switch}(a)}

\begin{slemma}[Routes]
\label{slem:routes}
Split a set of questions with any number of options into the routes A (visible, $g = 1$), B (visible, $g = 0$), C (not visible, $g = 0$) and D (not visible, $g = 1$). Let $v$ be the visible share, $\alpha$ the share of B among visible questions (miss rate), $\beta$ the share of D among the others (false-yes rate), and $R_k$ and $P_k$ the accuracies of the reader alone and of the counted prior alone on route $k$. Then \method has accuracy $v[(1-\alpha)R_A + \alpha P_B] + (1-v)[(1-\beta)P_C + \beta R_D]$. On any set of visible questions with miss rate $\alpha^c$ it is below the reader alone by $\alpha^c(R_B - P_B) \le \alpha^c$, with $R_B$ and $P_B$ taken on the missed questions of that set. For two options and a prior that is strictly canonical on every pair, $n_{XY} > n/2$ in Eq.~(\ref{eq:prior}), this loss is $\alpha^c R_B$ on visible conflict items and $-\alpha^c(1 - R_B) \le 0$ on the other visible items, so those cannot lose.
\end{slemma}

\noindent\emph{Proof.} On routes A and D \method outputs the reader's answer and on B and C the prior's, so its accuracy on route $k$ is $R_k$ or $P_k$, and the total is the sum weighted by the route shares $v(1-\alpha)$, $v\alpha$, $(1-v)(1-\beta)$ and $(1-v)\beta$. On a set of visible questions only A and B occur, where the reader alone has accuracy $(1-\alpha^c)R_A + \alpha^c R_B$ and \method $(1-\alpha^c)R_A + \alpha^c P_B$, a difference of $\alpha^c(R_B - P_B) \le \alpha^c$ since $R_B \le 1$ and $P_B \ge 0$. A strictly canonical two-option prior answers $X$ on every pair, wrong on every conflict item and right on every visible non-conflict item, so $P_B$ is $0$ on the first and $1$ on the second, which gives the two losses. \qed

\noindent\emph{Proof of Theorem~\ref{thm:switch}(a).} On a set of visible questions with miss share $\alpha$ the claim is the second statement of Lemma~\ref{slem:routes}, with $A_E = R_B$ and $A_{\bar p} = P_B$. On the questions without both named units only the routes C and D occur, so the counted prior alone has accuracy $(1-\beta)P_C + \beta P_D$ there and \method has $(1-\beta)P_C + \beta R_D$, and \method is below the prior alone by $\beta(P_D - R_D)$, the claim with $A_E = R_D$ and $A_{\bar p} = P_D$. With $\alpha = \beta = 0$, \method answers every visible question as the reader and every other question as the counted prior. \qed

With $\Delta_E$ and $\Delta_{\bar p}$ the accuracy of \method minus that of the reader alone $E$ and of the counted prior alone $\bar p$, the same algebra gives the two differences used in Proposition~\ref{sprop:tolerable},
\begin{align*}
\Delta_E &= v\alpha(P_B - R_B) + (1-v)(1-\beta)(P_C - R_C), \\
\Delta_{\bar p} &= v(1-\alpha)(R_A - P_A) + (1-v)\beta(R_D - P_D).
\end{align*} Lemma~\ref{slem:routes} holds for the realized routes of one seed and for counts pooled over seeds, so every check pools counts or pairs per-seed quantities.

\begin{sprop}[Tolerable check errors]
\label{sprop:tolerable}
In Lemma~\ref{slem:routes}, let $\alpha, \beta < 1$, and let $R_{\mathrm{vis}}, P_{\mathrm{vis}}$ and $R_{\mathrm{inv}}, P_{\mathrm{inv}}$ be the accuracies of the reader and of the prior on the visible questions and on the others, with the gains $G_{\mathrm{inv}} = (1-v)(P_{\mathrm{inv}} - R_{\mathrm{inv}}) > 0$ and $G_{\mathrm{vis}} = v(R_{\mathrm{vis}} - P_{\mathrm{vis}}) > 0$. Assume (A2.1): given visibility, the check errs independently of which part is right ($R_A = R_B$, $P_A = P_B$, $R_C = R_D$, $P_C = P_D$). Then \method beats the reader alone if and only if $\alpha < \alpha^\ast = (1-\beta)G_{\mathrm{inv}}/G_{\mathrm{vis}}$, and the prior alone if and only if $\beta < \beta^\ast = (1-\alpha)G_{\mathrm{vis}}/G_{\mathrm{inv}}$. Both can hold, for some ratio $G_{\mathrm{inv}}/G_{\mathrm{vis}}$, if and only if $\alpha + \beta < 1$. If instead $v = 1$, \method beats the prior alone if and only if $R_A > P_A$, also without (A2.1) and without $G_{\mathrm{inv}}, G_{\mathrm{vis}} > 0$.
\end{sprop}

\noindent\emph{Proof.} Under (A2.1), $\Delta_E = (1-\beta)G_{\mathrm{inv}} - \alpha G_{\mathrm{vis}}$ and $\Delta_{\bar p} = (1-\alpha)G_{\mathrm{vis}} - \beta G_{\mathrm{inv}}$. With $G_{\mathrm{inv}}, G_{\mathrm{vis}} > 0$ and $\alpha, \beta < 1$, dividing gives $\alpha^\ast$ and $\beta^\ast$. Both inequalities hold for the ratio $r = G_{\mathrm{inv}}/G_{\mathrm{vis}}$ if and only if $\alpha/(1-\beta) < r < (1-\alpha)/\beta$, with the right end $+\infty$ for $\beta = 0$. Such an $r$ exists if and only if $\alpha\beta < (1-\alpha)(1-\beta)$, that is $\alpha + \beta < 1$. For the last statement, at $v = 1$ (where $G_{\mathrm{inv}} = 0$) $\Delta_{\bar p} = (1-\alpha)(R_A - P_A)$ for any route accuracies, and $1 - \alpha > 0$. \qed

\subsection{Theorem \ref{thm:switch}(b)}

\noindent\textbf{Assumptions.} On synthetic skills the reader consumes the first $1600$ draws of a shuffled pool of $2016$ distinct episodes. For a pair $p$ of skill $s$, $n_p$ consumed windows contain both units, $j_i$ is the pair's offset in window $i$ and $q_i = a_{j_i}(\delta_s)$ of Lemma~\ref{slem:noise}.
\begin{enumerate}[label=(A3.\arabic*),leftmargin=*,nosep]
\item Given the windows of the consumed episodes, their presented orders are independent, each produced by the order noise. The pool draws every episode from its own seeded stream.
\item Conditional on the realized test items (questions, presented orders and naming orders), the consumed orders still follow (A3.1). This is exact if the test items are drawn independently of the training episodes, and the test sampler makes it an approximation through the excluded pool keys.
\end{enumerate}
\begin{sprop}[Counted prior]
\label{sprop:counts}
Assume (A3.1) and (A3.2). For a pair $p$ in $n_p \ge 1$ consumed windows, let $\bar q_p$ be the mean of the $q_i$, $S_p = n_{XY}$, $w_p = P(S_p < n_p/2)$ and $t_p = P(S_p = n_p/2)$.
(a) $S_p$ is Poisson-binomial with parameters $q_1, \dots, q_{n_p}$. The counted prior points the wrong way with probability $w_p$ and ties with probability $t_p$, and if $\bar q_p > \tfrac12$, then $w_p + t_p \le \exp\big(-2n_p(\bar q_p - \tfrac12)^2\big)$. Under the order noise, $\bar q_p \ge 1 - \delta_s$. On a visible conflict item of $p$ the prior alone is right with probability $w_p$, plus $t_p$ if the option named first is the gold.
(b) On a frame-free item of $p$ whose presented order is canonical with probability $a_s \ge \tfrac12$, named in an order that is a fair coin independent of the gold, the prior alone and \method with a correct check are right with probability $a_s - (2a_s - 1)(w_p + \tfrac12 t_p)$, and the oracle reference with probability $a_s$. The gap is at most $(2a_s - 1)\exp\big(-2n_p(\bar q_p - \tfrac12)^2\big)$ if $\bar q_p > \tfrac12$.
(c) Let $s'$ be $s$ with the two units of $p$ swapped in the canonical order. Consider a rule without frames whose answers about $p$ have the same law under $s$ and $s'$, such as one fitted without episodes of either. Its accuracies sum to $1$ on every question about $p$, so its gap to the oracle reference is at least $a_s - \tfrac12$ on $s$ or on $s'$.
\end{sprop}

\noindent\emph{Proof.} (a) Given the realized test items, $S_p$ is a sum of independent Bernoulli($q_i$) variables, the Poisson-binomial law, and Lemma~\ref{slem:direction} turns the direction of the prior into $S_p$ against $n_p/2$. Hoeffding's inequality for independent variables in $[0,1]$ gives $P(S_p - n_p\bar q_p \le -t) \le \exp(-2t^2/n_p)$, and $t = n_p(\bar q_p - \tfrac12) > 0$ gives $w_p + t_p = P(S_p \le n_p/2) \le \exp(-2n_p(\bar q_p - \tfrac12)^2)$. By Lemma~\ref{slem:noise} every $q_i \ge 1 - \delta_s$. A prior that points canonical is right exactly when the gold is canonical, one that points the wrong way exactly when it is not, a tie answers $x$, and a conflict item has the gold $Y$. (b) Averaging (a) over the gold and the naming coin gives $a_s(1 - w_p - t_p) + (1 - a_s)w_p + \tfrac12 t_p = a_s - (2a_s - 1)(w_p + \tfrac12 t_p)$. With a correct check \method answers frame-free items by the prior (Lemma~\ref{slem:routes}). The oracle reference answers the canonical option and is right with probability $a_s$. The bound on the gap uses $w_p + \tfrac12 t_p \le w_p + t_p$ and (a). (c) The swap does not move the canonical positions over which windows are drawn, so $s$ and $s'$ share the offset law of $p$ and $a_s$. A rule naming $X$, canonical in $s$, with probability $\phi$ has accuracies $\phi a_s + (1-\phi)(1-a_s)$ on $s$ and $\phi(1-a_s) + (1-\phi)a_s$ on $s'$, which sum to $1$. Its two gaps to the oracle reference thus sum to $2a_s - 1$, and the larger is at least $a_s - \tfrac12$. \qed

\noindent\emph{Proof of Theorem~\ref{thm:switch}(b).} Its hypothesis is the part of (A3.1) and (A3.2) that (a) and (b) use. By Proposition~\ref{sprop:counts}(b) the prior alone and \method with a correct check are right with probability $a_s - (2a_s - 1)(w_p + \tfrac12 t_p)$, which is at most $a_s$ and at least $a_s - (2a_s - 1)(w_p + t_p)$, since $2a_s - 1 \ge 0$, and by (a) $w_p + t_p \le \exp\big(-2n_p(\bar q_p - \tfrac12)^2\big)$. The checks apply Proposition~\ref{sprop:counts}(a) per realized test item, with the Poisson-binomial $w_p$ and $t_p$, which is (b) before the average over the gold and the naming coin. \qed

Part (c) separates rules by access to the skill's episodes: on held-out skills \method receives a library of the skill's episodes and the students receive none. Theorem~\ref{thm:switch}(a), Lemma~\ref{slem:routes} and Proposition~\ref{sprop:tolerable} hold for any number of options, so they cover next-step and CaptainCook4D questions.

\subsection{Corollary \ref{cor:regret}}
\label{sec:s_regret}

\noindent\emph{Proof of Corollary~\ref{cor:regret}.} The oracle reference is right on every visible question, so $A^\ast = v + (1-v)a^\ast$. With the routes of Lemma~\ref{slem:routes}, $R_{\mathrm{vis}} = (1-\alpha)R_A + \alpha R_B$ and $P_{\mathrm{inv}} = (1-\beta)P_C + \beta P_D$, and the lemma reads $A = v[R_{\mathrm{vis}} - \alpha(R_B - P_B)] + (1-v)[P_{\mathrm{inv}} - \beta(P_D - R_D)]$. Subtracting gives Eq.~(\ref{eq:regret}) with $L_\alpha = \alpha(R_B - P_B)$ and $L_\beta = \beta(P_D - R_D)$, the two losses of Theorem~\ref{thm:switch}(a). The identity holds for the realized routes of any set of questions with any number of options. For the bound, a correct check has $\alpha = \beta = 0$, so $L_\alpha = L_\beta = 0$. Under the condition of Proposition~\ref{prop:bayes}, a question of pair $p$ out of view is canonical with probability $a_s$, by (A3.2) independently of the counts, so Proposition~\ref{sprop:counts}(b) applies to it: $a^\ast = a_s$ and the prior is right with probability $a_s - (2a_s-1)(w_p + \tfrac12 t_p)$. The expected counting term is thus $(1-v)(2a_s - 1)(w_p + \tfrac12 t_p) \le (1-v)(2a_s-1)\exp\big(-2n_p(\bar q_p - \tfrac12)^2\big)$ by Proposition~\ref{sprop:counts}(a). Since $\bar q_p \ge 1 - \delta_s$, the bound tends to $0$ as $n_p$ grows for every $\delta_s < \tfrac12$. \qed

\subsection{Training loss and prior shift}
\label{sec:s_loss}

Theorem~\ref{thm:identify} measures how far the prior stays on in view. Three results add why training leaves it on and what conflict accuracy measures.

\begin{sprop}[Full windows leave the prior unconstrained]
\label{sprop:noident}
Let a loss be $\mathcal{L}(w) = \mathbb{E}_{(q, V) \sim \mathcal{D}}\,\ell\big(p_w(\cdot \mid q, V), q, V\big)$ for weights $w$ and a training law $\mathcal{D}$ with $\mathrm{vis}(V) = 1$ almost surely, as for a student that sees the full window. Then $\mathcal{L}(w) = \mathcal{L}(w')$ whenever $p_w$ and $p_{w'}$ agree almost surely under $\mathcal{D}$, whatever their log-odds on contexts with $\mathrm{vis}(V) = 0$. Given such a $V$ the gold is the named unit that comes first in $V$ (Section~\ref{sec:s_bayes}), so a target that follows the presented order carries no information on the order probability of the pair beyond $V$.
\end{sprop}

\noindent\emph{Proof.} $\mathcal{L}$ integrates under $\mathcal{D}$ a function of the values of $p_w$ on contexts drawn from $\mathcal{D}$. The gold is a function of $V$ and hence independent of every other variable given $V$. \qed

The loss of OPSD has this form: its student sees the full window and matches the teacher on its own samples (Section~\ref{sec:s_training}), and on a full window the teacher's privileged order is a function of $V$. The proposition leaves the frame-free prior $\pi = \mathbb{E}Z_F$ of Theorem~\ref{thm:identify} to the base model and the optimizer, and OPSD keeps at most $\ThOpsdPiMax$ nats of it per $\delta$ (Section~\ref{sec:s_checks}).

\begin{sprop}[The loss does not ask for a switch]
\label{sprop:incentive}
On visible items let a student have canonical log-odds $z = e + \kappa$ with a fixed evidence term $e$ and a leaked prior $\kappa \in \mathbb{R}$. Let the gold be canonical on a share $a \in [\tfrac12, 1)$ of them, and let $M$ be the margin toward the gold, $e$ on non-conflict and $-e$ on conflict items. The cross-entropy to the gold on visible items is
\begin{align*}
\mathcal{L}(\kappa) &= a\,\mathbb{E}_N\log\big(1 + \exp(-M - \kappa)\big) \\
&\quad + (1 - a)\,\mathbb{E}_C\log\big(1 + \exp(\kappa - M)\big),
\end{align*}
with expectations over the non-conflict and the conflict items, assumed finite.
(a) $\mathcal{L}$ is strictly convex with one minimizer. Under (A1.3), $\mathcal{L}'(0) = -(2a - 1)\,\mathbb{E}\,\sigma(-M) \le 0$ with $\sigma$ the logistic function, so the minimizer is nonnegative, and positive if $a > \tfrac12$.
(b) If $e$ is a calibrated log-likelihood ratio, $P(\text{gold canonical} \mid e) = \sigma(e + \logit a)$, then $\kappa = \logit a$ minimizes $\mathcal{L}$, also among all functions of $e$. With $a > \tfrac12$ and the frame-free prior at the same value $\pi = \logit a$ this is $\rho = 0$.
(c) With $\kappa > 0$ a conflict item of margin $M > 0$ pays at most $(\exp\kappa - 1)\exp(-M)$ more cross-entropy than with $\kappa = 0$, and its answer is wrong if $M < \kappa$ and right if $M > \kappa$.
\end{sprop}

\noindent\emph{Proof.} (a) $\mathcal{L}'(\kappa) = -a\,\mathbb{E}_N\,\sigma(-(M + \kappa)) + (1 - a)\,\mathbb{E}_C\,\sigma(-(M - \kappa))$, and $\mathcal{L}''(\kappa) > 0$ since $\sigma' > 0$ and $a$ and $1 - a$ are positive. Both terms are nonnegative, the first grows without bound as $\kappa \to -\infty$ and the second as $\kappa \to +\infty$, so $\mathcal{L}$ has one minimizer. By (A1.3) the two expectations agree at $\kappa = 0$, which gives $\mathcal{L}'(0)$, negative if $a > \tfrac12$. The minimizer is thus not below $0$, and not at $0$ if $\mathcal{L}'(0) < 0$. (b) The cross-entropy is a proper score, so given $e$ it is minimized by the true log-odds $e + \logit a$, which lie in the family at $\kappa = \logit a$. (c) The difference is $\log\frac{1 + \exp(\kappa - M)}{1 + \exp(-M)} \le \log\big(1 + (\exp\kappa - 1)\exp(-M)\big) \le (\exp\kappa - 1)\exp(-M)$. A conflict item has $z = \kappa - M$ and is right if $z < 0$ and wrong if $z > 0$. \qed

\noindent\textbf{Reading.} The proposition stands for training with a target close to the gold, as the OPSD teacher sees the privileged order. A student's evidence $e$ is finite where the frame term $\Lambda$ of Eq.~(\ref{eq:posterior}) is not, so the slope of (a) is not zero for $a > \tfrac12$. A student also trained on contexts with $\mathrm{vis}(V) = 0$, as in subsequence OPSD, the SFT variants and prior distillation, is driven there toward the log-odds of the order probability. By (a) the loss on visible contexts is least with that prior not switched off, and by (b) with all of it on for a calibrated reader. By (c) the loss charges a conflict item of margin $M$ at most $(\exp\kappa - 1)\exp(-M)$ while its answer flips once $0 < M < \kappa$, the items that $\mathcal{C}_\rho$ of Proposition~\ref{sprop:frontier} counts. Bleeding is thus a mismatch between an average proper score and conflict accuracy and not a failure to optimize, in line with Table~\ref{tab:s_budget}, where five times the steps do not restore conflict accuracy. On full windows alone the slope of (a) vanishes as reading saturates and Proposition~\ref{sprop:noident} leaves the frame-free log-odds free, in line with the small frame-free prior of OPSD.

\begin{sprop}[Prior shift on visible items]
\label{sprop:shift}
Let a share $\eta$ of the visible test items be conflict items, whatever the share in training. (a) Without ties, a reader of the family $R(\lambda, \rho)$ with constant shape, $\kappa = (1 - \rho)\lambda\psi \ge 0$ on visible items and (A1.3) has visible accuracy $A(\eta) = (1 - \eta)P(M > -\kappa) + \eta P(M > \kappa)$. It is nonincreasing in $\eta$ with slope $-P(-\kappa < M \le \kappa)$, its minimum over $\eta \in [0, 1]$ is the conflict accuracy $P(M > \kappa)$, and at $\eta = 1$ it is the bleeding $P(0 < M \le \kappa)$ below the same reader with $\kappa = 0$. (b) For two options and a strictly canonical prior, \method minus its reader alone is $(1 - \eta)\,\alpha_N(1 - R_B^N) - \eta\,\alpha_C R_B^C$ on visible items, with $\alpha_N$ and $\alpha_C$ the miss shares on non-conflict and conflict items and $R_B^N$, $R_B^C$ the reader's accuracies on the missed ones (route B of Lemma~\ref{slem:routes}). Its absolute value is at most $\max(\alpha_N, \alpha_C)$ at every $\eta$, and $0$ with a correct check.
\end{sprop}

\noindent\emph{Proof.} (a) By Proposition~\ref{sprop:frontier} a non-conflict item is right if and only if $M > -\kappa$ and a conflict item if and only if $M > \kappa$, with one law of $M$ on both by (A1.3). The difference of the two probabilities is the slope, and $P(M > 0) - P(M > \kappa)$ is the bleeding of Proposition~\ref{sprop:identify}. (b) Lemma~\ref{slem:routes} gives the loss $\alpha_C R_B^C$ on conflict items and the gain $\alpha_N(1 - R_B^N)$ on the others, two terms of opposite sign. \qed

By (a) the conflict accuracy of a shared-weight reader is its worst visible accuracy over the share of conflict items, and a reader with the prior off in view is flat in that share. This is the sense of criterion (ii) of Section~\ref{sec:intro} when a person or a website departs from the usual order more often than in training. By (b) \method stays within its miss shares of its reader at every share. Its prior is a table, so a shift of the order itself is met by new counts, as on held-out skills and after a drift (Section~\ref{sec:s_beyond}), in the manner of prior-shift correction~\cite{saerens2002adjusting,menon2021logit}.

\subsection{Numerical checks}
\label{sec:s_checks}

\begin{table}[t]
\centering
\scriptsize
\setlength{\tabcolsep}{3pt}
\caption{\textbf{Numerical checks.} Route identities, conflict loss and prior predictions hold in every listed case. Selected checks on seen skills and held-out test splits: accuracy band, route identities, conflict loss predictions, episode and library priors. Within: cases meeting each criterion.}
\label{tab:s_ledger}
\begin{tabular}{@{}>{\raggedright\arraybackslash}p{0.56\columnwidth}lr@{}}
\toprule
Check & Kind & Within \\
\midrule
\multicolumn{3}{l}{\textbf{\emph{Theorem~\ref{thm:identify}}}} \\
Conflict accuracy at the $\rho = 0$ end of the band, all and informative cases & prediction & $\TvAdIn/\TvAdN$, $\ThBandInfIn/\ThBandInf$ \\
\midrule
\multicolumn{3}{l}{\textbf{\emph{Theorem~\ref{thm:switch}(a)}}} \\
Lemma~\ref{slem:routes} recomposes every accuracy and conflict loss, a check of the data & identity & $\TvBaIn/\TvBaN$, $\TvBbIn/\TvBbN$ \\
Conflict loss of Lemma~\ref{slem:routes} with every quantity from development & prediction & $\TvBiIn/\TvBiN$ \\
\midrule
\multicolumn{3}{l}{\textbf{\emph{Theorem~\ref{thm:switch}(b)}}} \\
Frame-free prior accuracy given the test items & prediction & $\TvCcIn/\TvCcN$ \\
Held-out skills, prior counted from the library & prediction & $\TvCdIn/\TvCdN$ \\
\bottomrule
\end{tabular}
\end{table}

\begin{figure*}[t]
\centering
\includegraphics[width=\textwidth]{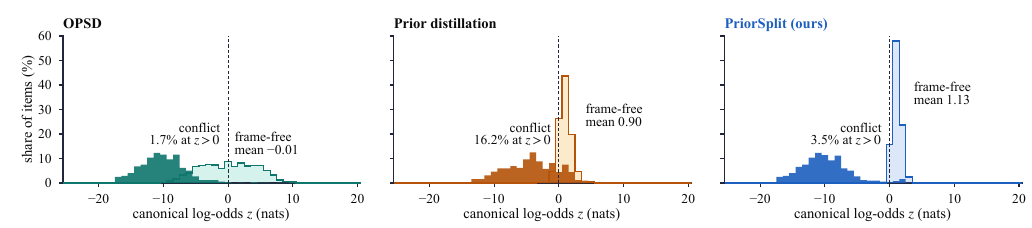}
\caption{\textbf{Where the order is stored.} Only $\OrderConfPastPS\%$ of \method's conflict scores cross zero, against $\OrderConfPastVOne\%$ for prior distillation. Log-odds $z$ on frame-free ($Z_F$, light) and conflict items ($Z_C$, solid), Qwen3-VL-4B.}
\label{fig:s_order_store}
\end{figure*}

Table~\ref{tab:s_ledger} selects checks of the band, route identities and accuracy predictions. Over all checks with a numerical criterion, $\TvAllIn/\TvAllN$ cases fall within it. The conflict loss predictions use development check rates and route accuracies. OPSD stores at most $\ThOpsdPiMax$ nats of frame-free prior per $\delta$ on synthetic skills with Qwen3-VL-4B and 8B, Qwen2.5-VL-7B and Idefics3-8B. As Fig.~\ref{fig:s_order_store} shows on Qwen3-VL-4B, over $\OrderNFree$ frame-free and $\OrderNConf$ conflict items of the synthetic test grid with three seeds and three $\delta$ pooled, the frame-free mean is $\OrderFreeMeanOPSD$ nats for OPSD and $\OrderFreeMeanVOne$ for prior distillation. For \method, $z$ is the score difference of Eq.~(\ref{eq:score}), so its frame-free scores come from the counts and its conflict scores from the OPSD reader wherever its check passes. A conflict item with $z > 0$ is answered with the canonical order. Of the $\ThBandInf$ informative cases, $\ThBandInfIn$ are consistent with the unswitched end $\rho = 0$ and $\ThBandInfAbove$ lie above it, and $\ThBandBelow$ of all $\ThBandAll$ cases fall below the band, every OPCD seed among them. On the main model the conflict loss predicted from development data is $\ThConfLossDevPred$ points against $\ThConfLossMain$ measured, and the held-out library prior is predicted at $\ThLibPred\%$ and measured at $\ThLibMeas\%$.

\section{Implementation details}
\label{sec:s_impl}

\begin{table}[t]
\centering
\scriptsize
\setlength{\tabcolsep}{3pt}
\caption{\textbf{Data per setting.} The usual answer is gold on most order questions. Episodes: split sizes, sampled windows on synthetic skills and Mind2Web. Conflict: full-context departures in view. Usual: canonical or most counted gold (\%). Blank: undefined.}
\label{tab:s_data}
\begin{tabular}{lrrrrrrrr}
\toprule
 & & \multicolumn{3}{c}{Episodes} & \multicolumn{3}{c}{Test questions} & \\
\cmidrule(lr){3-5}\cmidrule(lr){6-8}
Question & Skills & Train & Dev & Test & All & Short & Conflict & Usual \\
\midrule
\multicolumn{9}{l}{\textbf{\emph{Synthetic skills}}} \\
Order & 9 & 1600 & 300 & 1400 & 3200 & 1800 & 173 & 74.1 \\
\midrule
\multicolumn{9}{l}{\textbf{\emph{Mind2Web}}} \\
Order & 97 & 1600 & 300 & 1400 & 3200 & 1800 & 154 & 75.3 \\
Next step & 97 & 1600 & 500 & 700 & 1600 & 900 & 181 & 43.8 \\
\midrule
\multicolumn{9}{l}{\textbf{\emph{Tag-free Mind2Web}}} \\
Order & 96 & 1600 & 300 & 1400 & 2917 & 1626 & 146 & 74.3 \\
Next step & 97 & 1600 & 500 & 700 & 1531 & 864 & 175 & 43.2 \\
\midrule
\multicolumn{9}{l}{\textbf{\emph{HoloAssist}}} \\
Order & 20 & 1243 & 221 & 207 & 3264 & 2448 & 67 & 87.5 \\
Next step & 20 & 1243 & 221 & 207 & 3180 & 2385 & 184 & 71.6 \\
\midrule
\multicolumn{9}{l}{\textbf{\emph{CaptainCook4D}}} \\
Order & 24 & 213 & 62 & 109 & 5544 & 4158 & 344 & 75.2 \\
Next step & 24 & 213 & 62 & 109 & 3480 & 3480 &  & 37.2 \\
\midrule
\multicolumn{9}{l}{\textbf{\emph{ALFWorld}}} \\
Next step & 6 & 1596 & 140 & 134 & 670 &  &  & 99.6 \\
\bottomrule
\end{tabular}
\end{table}

Table~\ref{tab:s_data} counts the skills, episodes and test questions of every setting.

\subsection{Synthetic skills and test grid}
\label{sec:s_synth}

\begin{figure*}[t]
\centering
\includegraphics[width=\textwidth]{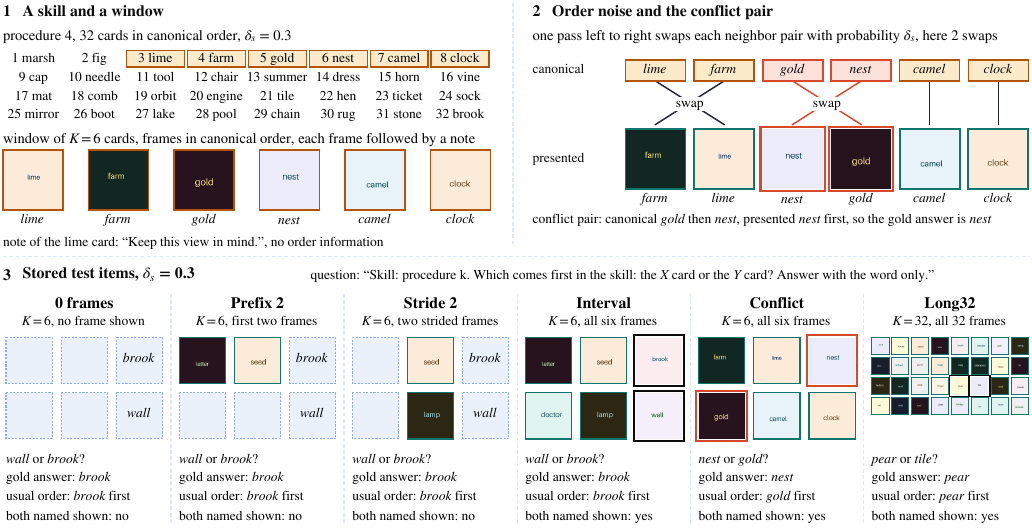}
\caption{\textbf{Construction of the synthetic skills.} One stored test question needs memory under Prefix 2 and Stride 2 and is answered by the frames under Interval. Amber: canonical order, teal: presented frames, coral: conflict pair, dashed: unseen, outlined: named.}
\label{fig:s_synth}
\end{figure*}

Each of nine skills orders $32$ word cards of about $256$ visual tokens, each with a step note free of order information. One pass from left to right through a window of $K$ consecutive units swaps right neighbors with probability $\delta_s$, continuing after each swapped pair (Lemma~\ref{slem:noise}). Adjacent pairs of training windows are canonical with mean probability $0.866$, $0.758$ and $0.642$ for $\delta_s = 0.15$, $0.3$ and $0.5$. Test cells contain $200$ questions per order noise. The short side pools $K=6$ with no frames, the first two (Prefix 2), or two three positions apart (Stride 2), totaling $1800$ questions. The interval cell shows all six ($600$). Long8 and Long32 show full windows with $\delta_s\in\{0.15,0.3\}$ ($400$ each). The sweep in Fig.~\ref{fig:protocol} has $100$ questions per order noise and $m$. Fig.~\ref{fig:s_synth} shows one skill with its order noise, and the first stored test question of each cell at $\delta_s = 0.3$ with the first conflict question of the interval cell.

\subsection{Training}
\label{sec:s_training}

Trained methods start from Qwen3-VL-4B-Instruct with LoRA rank $64$, alpha $128$ and dropout $0.05$. Training uses $200$ steps of $8$ prompts from one episode stream, or one pass over the training pool on HoloAssist ($\HAOrderSteps$ steps). It clips gradients at $0.1$ but not the OPSD divergence, draws one sample per prompt at temperature $1.0$ and top $p=0.95$ with at most $32$ new tokens, and keeps the last checkpoint. Development data set the shared learning rate to $5\times10^{-5}$ against OPSD's default $5\times10^{-6}$. The OPSD teacher has its adapter disabled and receives the full window and privileged order text. The student sees the full window and matches the teacher on its own samples with OPSD's forward KL divergence~\cite{zhao2026opsd}. OPCD diverges at the shared rate and takes $5\times10^{-6}$ from $\{10^{-6},5\times10^{-6}\}$ by best mean short and interval development accuracy. The same criterion chooses $5\times10^{-5}$ for CW-OPD from its prescribed $2\times10^{-6}$ and the shared rate.

\subsection{PriorSplit}
\label{sec:s_priorsplit}

\begin{algorithm}[t]
\caption{\method: Count, Check and Switch}
\label{alg:priorsplit}
\begin{algorithmic}[1]
\Require base model $\theta_0$, consumed episodes $\mathcal{D}$, question $q$ on $X, Y$ of skill $s$, context $V$
\State $E \gets \mathrm{OPSD}(\theta_0, \mathcal{D})$
\State $n \gets$ number of episodes of $s$ in $\mathcal{D}$ that hold $X, Y$
\State $n_{XY} \gets$ number of those with $X$ before $Y$
\State $\bar p_s(X \text{ first}) \gets \mathrm{clip}\big((n_{XY} + 1)/(n + 2)\big)$
\For{each $u \in \{X, Y\}$}
  \State $\ell_u \gets \log p_E(\text{yes} \mid u, V) - \log p_E(\text{no} \mid u, V)$
\EndFor
\State $g \gets \mathbf{1}[\ell_X > 0]\,\mathbf{1}[\ell_Y > 0]$
\If{$g = 1$}
  \State \Return $\argmax_{y} \log p_E(y \mid q, V)$
\Else
  \State \Return $\argmax_{y} \log \bar p_s(y)$
\EndIf
\end{algorithmic}
\end{algorithm}

\begin{table}[t]
\centering
\scriptsize
\setlength{\tabcolsep}{3pt}
\caption{\textbf{Check wordings.} The reader itself answers them, and no label trains the check. Asked once per named unit (A) or next-step question (B). Every prompt ends with ``Answer yes or no.'' Italic words are filled per question.}
\label{tab:s_wordings}
\begin{tabular}{@{}>{\raggedright\arraybackslash}p{0.19\columnwidth}>{\raggedright\arraybackslash}p{0.375\columnwidth}>{\raggedright\arraybackslash}p{0.375\columnwidth}@{}}
\toprule
Setting & Question A (named unit) & Question B (next step) \\
\midrule
Synthetic skills & Skill: \emph{name}. Is an image labelled \emph{word} among the images above? & none \\
Mind2Web, CaptainCook4D, HoloAssist (tag banner) & Skill: \emph{name}. Is a step tagged \emph{tag} among the steps above? & Skill: \emph{name}. Is the step tagged \emph{tag} followed by another step among the steps above? \\
Tag-free Mind2Web & Skill: \emph{name}. Is a screenshot of the step that performs `\emph{text}' among the images above? & Skill: \emph{name}. Is the step that performs `\emph{text}' followed by another step among the images above? \\
\bottomrule
\end{tabular}
\end{table}

Algorithm~\ref{alg:priorsplit} lists the steps for an order question. The reader $E$ is the same-seed OPSD model, an option's score is the log-probability of the option word as the first answer token, and the counts of Eq.~(\ref{eq:prior}) come from the episodes the reader consumed, clipped to $[0.02, 0.98]$, with $\bar p_s = \tfrac12$ for a pair without counts. The check asks question A of Table~\ref{tab:s_wordings} per named unit and question B per next-step question, then compares the summed first-token probabilities of yes, Yes and YES with those of no, No and NO, each also with a leading space. No label trains it or sets its threshold. Development data choose only the wording on tag-free Mind2Web, the tag banner on HoloAssist, the tag list of CaptainCook4D and the number of next-step check questions. On HoloAssist a question without frames goes to the counted prior. A key-value cache of the shared part of the three prompts lowers the two checks from $\KVCheckRecomp$ to $\KVCheckShared$ answer passes per question on $\KVNQ$ synthetic test questions and from $\KVCheckRecompLong$ to $\KVCheckSharedLong$ on the window of $32$ frames, with every check decision kept. The counts-in-prompt ablation prefixes the question with ``In this skill, \emph{A} usually comes before \emph{B}.'', \emph{A} first in the counts.

\subsection{Baselines}
\label{sec:s_baselines}

Every trained method runs in our trainer with the OPSD loss of the official code and shares with OPSD the teacher, the episodes, the adapter and, except all-pair subsequence OPSD (5$\times$ steps), the number of steps. Subsequence OPSD and both SFT variants show the student a random order-preserving subsequence of the window with $m$ uniform in $\{0, \dots, K\}$. All-pair subsequence OPSD asks every in-window adjacent pair of the same episodes, with five times the steps or within $1600$ prompts (same steps). Prior distillation adds to subsequence OPSD, with weight $1$, a distillation of the counted order into the frame-free answer. Context distillation~\cite{snell2022context}, OPCD~\cite{opcd}, the SKILL0-style curriculum~\cite{skill0} and NCA~\cite{nca} with weight $0.5$ follow their papers. The CW-OPD-style baseline~\cite{cwopd} swaps the named pair to form its second world and applies it, with weight $1$, only where both named units are visible. CAD~\cite{shi2024cad} with $\gamma = 0.5$, PMI~\cite{holtzman2021surface} (its limit in $\gamma$) and PAD~\cite{pad2024} with $\gamma = 1$ run on the OPSD reader. The base model, the all-pair and 5$\times$ steps variants, the oracle reference and the counts-in-prompt ablation are not ranked. \method uses the stored option scores and checks of OPSD with the same seed. Test episodes are separate from training and development, reusing training cards or screenshots on synthetic skills and Mind2Web. The paired $95\%$ bootstrap fixes trained seeds and resamples questions, recordings, sessions or episodes across methods. All-pair rows exclude questions either variant trained on. Results average three seeds, and differences use unrounded values.

\begin{table*}[t]
\centering
\scriptsize
\setlength{\tabcolsep}{2.8pt}
\caption{\textbf{Generality.} PriorSplit leads on Short throughout. Seen Mind2Web websites (order, Qwen3-VL-4B) and synthetic skills, seed means (\%). Bold, underline: best, second best per column and group. Shaded: PriorSplit. Italic: not ranked, all-pair rows on untrained questions only.}
\label{tab:general}
\begin{tabular}{>{\raggedright\arraybackslash}p{2.35cm}rrrrrrrrrrrrrrr}
\toprule
 & \multicolumn{3}{c}{Mind2Web} & \multicolumn{3}{c}{Qwen2.5-VL-7B} & \multicolumn{3}{c}{Qwen3-VL-8B} & \multicolumn{3}{c}{Idefics3-8B} & \multicolumn{3}{c}{InternVL3.5-8B} \\
\cmidrule(lr){2-4} \cmidrule(lr){5-7} \cmidrule(lr){8-10} \cmidrule(lr){11-13} \cmidrule(lr){14-16}
Method & Short & Interval & Conflict & Short & Interval & Conflict & Short & Interval & Conflict & Short & Interval & Conflict & Short & Interval & Conflict \\
\midrule
OPSD~\cite{zhao2026opsd} & 58.6 & 99.7 & \textbf{99.4} & 53.2 & 99.3 & \underline{98.5} & 52.2 & 98.2 & \textbf{98.5} & 54.5 & 97.8 & \underline{98.1} & 51.8 & 99.6 & \underline{99.0} \\
Subsequence OPSD & 54.3 & 98.4 & 97.0 & 54.9 & 98.1 & 97.1 & 55.8 & 95.6 & 94.2 & 55.3 & 91.3 & 87.3 & 56.7 & 99.9 & \textbf{99.8} \\
NCA~\cite{nca} & 55.1 & 94.3 & 90.9 & 54.5 & 92.3 & 87.1 & 56.0 & 85.6 & 75.9 & 56.6 & 80.3 & 70.1 & 55.1 & 95.4 & 91.1 \\
Prior distillation & \underline{61.4} & 99.4 & \underline{98.9} & \underline{69.0} & 99.8 & \textbf{99.6} & \underline{67.4} & 94.1 & 87.1 &  &  &  & \underline{69.7} & 99.2 & 97.9 \\
\grayed{All-pair subsequence OPSD (5$\times$ steps)} & \grayed{61.0} & \grayed{98.8} & \grayed{97.4} & \grayed{64.5} & \grayed{98.6} & \grayed{96.1} & \grayed{64.4} & \grayed{97.0} & \grayed{93.1} & \grayed{61.9} & \grayed{91.4} & \grayed{81.5} & \grayed{65.4} & \grayed{99.9} & \grayed{99.8} \\
PAD~\cite{pad2024} & 57.6 & 99.3 & \textbf{99.4} & 54.2 & 98.5 & 97.7 & 52.3 & 98.2 & \textbf{98.5} & \underline{58.4} & 98.3 & \textbf{99.0} & 51.1 & 99.3 & 98.5 \\
\hlours \textbf{PriorSplit} & \textbf{71.9} & 99.1 & 97.6 & \textbf{74.0} & 98.6 & 96.1 & \textbf{73.1} & 98.3 & \underline{97.3} & \textbf{73.9} & 93.6 & 82.1 & \textbf{74.0} & 99.6 & \underline{99.0} \\
\midrule
\grayed{Oracle reference (Def. 1)} & \grayed{76.7} & \grayed{100.0} & \grayed{100.0} & \grayed{73.8} & \grayed{100.0} & \grayed{100.0} & \grayed{73.8} & \grayed{100.0} & \grayed{100.0} & \grayed{73.8} & \grayed{100.0} & \grayed{100.0} & \grayed{73.8} & \grayed{100.0} & \grayed{100.0} \\
\bottomrule
\end{tabular}
\end{table*}

\begin{table}[t]
\centering
\scriptsize
\setlength{\tabcolsep}{4pt}
\caption{\textbf{Next step.} PriorSplit leads on Short. Next-step questions on seen Mind2Web websites (which of four actions comes next), accuracy (\%) over three training seeds.}
\label{tab:s_m2w_next}
\begin{tabular}{lrrr}
\toprule
Method & Short & Interval & Conflict \\
\midrule
OPSD & \underline{29.1} & \textbf{72.0} & \textbf{67.0} \\
Subsequence OPSD & 26.1 & 63.4 & 59.3 \\
NCA & 24.9 & 59.2 & 54.1 \\
Prior distillation & 26.9 & 66.6 & 60.4 \\
\hlours \textbf{PriorSplit} & \textbf{50.4} & \underline{71.1} & \underline{65.4} \\
\bottomrule
\end{tabular}
\end{table}

\subsection{Mind2Web skills and other backbones}
\label{sec:s_m2w}

Training uses $\MTraj$ Mind2Web trajectories from $\MWeb$ websites, among complete annotations of $4$ to $40$ steps. A skill is one trajectory of Multimodal-Mind2Web~\cite{zheng2024seeact}, with a median of eight steps, and a unit is a step: its screenshot with a top-left banner carrying a short tag, as on HoloAssist and CaptainCook4D keyframes, and its action text, such as \texttt{CLICK the button "Filter"}. A next-step question names the current step by its tag and offers four action texts of the trajectory, the gold being the direct successor of the current step in the presented order. The prior counts with Laplace smoothing how often each option's step directly follows the current one in the consumed episodes. The check asks questions A and B and sets $g=1$ when it finds the named step and another shown step after it. The further backbones see about $256$ visual tokens per frame ($169$ for Idefics3-8B) and use the recipe and learning rate of the main model.

\begin{figure}[t]
\centering
\includegraphics[width=0.86\columnwidth]{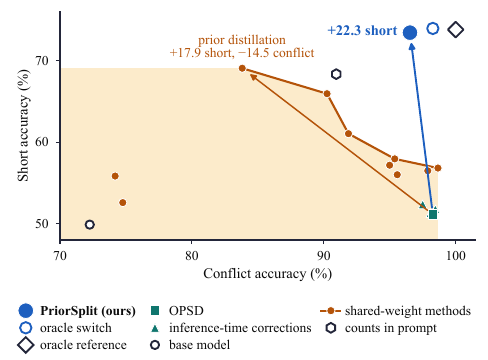}
\caption{\textbf{Short versus conflict accuracy.} PriorSplit is the only ranked method above the frontier. Seen synthetic skills, seed means, arrows from OPSD. Off the axis: context distillation (42.6, 59.9) and counts in prompt with the base model (9.1, 70.0).}
\label{fig:tradeoff_full}
\end{figure}
\begin{table*}[t]
\centering
\scriptsize
\setlength{\tabcolsep}{4pt}
\caption{\textbf{Ablation.} On Short the switch beats both parts and every unchecked combination. Seen synthetic skills, main model, three training seeds. Oracle switch: true visibility. Reader plus prior, checked, uses option scores. Reader ratio plus prior: every check set to yes.}
\label{tab:s_ablation}
\begin{tabular}{lrrrrrrrr}
\toprule
Variant & Short & 0 frames & Prefix 2 & Stride 2 & Interval & Conflict & Long8 & Long32 \\
\midrule
\hlours \textbf{PriorSplit (switch)} & \textbf{73.5} & \textbf{71.3} & \textbf{78.5} & \underline{70.6} & 98.6 & 96.5 & \underline{97.4} & 83.0 \\
\grayed{Oracle switch} & \grayed{74.0} & \grayed{71.3} & \grayed{79.3} & \grayed{71.3} & \grayed{98.8} & \grayed{98.3} & \grayed{97.2} & \grayed{80.3} \\
Reader only (OPSD) & 51.1 & 45.8 & 57.7 & 49.9 & \textbf{98.8} & \textbf{98.3} & 97.2 & 80.3 \\
Prior only & 71.3 & \textbf{71.3} & 71.3 & \textbf{71.3} & 71.3 & 1.5 & 81.0 & \textbf{85.0} \\
Reader plus prior & 54.7 & \underline{50.4} & 61.2 & 52.4 & \underline{98.7} & \underline{97.7} & \textbf{97.6} & 82.5 \\
Reader ratio plus prior, checked & \underline{73.4} & \textbf{71.3} & \underline{78.4} & \underline{70.6} & 98.3 & 94.8 & 96.0 & 80.8 \\
Reader plus prior, checked & \textbf{73.5} & \textbf{71.3} & \underline{78.4} & \underline{70.6} & 98.6 & 96.0 & \textbf{97.6} & \underline{84.8} \\
Reader ratio plus prior & 63.3 & \textbf{71.3} & 64.9 & 53.7 & 98.4 & 96.1 & 95.9 & 78.3 \\
\midrule
\grayed{PriorSplit with the base model as reader} & \grayed{73.3} & \grayed{71.3} & \grayed{77.3} & \grayed{71.2} & \grayed{82.0} & \grayed{52.6} & \grayed{77.8} & \grayed{70.3} \\
\grayed{PriorSplit with SFT on gold answers as reader} & \grayed{73.6} & \grayed{71.2} & \grayed{78.3} & \grayed{71.2} & \grayed{97.6} & \grayed{93.6} & \grayed{98.3} & \grayed{85.1} \\
\midrule
\multicolumn{9}{l}{\textbf{\emph{Counts in the prompt}}} \\
\grayed{OPSD reader} & \grayed{68.4} & \grayed{71.3} & \grayed{73.8} & \grayed{59.9} & \grayed{97.2} & \grayed{90.9} & \grayed{97.1} & \grayed{84.6} \\
\grayed{Base model} & \grayed{70.0} & \grayed{71.3} & \grayed{73.4} & \grayed{65.3} & \grayed{72.5} & \grayed{9.1} & \grayed{81.0} & \grayed{83.3} \\
\bottomrule
\end{tabular}
\end{table*}

\begin{table*}[t]
\centering
\scriptsize
\setlength{\tabcolsep}{3pt}
\caption{\textbf{Own check.} PriorSplit gains on Short throughout. Seen skills and websites, seed means. PriorSplit minus OPSD (points), self: own check, true: true visibility (0.0 on Interval and Conflict, not shown). Miss conflict: conflict items handed to the prior (\%).}
\label{tab:s_gate}
\begin{tabular}{lrrrrrrr}
\toprule
Setting & Miss & False yes & Miss conflict & Short self & Short true & Interval & Conflict \\
\midrule
Qwen3-VL-4B & 5.3 & 2.1 & 2.3 & 22.3 & 22.8 & $-$0.2 & $-$1.7 \\
Mind2Web & 3.6 & 0.5 & 2.2 & 13.3 & 13.8 & $-$0.6 & $-$1.7 \\
Qwen2.5-VL-7B & 2.3 & 0.1 & 2.3 & 20.8 & 20.7 & $-$0.7 & $-$2.3 \\
Qwen3-VL-8B & 2.6 & 3.6 & 1.3 & 20.9 & 21.7 & 0.1 & $-$1.2 \\
Idefics3-8B & 25.0 & 0.1 & 16.2 & 19.4 & 19.5 & $-$4.3 & $-$16.0 \\
InternVL3.5-8B & 0.0 & 0.1 & 0.0 & 22.2 & 22.2 & 0.0 & 0.0 \\
\bottomrule
\end{tabular}
\end{table*}

\subsection{CaptainCook4D}
\label{sec:s_cc4d}

A skill is one of the $24$ recipes of CaptainCook4D~\cite{captaincook4d}, and a unit is an annotated step, shown by the middle frame of its segment with the tag banner and a description without order information. The canonical order follows the recipe's task graph, with the mean position over training recordings where the graph leaves it open, and the presented order is the cook's. We use the $213$ training, $\CDevNClusters$ development and $\CNClusters$ test recordings of the official split, six-step windows with whole recordings as longer contexts, and the synthetic tag words without $47$ for things seen in kitchens. Three OPSD readers are compared with prior distillation and PAD, and next-step questions offer four tag words. On development recordings the check misses a step on $\CGoneMissMin\%$ to $\CGoneMissMax\%$ of visible full-window items, above the $5\%$ bound (Section~\ref{sec:s_probe}) on $\CGoneNFail$ seeds.

\subsection{Tag-free Mind2Web}
\label{sec:s_tagfree}

Tag-free Mind2Web drops the banner and names each step by its action text, which stays as its note, with OPSD, prior distillation and \method as on tagged Mind2Web. Its test cells are those of tagged Mind2Web and hold the $\TFOrdNCells$ order and $\TFNxtNCells$ next-step questions whose named steps have action texts unique in their skill.

\noindent\phantomsection\label{sec:s_heldout_m2w_setup}\textbf{Held-out skills.} Two splits test skills that no reader trained on. Held-out trajectories of seen websites hold out one seeded trajectory of each of the $\HOTNSites$ websites with two or more trajectories, and six new readers train on the other $\HOTNTrain$ of all $\HOTNAll$ trajectories as above. Held-out websites are the $\HOSNSites$ websites of the $\HOSNSkills$ trajectories that the readers of the other tag-free tests never saw. The counts of the held-out skills come from $\HOTNConsume$ of their $\HOTNPool$ logged executions and $\HOSNConsume$ of $\HOSNPool$, the same share as a training run takes of its pool, and no test question repeats a logged one. Each split asks $\HOTNQ$ order questions, each with no frame, two prefix or strided frames, and every frame of a six-step window, which ends early with its trajectory. The probe is fitted on training trajectories only.

\begin{table*}[t]
\centering
\scriptsize
\setlength{\tabcolsep}{2pt}
\caption{\textbf{Held-out skills.} PriorSplit leads OPSD without frames and matches it with frames. Unranked controls, main model, seed mean accuracy (\%). PriorSplit and the counted prior alone (78.0\% in every cell) count those skills' episodes. Gain: points [95\% interval].}
\label{tab:s_heldout}
\begin{tabular}{lrrrrrrrrrrl}
\toprule
 &  &  & \multicolumn{2}{c}{200 steps} & \multicolumn{2}{c}{5$\times$ steps} &  & \multicolumn{2}{c}{Counts in prompt} &  &  \\
\cmidrule(lr){4-5}\cmidrule(lr){6-7}\cmidrule(lr){9-10}
Cell & Questions & OPSD & \shortstack{Subsequence\\OPSD} & \shortstack{Prior\\distillation} & \shortstack{Subsequence\\OPSD} & \shortstack{Prior\\distillation} & PriorSplit & \shortstack{OPSD\\reader} & \shortstack{Base\\model} & \shortstack{Oracle\\reference} & Gain over OPSD \\
\midrule
$m=0$, $\delta=0.3$ & 200 & 51.2 & 51.8 & 53.3 & 50.0 & 50.5 & 78.0 & 78.5 & 77.0 & 78.5 & 26.8 [18.7, 34.7] \\
$m=6$, $\delta=0.3$ & 200 & 98.2 & 91.3 & 89.2 & 91.7 & 90.2 & 98.2 & 97.7 & 80.0 & 100.0 & 0.0 [$-$1.0, 1.0] \\
$m=32$, $\delta=0.3$ & 100 & 79.0 & 72.3 & 62.7 & 72.3 & 60.0 & 79.3 & 81.0 & 76.0 & 100.0 & 0.3 [$-$2.0, 2.7] \\
\bottomrule
\end{tabular}
\end{table*}

\begin{table*}[t]
\centering
\scriptsize
\setlength{\tabcolsep}{3pt}
\caption{\textbf{Learned prior.} Within PriorSplit, counts match a learned prior on seen skills and beat it on held-out skills. Unranked controls, main model. Learned prior: adapter fitted to training counts, queried without frames. Seed means and differences [95\% interval].}
\label{tab:s_learned_prior}
\begin{tabular}{lrrrrr}
\toprule
Method & Short & 0 frames & Interval & Conflict & Held-out, 0 frames \\
\midrule
OPSD & 51.1 & 45.8 & 98.8 & 98.3 & 51.2 \\
Prior only (counts) & 71.3 & 71.3 & 71.3 & 1.5 & 78.0 \\
Learned prior only & 70.6 & 70.6 & 70.6 & 8.3 & 48.7 \\
\hlours \textbf{PriorSplit} & 73.5 & 71.3 & 98.6 & 96.5 & 78.0 \\
PriorSplit, learned prior & 72.8 & 70.6 & 98.6 & 96.5 & 48.7 \\
\midrule
Learned minus counted prior & $-$0.7 [$-$1.9, 0.5] & $-$0.7 [$-$1.9, 0.5] & $-$0.7 [$-$1.9, 0.5] & 6.7 [4.4, 9.2] & $-$29.3 [$-$37.7, $-$20.8] \\
PriorSplit, learned minus counted & $-$0.6 [$-$1.8, 0.5] & $-$0.7 [$-$2.0, 0.5] & 0.0 [0.0, 0.0] & 0.0 [0.0, 0.0] & $-$29.3 [$-$37.8, $-$20.8] \\
\bottomrule
\end{tabular}
\end{table*}

\begin{table*}[t]
\centering
\scriptsize
\setlength{\tabcolsep}{2.5pt}
\caption{\textbf{Training budget.} PriorSplit leads both 5$\times$ rows on Interval and Conflict. Unranked controls, main model. Steps: 999 to 1007 versus 200. Prior distillation draws: 7992 to 8056 versus 1600, about 4.0 per training episode. Seed means and differences [95\% interval].}
\label{tab:s_budget}
\begin{tabular}{>{\raggedright\arraybackslash}p{3.9cm}rrrrrr}
\toprule
Method & Short & Interval & Conflict & Long8 & Long32 & Held-out, 0 frames \\
\midrule
OPSD & 51.1 & 98.8 & 98.3 & 97.2 & 80.3 & 51.2 \\
Subsequence OPSD & 56.0 & 96.2 & 95.6 & 94.6 & 76.4 & 51.8 \\
Subsequence OPSD (5$\times$ steps) & 63.5 & 94.4 & 89.0 & 93.2 & 83.3 & 50.0 \\
Prior distillation & 69.1 & 92.7 & 83.8 & 91.3 & 75.4 & 53.3 \\
Prior distillation (5$\times$ steps) & 72.6 & 93.8 & 85.4 & 89.8 & 75.9 & 50.5 \\
\hlours \textbf{PriorSplit} & 73.5 & 98.6 & 96.5 & 97.4 & 83.0 & 78.0 \\
\midrule
Subsequence OPSD (5$\times$ steps) minus OPSD & 12.4 [9.4, 15.4] & $-$4.3 [$-$5.6, $-$3.2] & $-$9.2 [$-$12.5, $-$6.2] & $-$4.0 [$-$5.6, $-$2.5] & 3.1 [$-$0.1, 6.3] & $-$1.2 [$-$8.5, 6.3] \\
Prior distillation (5$\times$ steps) minus OPSD & 21.4 [17.9, 24.7] & $-$5.0 [$-$6.5, $-$3.6] & $-$12.9 [$-$17.3, $-$8.9] & $-$7.4 [$-$9.7, $-$5.3] & $-$4.3 [$-$7.6, $-$1.1] & $-$0.7 [$-$8.7, 7.2] \\
PriorSplit minus subsequence OPSD (5$\times$ steps) & 10.0 [7.1, 12.8] & 4.2 [3.1, 5.3] & 7.5 [4.8, 10.6] &  &  &  \\
PriorSplit minus prior distillation (5$\times$ steps) & 0.9 [$-$0.1, 1.9] & 4.8 [3.4, 6.3] & 11.2 [7.3, 15.6] &  &  &  \\
\bottomrule
\end{tabular}
\end{table*}

\begin{table}[t]
\centering
\scriptsize
\setlength{\tabcolsep}{3pt}
\caption{\textbf{Held-out Mind2Web skills without tags.} PriorSplit with the probe leads on the short side and matches OPSD with all frames. Seed mean accuracy (\%), three training seeds, 550 questions per cell. Italic: oracle switch, true visibility.}
\label{tab:s_heldout_m2w}
\begin{tabular}{lrrrrrr}
\toprule
Method & Short & 0 frames & Prefix 2 & Stride 2 & Interval & Conflict \\
\midrule
\multicolumn{7}{l}{\textbf{\emph{Held-out trajectories of seen websites (35 trajectories)}}} \\
OPSD & \underline{61.9} & \underline{59.0} & \underline{69.3} & \underline{57.4} & \textbf{94.0} & \textbf{88.6} \\
Prior distillation & 59.2 & 58.2 & 62.9 & 56.3 & \underline{92.8} & \underline{87.0} \\
\hlours \textbf{PriorSplit} & \textbf{73.4} & \textbf{71.2} & \textbf{78.4} & \textbf{70.5} & \textbf{94.0} & \textbf{88.6} \\
\grayed{Oracle switch} & \grayed{73.8} & \grayed{71.2} & \grayed{79.2} & \grayed{71.2} & \grayed{94.0} & \grayed{88.6} \\
\midrule
\multicolumn{7}{l}{\textbf{\emph{Held-out websites (25 trajectories)}}} \\
OPSD & \underline{60.7} & 53.6 & \underline{69.1} & \underline{59.4} & \underline{94.9} & \underline{87.4} \\
Prior distillation & 59.1 & \underline{56.2} & 63.2 & 57.9 & \textbf{97.3} & \textbf{95.7} \\
\hlours \textbf{PriorSplit} & \textbf{72.9} & \textbf{70.2} & \textbf{78.1} & \textbf{70.3} & \underline{94.9} & \underline{87.4} \\
\grayed{Oracle switch} & \grayed{73.1} & \grayed{70.2} & \grayed{78.7} & \grayed{70.4} & \grayed{94.9} & \grayed{87.4} \\
\bottomrule
\end{tabular}
\end{table}

\subsection{HoloAssist}
\label{sec:s_holoassist}

Skills are the $20$ tasks of HoloAssist~\cite{holoassist}, and a unit is a coarse action at its first occurrence in a session, shown by the middle frame of its first segment with the tag banner. The canonical order sorts units by mean relative position over training sessions, and questions pair units adjacent in it among those a session performs. The test split is the $207$ official validation sessions, sharing tasks, participants and recording days with training, and the development split $15\%$ of each task's training sessions. Short contexts show no frame or the last two or four steps before a cut. Three OPSD readers each of Qwen3-VL-4B and 8B train in one pass and are compared with prior distillation and PAD. With 4B readers OPSD and prior distillation are also trained on both question types for $\HANSSteps$ steps instead of $\HAOrderSteps$. The order departures are the $\HANConfOrder$ full-context test departures that two independent annotation passes kept, and the next-step departures the $\HANConfNext$ full-context questions whose shown next step is not among the most counted. The order check hands $\HADevOrderRouteLo\%$ and $\HADevOrderRouteHi\%$ of full development contexts to the prior on two 4B seeds and $\HAEightDevOrderRoute\%$ on one 8B seed, above the $5\%$ bound.

\begin{figure*}[t]
\centering
\includegraphics[width=0.82\textwidth]{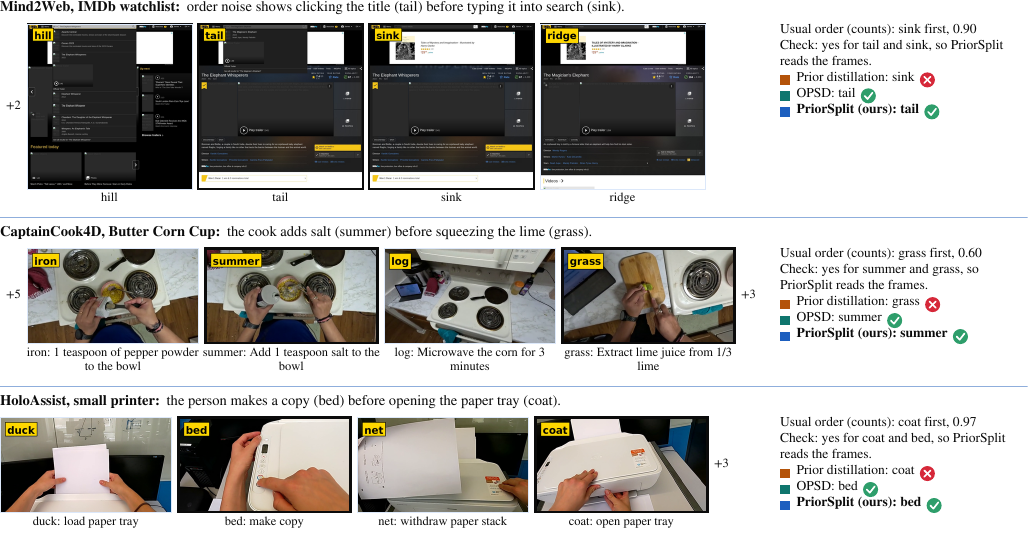}
\caption{\textbf{Real examples.} In each row PriorSplit follows frames that contradict the usual order. Selected test questions on seen skills, one per dataset, picked to favor PriorSplit. Named steps outlined, +n counts omitted frames. Qwen3-VL-4B readers, seed $0$.}
\label{fig:s_real}
\end{figure*}

\begin{figure*}[t]
\centering
\includegraphics[width=\textwidth]{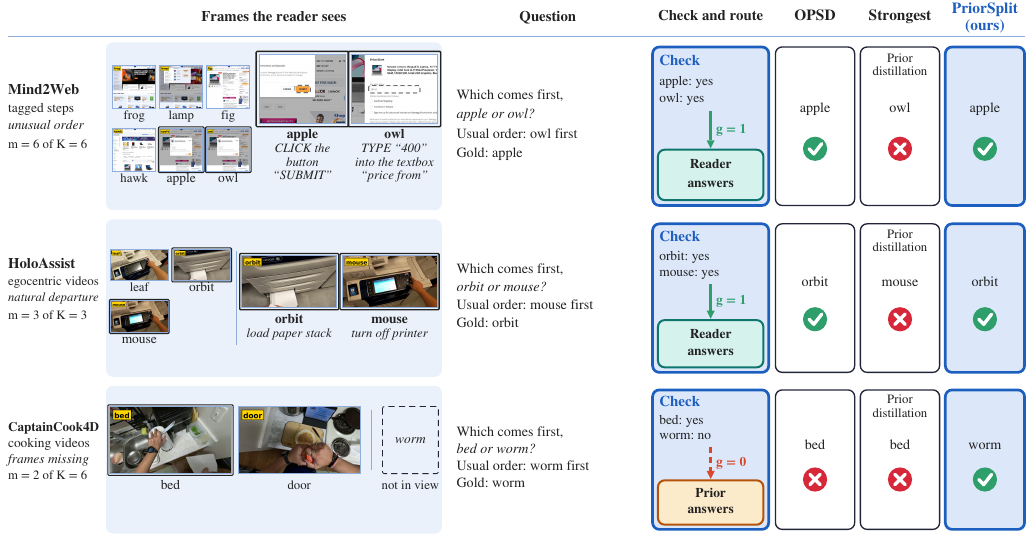}
\caption{\textbf{The other situation of each dataset.} PriorSplit is right in every row and the strongest compared method (Strongest) in none. Rows 1 and 2 show an unusual order, row 3 a missing frame. Selected examples, Qwen3-VL-4B, seed $0$.}
\label{fig:s_qualitative}
\end{figure*}

\begin{table*}[t]
\centering
\scriptsize
\setlength{\tabcolsep}{3pt}
\caption{\textbf{Strongest method.} PriorSplit leads on Short throughout the first two groups. Strongest ranked method by Short, seen skills and real test splits, points [95\% interval]. Blank: unpaired. Italic: HoloAssist next step. (both): training on both question types.}
\label{tab:s_best}
\begin{tabular}{llrrlll}
\toprule
& & \multicolumn{2}{c}{Short accuracy} & \multicolumn{3}{c}{PriorSplit minus the strongest method} \\
\cmidrule(lr){3-4} \cmidrule(lr){5-7}
Setting & Strongest method & Strongest & PriorSplit & Short & Interval & Conflict \\
\midrule
Qwen3-VL-4B & Prior distillation & 69.1 & 73.5 & 4.4 [2.4, 6.4] & 5.9 [4.6, 7.3] & 12.7 [9.4, 16.2] \\
Qwen2.5-VL-7B & Prior distillation & 69.0 & 74.0 & 5.0 [3.2, 6.9] & \ensuremath{-}1.2 [\ensuremath{-}2.1, \ensuremath{-}0.4] & \ensuremath{-}3.5 [\ensuremath{-}6.2, \ensuremath{-}1.2] \\
Qwen3-VL-8B & Prior distillation & 67.4 & 73.1 & 5.7 [3.5, 7.9] & 4.2 [2.9, 5.6] & 10.2 [6.9, 13.7] \\
Idefics3-8B & PAD & 58.4 & 73.9 & 15.6 [11.6, 19.4] & \ensuremath{-}4.7 [\ensuremath{-}5.9, \ensuremath{-}3.6] & \ensuremath{-}17.0 [\ensuremath{-}19.8, \ensuremath{-}14.1] \\
InternVL3.5-8B & Prior distillation & 69.7 & 74.0 & 4.2 [2.5, 6.1] & 0.4 [\ensuremath{-}0.1, 0.8] & 1.2 [0.0, 2.5] \\
Mind2Web order & Prior distillation & 61.4 & 71.9 & 10.4 [7.3, 13.5] & \ensuremath{-}0.3 [\ensuremath{-}1.0, 0.3] & \ensuremath{-}1.3 [\ensuremath{-}3.5, 0.6] \\
Mind2Web order, 8B & Prior distillation & 60.1 & 72.0 & 11.9 [8.8, 15.0] & \ensuremath{-}0.7 [\ensuremath{-}1.1, \ensuremath{-}0.3] & \ensuremath{-}1.3 [\ensuremath{-}2.6, \ensuremath{-}0.2] \\
Mind2Web next step & OPSD & 29.1 & 50.4 & 21.3 [16.6, 25.8] & \ensuremath{-}0.9 [\ensuremath{-}2.1, 0.2] & \ensuremath{-}1.7 [\ensuremath{-}3.5, 0.0] \\
\midrule
HoloAssist order & Prior distillation & 86.2 & 90.1 & 3.9 [2.6, 5.3] & \ensuremath{-}4.8 [\ensuremath{-}6.7, \ensuremath{-}3.0] &  \\
HoloAssist order, 8B & Prior distillation & 86.1 & 90.5 & 4.4 [2.7, 6.1] & \ensuremath{-}0.3 [\ensuremath{-}1.2, 0.6] &  \\
CaptainCook4D, test & Prior distillation & 69.5 & 75.0 & 5.5 [4.0, 7.0] & \ensuremath{-}2.3 [\ensuremath{-}3.3, \ensuremath{-}1.3] & \ensuremath{-}5.1 [\ensuremath{-}7.2, \ensuremath{-}3.2] \\
CaptainCook4D next step & Prior distillation & 30.0 & 44.4 & 14.3 [11.1, 17.6] &  &  \\
Tag-free Mind2Web order & OPSD & 62.5 & 69.8 & 7.3 [4.1, 10.5] & \ensuremath{-}2.6 [\ensuremath{-}3.9, \ensuremath{-}1.4] & \ensuremath{-}7.8 [\ensuremath{-}11.4, \ensuremath{-}4.6] \\
Tag-free Mind2Web next step & Prior distillation & 35.9 & 48.2 & 12.3 [8.2, 16.6] & \ensuremath{-}7.4 [\ensuremath{-}11.2, \ensuremath{-}3.7] & \ensuremath{-}15.4 [\ensuremath{-}20.6, \ensuremath{-}10.3] \\
\midrule
\grayed{HoloAssist next step (both)} & \grayed{Prior distillation (both)} & \grayed{71.4} & \grayed{69.1} & \grayed{\ensuremath{-}2.4 [\ensuremath{-}3.9, \ensuremath{-}0.9]} &  &  \\
\grayed{HoloAssist next step, 8B} & \grayed{Prior distillation} & \grayed{31.5} & \grayed{66.7} & \grayed{35.2 [31.9, 38.4]} &  &  \\
\bottomrule
\end{tabular}
\end{table*}

\begin{table*}[t]
\centering
\scriptsize
\setlength{\tabcolsep}{4pt}
\caption{\textbf{Shown frames.} PriorSplit leads OPSD up to two frames and every trained row from twelve. Seed mean accuracy (\%) by shown frames $m$, seen skills, main model. Window: six frames for $m\le 2$, $m$ frames for $m\ge 4$. Differences in points.}
\label{tab:s_protocol}
\begin{tabular}{lrrrrrrrrrr}
\toprule
Method & $m=0$ & $m=1$ & $m=2$ & $m=4$ & $m=6$ & $m=8$ & $m=12$ & $m=16$ & $m=24$ & $m=32$ \\
\midrule
OPSD & 44.8 & 50.7 & 53.8 & 99.8 & 98.4 & 96.4 & 92.1 & 87.8 & 84.7 & 80.0 \\
Subsequence OPSD & 53.7 & 52.9 & 56.3 & 98.4 & 96.4 & 94.9 & 92.0 & 87.1 & 82.6 & 76.1 \\
Prior distillation & 69.4 & 69.2 & 69.9 & 96.1 & 93.3 & 89.8 & 84.9 & 80.1 & 78.3 & 75.4 \\
Subsequence OPSD (5$\times$ steps) & 61.1 & 61.2 & 62.2 & 95.7 & 94.1 & 93.1 & 91.0 & 88.2 & 81.9 & 79.2 \\
Prior distillation (5$\times$ steps) & 72.2 & 72.2 & 73.3 & 96.8 & 93.9 & 89.1 & 82.8 & 78.0 & 76.6 & 73.1 \\
\hlours \textbf{PriorSplit} & 73.7 & 72.4 & 72.7 & 99.7 & 98.7 & 96.1 & 94.2 & 89.3 & 87.0 & 82.7 \\
Prior only & 73.7 & 73.7 & 73.7 & 75.9 & 73.7 & 76.3 & 76.8 & 82.1 & 76.9 & 75.6 \\
Oracle reference (Def. 1) & 73.3 & 73.3 & 73.3 & 100.0 & 100.0 & 100.0 & 100.0 & 100.0 & 100.0 & 100.0 \\
\midrule
Oracle reference minus PriorSplit & $-$0.3 & 0.9 & 0.7 & 0.3 & 1.3 & 3.9 & 5.8 & 10.7 & 13.0 & 17.3 \\
OPSD minus PriorSplit & $-$28.9 & $-$21.8 & $-$18.9 & 0.1 & $-$0.2 & 0.3 & $-$2.1 & $-$1.6 & $-$2.3 & $-$2.7 \\
\bottomrule
\end{tabular}
\end{table*}

\section{Further results}
\label{sec:s_results}

\subsection{Mind2Web and other backbones}

The marks of Table~\ref{tab:general}'s caption hold for every later table. Tables~\ref{tab:general} and~\ref{tab:s_m2w_next} put the Mind2Web short-side gain over OPSD at $\MGainShort$ points $[\MGainShortlo, \MGainShorthi]$ on order and $\NGainShort$ $[\NGainShortlo, \NGainShorthi]$ on next-step questions. Prior distillation needs a single-form option score, which three of the four further backbones give. Fig.~\ref{fig:tradeoff_full} places every main-model method by short and conflict accuracy.

\begin{table}[t]
\centering
\scriptsize
\setlength{\tabcolsep}{3pt}
\caption{\textbf{Fitted probes.} They meet every reported bound on all seeds and lift conflict accuracy. Qwen3-VL-4B readers, own checks exceeding development bounds. Rates: seed maxima. Within: seeds meeting reported bounds. Blank To prior: only miss and false yes checked.}
\label{tab:s_check}
\begin{tabular}{lrrrrr}
\toprule
 & \multicolumn{4}{c}{Development} & Test \\
\cmidrule(lr){2-5}\cmidrule(lr){6-6}
Check & Miss & False yes & To prior & Within & Conflict \\
\midrule
\multicolumn{6}{l}{\textbf{\emph{Tag-free Mind2Web order}}} \\
Own check & \TFDevMissMax & \TFDevFalseYesMax &  & \TFDevNPassOrder & \HeadTFPSConf \\
Probe & \HeadTFMissMax & \HeadTFFYMax &  & \HeadTFNPass & \HeadTFConf \\
\grayed{Oracle switch} &  &  &  &  & \grayed{\HeadTFOracleConf} \\
\midrule
\multicolumn{6}{l}{\textbf{\emph{HoloAssist order, 4B}}} \\
Own check & \HADevMissMax & \HADevFalseYesMax & \HADevOrderRouteMax & \HADevOrderNPass & \HeadHAPSConf \\
Probe & \HeadHADevMissMax & \HeadHADevFYMax &  & \HeadHANPass & \HeadHAConf \\
\grayed{Oracle switch} &  &  &  &  & \grayed{\HeadHAOracleConf} \\
\midrule
\multicolumn{6}{l}{\textbf{\emph{CaptainCook4D order}}} \\
Own check & \CGoneMissMax &  &  & \CGoneNPass & \HeadCCPSConf \\
Probe & \HeadCCDevMissMax & \HeadCCDevFYMax &  & \HeadCCNPass & \HeadCCConf \\
\grayed{Oracle switch} &  &  &  &  & \grayed{\HeadCCOracleConf} \\
\bottomrule
\end{tabular}
\end{table}

\begin{table}[t]
\centering
\scriptsize
\setlength{\tabcolsep}{3pt}
\caption{\textbf{Threshold.} PriorSplit leads on departures with both readers. HoloAssist next step, seed mean accuracy (\%). Threshold: per reader and question, fitted on development checks (PriorSplit: $0$). Online: next step not yet shown. Departures: next step not among the most counted.}
\label{tab:s_threshold}
\begin{tabular}{lrrrrr}
\toprule
 & \multicolumn{3}{c}{Online} & \multicolumn{2}{c}{Departures} \\
\cmidrule(lr){2-4}\cmidrule(l){5-6}
Reader & PriorSplit & Threshold & Oracle & PriorSplit & Threshold \\
\midrule
Qwen3-VL-4B & $\CalibHAFourNextPSOnline$ & $\CalibHAFourNextCalOnline$ & $\CalibHAFourNextOracleOnline$ & $\CalibHAFourNextPSConf$ & $\CalibHAFourNextCalConf$ \\
Qwen3-VL-8B & $\CalibHAEightNextPSOnline$ & $\CalibHAEightNextCalOnline$ & $\CalibHAEightNextOracleOnline$ & $\CalibHAEightNextPSConf$ & $\CalibHAEightNextCalConf$ \\
\bottomrule
\end{tabular}
\end{table}

\subsection{Ablation and the check}

Table~\ref{tab:s_ablation} gives synthetic ablations. The self-distilled reader and check lift \method to $\MConfPS\%$ on Mind2Web order conflict items, from $\MSwapBaseConf\%$ with the base model in both roles. In Table~\ref{tab:s_gate}, true visibility removes every interval and conflict difference to OPSD. Miss and False yes are the percentages of visible items checked no and invisible items checked yes.

\subsection{Held-out skills and training budget}

In Table~\ref{tab:s_heldout}, \method counts a library of the three held-out skills ($\delta_s = 0.3$), reaching $\HeldoutPS\%$ against $\HeldoutOPSD\%$ for OPSD without frames and matching it with frames. Table~\ref{tab:s_learned_prior} replaces counts with a learned prior, which alone changes the short side by $\DeltaLPriorShort$ points $[\DeltaLPriorShortlo, \DeltaLPriorShorthi]$ and held-out accuracy without frames by $\DeltaLPriorHeldout$ $[\DeltaLPriorHeldoutlo, \DeltaLPriorHeldouthi]$. Trained for five times the steps (Table~\ref{tab:s_budget}), subsequence OPSD and prior distillation change conflict accuracy by $\DFixBMConf$ and $\DVOneBMConf$ points against OPSD and stay below \method there.

\noindent\phantomsection\label{sec:s_heldout_m2w}\textbf{Held-out Mind2Web skills without tags.} In Table~\ref{tab:s_heldout_m2w}, \method with the probe leads OPSD by $\HOTGainShort$ $[\HOTGainShortlo, \HOTGainShorthi]$ short-side points on held-out trajectories and $\HOSGainShort$ $[\HOSGainShortlo, \HOSGainShorthi]$ on held-out websites, and prior distillation by $\HOTGainShortPD$ $[\HOTGainShortPDlo, \HOTGainShortPDhi]$ and $\HOSGainShortPD$ $[\HOSGainShortPDlo, \HOSGainShortPDhi]$. With every frame it matches OPSD, conflicts included. Three subsets each leave out one kind of question: those whose window and order a logged execution also shows, those whose two step texts occur in a training trajectory of the same website, and those whose screenshots have near twins among the training screenshots of the same website. In each, the lead over OPSD stays between $\HOLeadSubMin$ and $\HOLeadSubMax$ points, every lower bound at least $\HOLeadSubLoMin$. On test the probe misses at most $\HOProbeMissMax\%$ of shown steps and says yes on at most $\HOProbeFYMax\%$ of hidden ones.

\noindent\phantomsection\label{sec:s_beyond}\textbf{Three steps, drift and trajectories.} No reader is retrained for these tests. Three-step questions ask which of three consecutive synthetic steps comes first, or their order. The reader answers when all three checks say yes, else a Copeland count of the two pair majorities ranks them. With none or half of six frames shown, PriorSplit leads OPSD by $\TriFirstGain$ $[\TriFirstGainlo, \TriFirstGainhi]$ points on the first step and $\TriOrderGain$ $[\TriOrderGainlo, \TriOrderGainhi]$ on the full order. Counts with a $\DriftHalfLife$-episode half-life, chosen on development questions, restore a $\DriftGain$ $[\DriftGainlo, \DriftGainhi]$ short-side lead $\DriftT$ episodes after a synthetic drift. A trajectory is solved offline when every stored question asked of it in a format is right. On the short side PriorSplit solves $\TrajPS\%$ of $\TrajN$ Mind2Web trajectories ($\TrajGain$ $[\TrajGainlo, \TrajGainhi]$ over OPSD), OPSD $\TrajOPSD\%$ and prior distillation $\TrajPD\%$.

\noindent\phantomsection\label{sec:s_alfworld}\textbf{ALFWorld.} Skills are the six task types of ALFWorld~\cite{alfworld} and units the subgoals of its handcoded expert, whose training runs give counts and episodes. In SKILL0's style~\cite{skill0}, frames are images of completed units' text with the tag banner. On $\AlfNItems$ next-step questions from $\AlfNItemGames$ of the $\AlfNGames$ unseen games, with three Qwen3-VL-4B readers per method, PriorSplit reaches $\AlfItemPS\%$, OPSD $\AlfItemOPSD\%$ ($\AlfItemGainOPSD$ $[\AlfItemGainOPSDlo, \AlfItemGainOPSDhi]$) and prior distillation $\AlfItemPD\%$ ($\AlfItemGainPD$ $[\AlfItemGainPDlo, \AlfItemGainPDhi]$). With each answer as the next-step hint of a Qwen3-VL-8B agent on the $\AlfNGames$ unseen games, task success is $\AlfSuccPS\%$ for PriorSplit, $\AlfSuccOPSD\%$ for OPSD, $\AlfSuccPD\%$ for prior distillation, $\AlfSuccRecipe\%$ with the step list in context and $\AlfSuccNone\%$ without a skill, $\AlfSuccGainNone$ $[\AlfSuccGainNonelo, \AlfSuccGainNonehi]$ points below PriorSplit. PriorSplit calls its reader on $\AlfReaderShare\%$ of steps, while OPSD and prior distillation call theirs on every step. Per episode its agent takes $\AlfStepSavePD$ $[\AlfStepSavePDlo, \AlfStepSavePDhi]$ fewer steps than with prior distillation and reads $\AlfTokSave$ $[\AlfTokSavelo, \AlfTokSavehi]$ fewer prompt tokens than with the step list.

\subsection{Strongest compared method and shown frames}

On seed means and with every $95\%$ interval above zero, \method leads the strongest ranked method of Table~\ref{tab:s_best} on the short side in its first two groups, as on HoloAssist next step with 8B. In Table~\ref{tab:s_protocol}, \method is within $\ProtoBayesGapMax$ points of the oracle reference up to two shown frames and within $\ProtoOPSDGapMax$ of OPSD or above it from four, ahead of every other trained row from twelve.

\subsection{The check on real frames}
\label{sec:s_probe}

The development bounds let a check miss at most $5\%$ of shown named units, say yes on at most $10\%$ of absent ones and route at most $5\%$ of full contexts with both named units visible to the prior. Table~\ref{tab:s_check} compares 4B checks where the own check exceeds a development bound. Mind2Web without tags retains $\HeadTFNConf$ conflict items. Tagged HoloAssist and CaptainCook4D pool every annotated visible conflict across test contexts ($\HeadHANConf$ and $\HeadCCNConf$ items). The probe, fitted on training episodes, takes regularization and threshold from development cells. Own check misses use full contexts and false yes short ones. Probes of Table~\ref{tab:s_check} meet every measured bound on three seeds, using all development cells. The probe matches oracle switch and OPSD conflict accuracy on Mind2Web without tags, held-out Mind2Web skills, HoloAssist order and CaptainCook4D, gaining $\HeadHADeltaConf$ points $[\HeadHADeltaConflo, \HeadHADeltaConfhi]$ and $\HeadCCDeltaConf$ $[\HeadCCDeltaConflo, \HeadCCDeltaConfhi]$ over the own check on the latter two. On CaptainCook4D it gains $\HeadCCGainShort$ short-side points $[\HeadCCGainShortlo, \HeadCCGainShorthi]$ over OPSD.

In Table~\ref{tab:s_threshold}, \method's zero threshold scores higher on departures and one threshold per reader and check question, fitted on development checks, scores higher before the next step is shown.

\noindent\phantomsection\label{sec:s_content}\textbf{Frame content.} With the screenshot area filled with one color at test time and banners and notes kept, the trained readers move little on Mind2Web order: OPSD by $\MaskMOPSDShort$ short-side and $\MaskMOPSDConf$ conflict points, \method by $\MaskMPSShort$ and $\MaskMPSConf$. Its check says yes on $\MaskMCheckYesMask\%$ of shown and no on $\MaskMCheckNoMask\%$ of hidden pairs. With screenshots alone, a constant note and steps named by action text, the base model's order check agrees with the true visibility on $\PixelCheckAgreeMin\%$ to $\PixelCheckAgreeMax\%$ of development checks across wordings, so no reader is trained there. On CaptainCook4D with the banner removed, a constant note and steps named by their description, readers train as on the tagged recipes ($\RawNRec$ test recordings). The base check agrees on $\RawCheckAgreeBase\%$ of development checks (always yes $\RawCheckAlwaysYes\%$). On test the trained one misses $\RawCheckMissMin\%$ to $\RawCheckMissMax\%$ of shown steps with $\RawCheckFYMin\%$ to $\RawCheckFYMax\%$ false yes, and the probe says yes on every full context but one. Frames lift the untrained base model by $\RawBaseFrameGain$ $[\RawBaseFrameGainlo, \RawBaseFrameGainhi]$ points on the departures, about as much as OPSD. PriorSplit's short-side lead there equals that of the prior alone ($\RawPSMinusPriorShort$ $[\RawPSMinusPriorShortlo, \RawPSMinusPriorShorthi]$).

\subsection{Real examples}
\label{sec:s_real_examples}

Fig.~\ref{fig:s_real} shows downscaled Multimodal-Mind2Web~\cite{deng2023mind2web,zheng2024seeact} (OpenRAIL), CaptainCook4D~\cite{captaincook4d} (Apache 2.0) and HoloAssist~\cite{holoassist} (CDLA-Permissive-2.0), with tag banners. Rows show four frames in viewing order, the usual order and its probability. Fig.~\ref{fig:s_qualitative} shows each dataset of Fig.~\ref{fig:qualitative} in its other situation, in the same layout.

Examples use test data and training seed $0$, ordered by seed, cell and question id. Rule (1): a named step has no frame, a check says no, and the counted prior is right where OPSD is wrong. Rule (2): both named steps appear in unusual order, their checks say yes, and \method answers correctly while prior distillation gives the usual order. Fig.~\ref{fig:qualitative} and Fig.~\ref{fig:s_qualitative} together take the first Mind2Web, CaptainCook4D and HoloAssist questions meeting each rule, with prior distillation also wrong on rule (1). The HoloAssist rule (2) row instead has the smallest session id among those with both named steps in at most six keyframes and the prior also giving the usual order. Fig.~\ref{fig:s_real} takes per dataset the question with the larger reader margin among the first two unused rule (2) questions. The teaser asks one Mind2Web question in both rows: the first rule (1) and rule (2) questions on the same skill and named steps where prior distillation is also right on rule (1), OPSD is also right on rule (2), \method changes its answer and both checks match the true visibility, skipping tasks drawn in other figures.

\end{document}